\documentclass[10pt,journal,compsoc]{IEEEtran}

\PassOptionsToPackage{numbers, compress}{natbib}

\ifCLASSOPTIONcompsoc
  \usepackage[nocompress]{cite}
\else
  \usepackage{cite}
\fi
\ifCLASSINFOpdf
\else
\fi

\usepackage{color}
\usepackage{amsmath}
\usepackage{amssymb}
\usepackage{amsthm}
\usepackage{graphicx}
\usepackage{caption}
\usepackage{bm}
\usepackage{mathrsfs}
\usepackage{multirow}
\usepackage{graphicx}
\usepackage{multirow}
\usepackage{graphics}

\allowdisplaybreaks[4]

\newtheorem{theorem}{Theorem}

\newcommand{\cdummy}{\cdot}
\newcommand{\mathd}{\mathrm{d}}
\newcommand{\tmmathbf}[1]{\ensuremath{\boldsymbol{#1}}}
\newcommand{\mathLaplace}{\Delta}

\newcommand{\tmop}[1]{\ensuremath{\operatorname{#1}}}
\newcommand{\assign}{:=}

\begin{document}
%
\title{
Physics-Informed Machine Learning: A Survey on Problems, Methods and Applications}
%
%
%
%

\author{Zhongkai Hao, Songming Liu, Yichi Zhang, Chengyang Ying, Yao Feng,\\ Hang Su, Jun Zhu 
\IEEEcompsocitemizethanks{
\IEEEcompsocthanksitem Zhongkai Hao, Songming Liu, Yichi Zhang, Chengyang Ying, Yao Feng, Hang Su, Jun Zhu are with Dept. of Comp. Sci. $\&$ Techn., Institute for AI, BNRist Center, Tsinghua-Bosch Joint ML Center, Tsinghua University, Email: \{hzj21, liusm18, zyc22, ycy21, y-feng20\}@mails.tsinghua.edu.cn, \{suhangss, dcszj\}@mail.tsinghua.edu.cn,
\IEEEcompsocthanksitem Jun Zhu is also with RealAI.
}
}

\IEEEtitleabstractindextext{%
\begin{abstract}
Recent advances of data-driven machine learning have revolutionized fields like computer vision, reinforcement learning, and many scientific and engineering domains. In many real-world and scientific problems, systems that generate data are governed by physical laws. Recent work shows that it provides potential benefits for machine learning models by incorporating the physical prior and collected data, which makes the intersection of machine learning and physics become a prevailing paradigm. By integrating the data and mathematical physics models seamlessly, it can guide the machine learning model towards solutions that are physically plausible, improving accuracy and efficiency even in uncertain and high-dimensional contexts. In this survey, we present this learning paradigm called Physics-Informed Machine Learning (PIML) which is to build a model that leverages empirical data and available physical prior knowledge to improve performance on a set of tasks that involve a physical mechanism. We systematically review the recent development of physics-informed machine learning from three perspectives of machine learning tasks, representation of physical prior, and methods for incorporating physical prior. We also propose several important open research problems based on the current trends in the field. We argue that encoding different forms of physical prior into model architectures, optimizers, inference algorithms, and significant domain-specific applications like inverse engineering design and robotic control is far from being fully explored in the field of physics-informed machine learning. We believe that the interdisciplinary research of physics-informed machine learning will significantly propel research progress, foster the creation of more effective machine learning models, and also offer invaluable assistance in addressing long-standing problems in related disciplines.
\end{abstract}

}

\maketitle

\IEEEdisplaynontitleabstractindextext

%
\IEEEpeerreviewmaketitle



\section{Introduction}
\label{sec:introduction}

The paradigm of scientific research in recent decades has undergone a revolutionary change with the development of computer technology. Traditionally, researchers used theoretical derivation combined with experimental verification to study natural phenomena. With the development of computational methods, a large number of methods based on computer numerical simulation have been developed to understand complex real systems. Nowadays, with the automation and batching of scientific experiments, scientists have accumulated a large amount of observational data. \emph{The paradigm of (data-driven) machine learning is to understand and build models that leverage empirical data to improve performance on some set of tasks\cite{mitchell1990machine}.}
It is an important research area to promote the development of modern science and engineering technology with the aid of learning from observational data since we could extract a lot of information from data.

As part of the remarkable progress of machine learning in recent years, deep neural networks \cite{lecun2015deep} have achieved milestone breakthroughs in the fields of computer vision \cite{he2016deep}, natural language processing \cite{devlin2018bert}, speech processing \cite{amodei2016deep}, and reinforcement learning \cite{silver2016mastering}. Their flexibility and scalability allow neural networks to be easily applied to many different domains, as long as there is a sufficient amount of data. The powerful abstraction ability of deep neural networks also motivates researchers to apply them on scientific problems in modeling physical systems.
For example, AlphaFold 2 \cite{jumper2021highly} has revolutionized the paradigm of protein structure prediction. Similarly, FourCastNet \cite{pathak2022fourcastnet} has built an ultra-large learning-based weather forecasting system that surpasses traditional numerical forecasting systems. Deep Potential\cite{zhang2018deep} proposed neural models for learning large-scale molecular potential satisfying symmetry. The integration of prior knowledge of physics, which represents a high-level abstraction of natural phenomena or human behaviors, with data-driven machine learning models is becoming a new paradigm since it has the potential to facilitate novel discoveries and solutions to challenges across a diverse range of domains.


Moreover, despite the impressive advancements of machine learning based models, there remain significant limitations when deploying purely data-driven models in real-world applications. 
In particular, data-driven machine learning models can suffer from several limitations such as a lack of robustness, interpretability, and adherence to physical constraints or commonsense reasoning. In computer vision, recognizing and understanding the geometry, shape, texture, and dynamics from images or videos can pose a significant challenge for deep neural networks, which can lead to limitations in their ability to extrapolate beyond their training data. Additionally, such models have demonstrated suboptimal performance outside of their training distribution \cite{shen2021towards} and are susceptible to adversarial attacks via human-imperceptible noise \cite{goodfellow2014explaining}. In deep reinforcement learning, an agent may learn to take actions that result in higher rewards through trial and error, but it may not understand the underlying physical mechanisms. These issues are particularly pertinent in scientific problems where the laws of physics and scientific principles govern the behavior of the system under study.
For example, data obtained from scientific and engineering experiments often tends to be sparse and noisy due to the high cost and the presence of environmental and device-related noise, which can result in significant generalization errors in common machine learning models. One possible explanation for the generalization errors observed in common statistical learning models is their sole reliance on empirical data without incorporating any understanding of the internal physical mechanisms that generate the data. By contrast, humans have the capacity to extract concise physical laws from data, which allows them to interact with the world more efficiently and robustly \cite{karniadakis2021physics,thuerey2021physics}. \emph{The integration of physical laws or constraints into machine learning models, therefore, presents new opportunities for traditional scientific research, substantially advancing the discovery of new knowledge, and facilitating the research in persistent issues of machine learning, such as robustness, interpretability, and generalization \cite{karniadakis2021physics,clark2013whatever}. }

Numerous methods have been proposed by researchers to integrate physical knowledge with machine learning, which are tailored to the specific context of the problem and the representation of physical constraints. While the existing literature on this topic is extensive and multifaceted, \emph{we propose to establish a concise and formalized concept in the form of Physics-Informed Machine Learning (PIML), 
which is a paradigm that seeks to construct models that make use of both empirical data and prior physical knowledge to enhance performance on tasks that involve a physical mechanism.} In this survey, we propose a concise theoretical framework for machine learning problems with physical constraints, based on probabilistic graphical models using latent variables to represent the real state of a system that satisfies physical prior constraints. Our framework provides a unified view of such problems and is flexible in handling physical systems with various constraints, including high-dimensional observational data. It can be combined with methods like autoencoders and dynamic mode decomposition. Moreover, we introduce a physical bottleneck network that can learn low-dimensional, physics-aware representations from high-dimensional, noisy data based on the choice of physical priors.

As an attractive research area, several surveys have been recently published. Karniadakis \cite{karniadakis2021physics} provides a comprehensive overview of the historical development of PIML. Cuomo et al. \cite{cuomo2022scientific} focus on algorithms and applications of PINNs. Beck et al. \cite{beck2020overview} review the theoretical results obtained using NNs for solving PDEs. Other studies have focused on subdomains or applications of PIML, such as fluid mechanics \cite{cai2021physics}, uncertainty quantification \cite{psaros2022uncertainty}, domain decomposition \cite{heinlein2021combining}, and dynamic systems \cite{wang2021physics}. Zubov et al. \cite{zubov2021neuralpde}, Cheung et al. \cite{cheung2021recent}, Blechschmidt et al. \cite{blechschmidt2021three}, Pratama et al. \cite{pratama2021anns}, and Das et al. \cite{das2022state} provide further examples and tutorials with software. Additionally, Rai et al. \cite{rai2020driven}, Meng et al. \cite{meng2022physics}, Willard et al. \cite{willard2020integrating}, and Frank et al. \cite{frank2020machine} focus on other hybrid modeling paradigms that integrate machine learning with physical knowledge. In this survey, we summarize the developments in PIML from the perspective of machine learning researchers, providing a comprehensive review of algorithms, theory, and applications, and proposing future challenges for PIML that will advance interdisciplinary research in this area.

In this review paper, we begin by presenting mathematical preliminaries and background. We then discuss the development of physics-informed machine learning methods for both scientific problems and traditional machine learning tasks, such as computer vision and reinforcement learning. For scientific problems, we focus on representative methods like PINNs and DeepONet, as well as current improvements, theories, applications, and unsolved challenges. We also summarize the methods that incorporate physical prior knowledge into computer vision and reinforcement learning, respectively. Finally, we describe some representative and challenging tasks for the machine learning community.

\section{Problem Formulation}

\begin{figure*}[!t]
    \centering
    \includegraphics[width=1\textwidth]{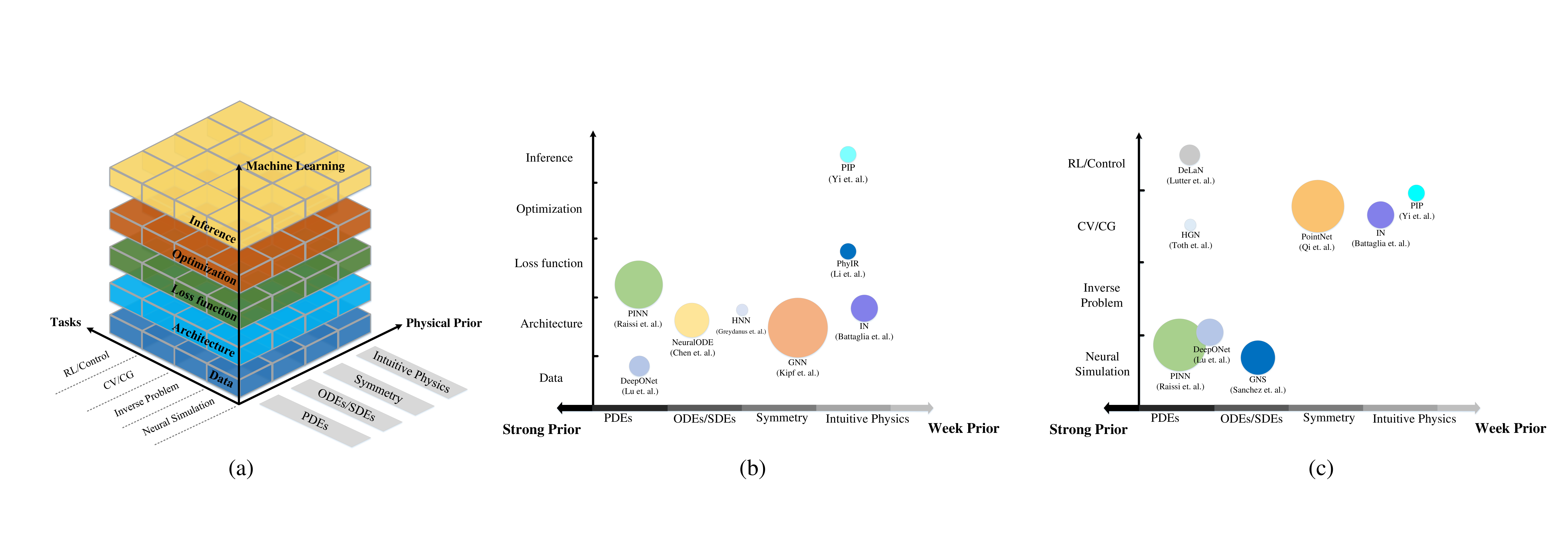}
    \caption{
    An overview of physics-informed machine learning. We review various methods of incorporating physical prior knowledge into machine learning models, ranging from strong to weak forms, such as PDEs/ODEs/SDEs, symmetry, and intuitive physics. These physical priors can be incorporated into different aspects of machine learning models, such as data, model architecture, loss function, optimizer, and inference algorithm. We also highlight different applications of physics-informed machine learning in tasks such as neural simulation, inverse problems, CV/NLP, and RL/control. Finally, we identify some significant areas for exploration in the PIML field, such as physics-informed optimizers and physics-informed inference methods.
    }
    \vspace{-1em}
    \label{summary}
\end{figure*}

In this section, we introduce the concept and commence by examining fundamental problems in physics-informed machine learning (PIML). We elaborate on the representation methodology for physical knowledge, the approach for integrating physical knowledge into machine learning models, and the practical problems that PIML resolves, as is illustrated in Figure~\ref{summary}.

\iftrue

\fi

\subsection{Representation of Physics Prior}


Physical prior knowledge refers to the understanding of the fundamental laws and principles of physics that describe the behavior of the physical world. This knowledge can be categorized into various types, ranging from strong to weak inductive biases, such as partial differential equations (PDEs),   symmetry constraints, and intuitive physical constraints. PDEs, ODEs, and SDEs are prevalent in scientific and engineering domains and can be easily integrated into machine learning models, as they have analytical mathematical expressions. For example, PINNs \cite{raissi2019physics} use PDEs and ODEs as regularization terms in the loss function, while NeuralODE \cite{chen2018neural} construct a neural architecture that obeys ODEs. 

Symmetry and intuitive physical constraints are weaker inductive biases than PDEs/ODEs, which can be represented in various ways, such as designing network architectures that respect these constraints or incorporating them as regularization terms in the loss function. Symmetry constraints include translation, rotation, and permutation invariance or equivariance, which are widely used when designing novel network architectures, e.g., PointNet \cite{qi2017pointnet} and Graph Convolutional Networks (GCN) \cite{kipf2016semi}. 
Intuitive physics, also known as naive physics, is the interpretable physical commonsense about the dynamics and constraints of objects in the physical world. Although intuitive physical constraints are essential and straightforward, mathematically and systematically representing them remains a challenging task. We will elaborate on the different types of physical priors in the following.

\subsubsection{Differential Equations}

Differential equations represent precise physical laws that can effectively describe various scientific phenomena. In this paper, we consider a physical system that exists on a spatial or spatial-temporal domain $\Omega \subseteq \mathbb{R}^d$, where $u (\tmmathbf{x}): \mathbb{R}^d \rightarrow \mathbb{R}^m$ denotes a vector of state variables, which are the physical quantities of interest, and also functions of the spatial or spatial-temporal coordinates $\tmmathbf{x}$. The physical laws governing this system are characterized by partial differential equations (PDEs), ordinary differential equations (ODEs), or stochastic differential equations (SDEs). These equations are known as the \textit{governing equations}, with the unknowns being the state variables $u$. The system is either parameterized or controlled by $\theta \in \Theta$, where $\theta$ could be either a vector or a function incorporated in the governing equations. Unless otherwise stated, Table~\ref{tb1} provides a detailed list of the notations used in the following sections.

\begin{table}[h]
  \begin{tabular}{c|c}
\hline
Notations & Description \\ \hline
$u$ & state variables of the physical system \\ \hline
$\boldsymbol{x}$ & spatial or spatial-temporal coordinates \\ \hline
$x$ & spatial coordinates \\ \hline
$t$ & temporal coordinates \\ \hline
$\theta$ & parameters for a physical system \\ \hline
$w$ & weights of neural networks \\ \hline
$\frac{\partial}{\partial x_i}$ & partial derivatives operator \\ \hline
$\mathcal{D}^k_i$ & $\frac{\partial^k}{\partial x_i^k}$, $k$-order derivatives for variable $x_i$ \\ \hline
$\nabla$ & nabla operator (gradient) \\ \hline
$\Delta$ & Laplace operator \\ \hline
$\int$ & integral operator \\ \hline
$\mathcal{F}$ & differential operator representing the PDEs/ODEs \\ \hline
$\mathcal{I}$ & initial conditions (operator) \\ \hline
$\mathcal{B}$ & boundary conditions (operator) \\ \hline
$\Omega$ & spatial or spatial-temporal domain of the system \\ \hline
$\Theta$ & space of the parameters $\theta$ \\ \hline
$W$ & space of weights of neural networks \\ \hline
$\mathcal{L}$ & loss functions \\ \hline
$\mathcal{L}_r$ & residual loss \\ \hline
$\mathcal{L}_b$ & boundary condition loss \\ \hline
$\mathcal{L}_i$ & initial condition loss \\ \hline
$l_k$ & residual (error) terms \\ \hline
$\| \cdummy \|$ & norm of a vector or a function \\ \hline
\end{tabular}
  
  \caption{A table of mathematical notations.}
  \label{tb1}
\end{table}

\begin{figure*}[t]
    \centering
    \includegraphics[width=18cm]{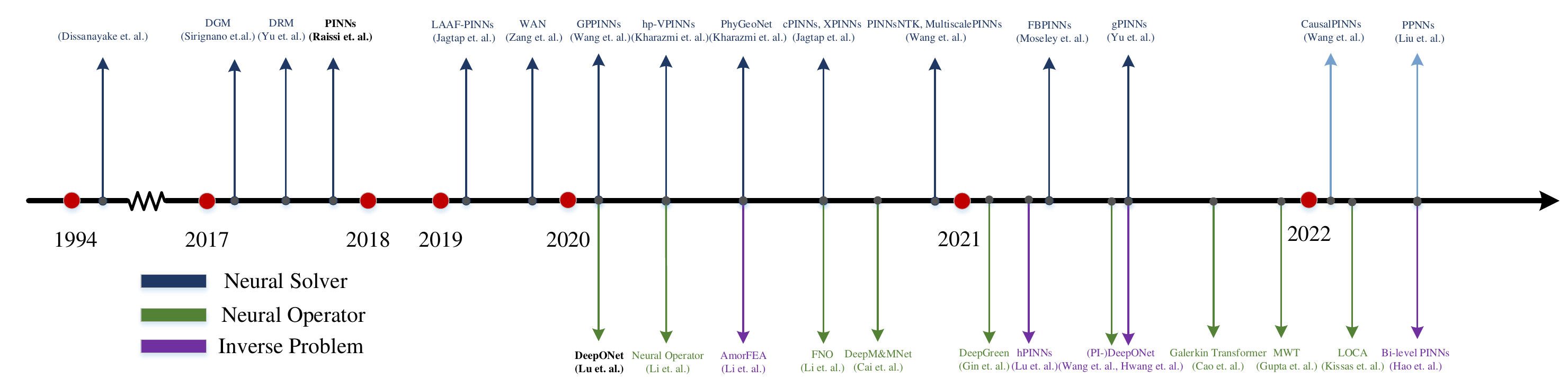}
    \caption{ A chronological overview of important methods for neural simulation (neural solver and neural operator) and inverse problems (inverse design) of physics-informed machine learning. The earliest work could be traced back to \cite{dissanayake1994neural}.}
    \label{timeline}
\end{figure*}




In many domains of science and engineering, real-world physical systems can be modeled using differential equations that are based on different domain-specific assumptions and simplifications. These models can be used to approximate the behavior of these systems. In this paper, we introduce the basic concepts of differential equations. Formally, we consider a system with state variables $u (\mathbf{x}) \in \mathbb{R}^m$, where $\mathbf{x} \in \Omega$ is the domain of definition. For simplicity, we use $\mathbf{x}$ to denote the spatial-temporal coordinates, i.e., $\mathbf{x}=(x_1, \ldots, x_d) \in \Omega$ for time-independent systems and $\mathbf{x}=(x_1, \ldots, x_{d-1}, t) \in \Omega$ for time-dependent systems. The behavior of the system can be represented by ordinary or partial differential equations (ODEs/PDEs) as follows:
\begin{align}
&\text{Partial differential equation: } \mathcal{F} (u ; \theta) (\tmmathbf{x})  =  0, \label{f1} \\
&\text{Initial Conditions: } 
  \mathcal{I} (u ; \theta) (x, t_0)  =  0, \\
&\text{Boundary Conditions: }   \mathcal{B} (u ; \theta) (x, t)  = 0. 
  \label{pde1}
  \end{align}
Without confusion of notation, we rewrite equivalent forms of Eq.~\eqref{f1} as:
\begin{equation}
  \mathcal{F} (u ; \theta) (\tmmathbf{x}) \equiv \mathcal{F} (u, \tmmathbf{x};
  \theta) = 0, x \in \Omega. \label{f2}
\end{equation}
For time-dependent cases (i.e., dynamic systems), we need to pose the initial conditions for state
variables and sometimes their derivatives at a certain time $t_0$ that can be described as 
$ \mathcal{I} (u ; \theta) (x, t_0) = 0, x \in \Omega_0$. 
For systems characterized by PDEs, 
we also need constraints for state variables on the boundary of the spatial domain $\partial \Omega$ to make the system well-posed. For boundary points $ x \in \partial \Omega$, we have the boundary conditions as $\mathcal{B} (u ; \theta) (x, t)  =  0, x \in \partial \Omega$. 
If there are no corresponding constraints of initial conditions and boundary conditions, we can define $\mathcal{I} (u ; \theta) \triangleq 0$ and $\mathcal{B} (u ; \theta) \triangleq 0$.

\subsubsection{Symmetry Constraints}
Symmetry constraints are considered a weaker inductive bias compared to partial differential equations (PDEs) or ordinary differential equations (ODEs). Symmetry constraints refer to a collection of transformations that can be applied to objects,where the abstract set of symmetries is capable of transforming diverse objects. Examples of symmetry constraints are translation, rotation, and permutation invariance or equivariance. In mathematics, symmetries are represented as invertible transformations that can be composed which can be formulated as the concept of groups~\cite{cohen2016group}.

Symmetries or invariants can be incorporated into machine learning to improve the performance of algorithms, depending on the type of data and problem being addressed. There are several types of symmetries, such as translation, rotation, reflection, scale, permutation, and topological invariance, which can be useful in different scenarios~\cite{bronstein2021geometric}. For example, translation invariance is important for data that is shift-invariant, like images or time-series data. Similarly, rotational symmetry is essential for data that is invariant to rotations~\cite{qi2017pointnet}, like images or point clouds, and reflection symmetry is critical for data that is invariant to reflections, such as images or shapes. Scale invariance is useful for data that is invariant to changes in scale, such as images or graphs, while permutation invariance is significant for data that is invariant to permutations of its elements, such as sets or graphs~\cite{kipf2016semi}. Finally, topological invariance is important for data that is invariant to topological transformations, such as shape or connectivity changes.

 The symmetry constraint is that for data $\tmmathbf{x} \in \mathcal{D}$,
there exist an operation $s : \mathcal{D} \rightarrow \mathcal{D}$, \ such
that the property function $\varphi (\cdummy): \mathcal{D}\rightarrow \mathbb{R}^k$ is the same under the symmetric
operation, i.e.
\begin{equation}
  \varphi (\tmmathbf{x}) = \varphi (s (\tmmathbf{x})).
\end{equation}

Incorporating symmetries or invariants can provide numerous advantages for machine learning models. These benefits include improved generalization performance, reduced data redundancy, increased interpretability, and better handling of complex data structures. Symmetries or invariants can aid in improving generalization by providing prior knowledge about the data and by training the model on a representative subset of the data, reducing redundancy. By incorporating symmetries or invariants, we can also gain insights into the underlying structure of the data, making the models more interpretable, especially in scientific or engineering applications. Finally, incorporating symmetries or invariants can be useful for handling complex data structures such as graphs or manifolds, which may not have a simple Euclidean structure. By respecting the underlying geometry of the data, we can design algorithms that can handle these complex symmetries or invariants.

\subsubsection{Intuitive Physics}

Intuitive physics refers to the common-sense knowledge about the physical world that humans possess that they use to reason about and make predictions, such as the understanding that objects fall to the ground when dropped. 
Integrating intuitive physics into machine learning involves incorporating this prior knowledge into the design of machine learning algorithms to improve their performance~\cite{piloto2022intuitive,ye2018interpretable}. There are several commonly used intuitive physics principles that can be incorporated into machine learning models such as ~\cite{duan2022survey}
\begin{itemize}
    \item Object permanence: The understanding that objects continue to exist even when they are no longer visible; 
    \item Gravity: The understanding that objects are attracted to each other with a force proportional to their mass and inversely proportional to the square of their distance; 
    \item Newton's laws of motion: The principles that describe the relationship between an object's motion and the forces acting upon it; 
    \item Conservation laws: The principles that describe the conservation of energy, momentum, and mass in physical systems. 
\end{itemize}

These principles can be used as physical priors or constraints in machine learning models to improve their accuracy, robustness, and interpretability including computer vision, robotics, and natural language processing. For example, object permanence can be used to improve object tracking algorithms by predicting the future location of an object based on its previous motion. Gravity can be used to simulate the behavior of objects in a physical environment, such as in physics-based games or simulations. Therefore, intuitive physics can help us to develop machine learning models that can reason about and predict the behavior of objects in the physical world.

However, intuitive physics is a challenging concept to formalize using traditional mathematical models and equations, hindering its integration into machine learning algorithms. In general, intuitive physics can be incorporated as constraints or regularizers to enhance machine learning models~\cite{raissi2019physics}. For instance, by including the conservation of energy or momentum as constraints, we can design models to predict the behavior of physical systems. Additionally, physical simulations can generate training data for machine learning models, improving their understanding of physical phenomena and validating their performance~\cite{sanchez2020learning,faroughi2022physics}. Finally, hybrid models that combine machine learning and physics can leverage the strengths of both approaches~\cite{xu2021bayesian}. For example, a physics-based model can generate initial conditions for a machine learning model, which can refine those predictions using observed data.


\subsection{Possible Ways towards PIML}
A fundamental issue for PIML is how physical prior knowledge is integrated into machine learning models. As is illustrate in Figure~\ref{summary},
the training of a machine learning model involves several fundamental components including data, model architecture, loss functions, optimization algorithms, and inference. The incorporation of physical prior knowledge can be achieved through modifications to one or more of these components.

Formally, let $\mathcal{D}= \{ (\tmmathbf{x}_i, \tmmathbf{y}_i) \}$ denote a given training dataset. Machine learning tasks can be generally put as searching for a model $f$ from a hypothesis space $\mathcal{H}$. The performance of a particular model on dataset $\mathcal{D}$ is often characterized by a loss function 
$\mathcal{L} (f ; \mathcal{D})$. Then the problem is cast as solving an optimization objective as 
\begin{equation}\label{eq:ML-opt}
  \min_{f \in \mathcal{H}} \mathcal{L} (f ; \mathcal{D}) + \Omega (f) ,
\end{equation}
where $\Omega (f)$ is a regularization term that introduces some inductive bias for better generalization. 
Then, we solve problem \eqref{eq:ML-opt} using an optimizer $OPT(\cdot)$ that outputs a model $f$ from some initial guess $f_0$, i.e., $f  = OPT(\mathcal{H}, f_0)$.

Physics-informed machine learning is a direction of ML that aims to leverage physical prior knowledge and empirical data to improve performance on a set of tasks that involve a physical mechanism. 
Training a machine learning model consisting of several basic components, i.e. data, model architecture, loss functions, optimization algorithms, and inference. 
In general, there are various approaches to incorporating physical prior into different components of machine learning: 
\begin{itemize}
    \item \emph{Data}: we could augment or process the dataset utilizing
available physical prior like symmetry. Mathematically we have $\mathcal{D}_p
= P (\mathcal{D})$ where $P (\cdummy)$ denotes a preprocessing or augmentation
operation using physical prior. 
    \item \emph{Model}: we could embed physical prior into the
model design (e.g., network architecture). We usually achieve this by introducing inductive biases
guided by the physical prior into the hypothesis space, i.e., $f \in
\mathcal{H}_p \subseteq \mathcal{H}$.
    \item \emph{Objective}: we could design better loss
functions or regularization terms using given physical priors like ODE/PDE/SDEs,
i.e. replace $\mathcal{L} (f ; \mathcal{D})$ or $\Omega (f )$
with $\mathcal{L}_p (f ; \mathcal{D})$ or $\Omega_p (f )$.
    \item \emph{Optimizer}: we could design better optimization methods that are more stable or converge faster. We use $OPT_p$ to denote the optimizer that incorporates the physical prior.
    \item \emph{Inference}: we could enforce the physical constraints by using modifying the inference algorithms. For example, we could design a post-processing function $g_p$, we use $g_p (x, f (\tmmathbf{x}))$ instead of $f (\tmmathbf{x})$ when inferencing.
\end{itemize}

First, data could be augmented or synthesized for problems with symmetry constraints or known PDEs/ODEs. Models could learn from these generated data. Second, the architecture of the model may need to be redesigned and evaluated. Physical laws such as PDEs/ODEs, symmetry, conservation laws, and the possible periodicity of data may require us to redesign the structure of the current neural network to meet the needs of practical problems. Third, loss functions and optimization methods for general deep neural networks may not be optimal for training models that incorporate physical constraints. For example, when physical constraints are used as regular term losses, the weight adjustment of each loss function is very important, and commonly used first-order optimizers such as Adam \cite{kingma2014adam} are not necessarily suitable for the training of such models. Finally, for pre-trained machine learning models, we might also design different inference algorithms to enforce physical prior or enhance interpretability.  

First, physical prior knowledge can be integrated into the data by augmenting or synthesizing it for problems with symmetry constraints or known partial differential equations (PDEs) or ordinary differential equations (ODEs). By training models on such generated data, they can learn to account for the physical laws that govern the problem.
Second, the model architecture may need to be redesigned and evaluated to accommodate physical constraints. Physical laws such as PDEs/ODEs, symmetry, conservation laws, and periodicity of data may necessitate a rethinking of the structure of the neural network. Third, standard loss functions and optimization algorithms for deep neural networks may not be optimal for models that incorporate physical constraints. For instance, when physical constraints are used as regular term losses, the weight adjustment of each loss function is crucial, and commonly used first-order optimizers such as Adam are not necessarily suitable for training such models. Finally, for pre-trained machine learning models, different inference algorithms can be designed to enforce physical prior knowledge or improve interpretability. By incorporating physical prior knowledge into one or more of these components, machine learning models can achieve improved performance and better align with practical problems that adhere to the laws of physics.

\subsection{Tasks of PIML}

 Physics-Informed Machine Learning (PIML) can be applied to various problem settings of statistical machine learning such as supervised learning, unsupervised learning, semi-supervised learning, reinforcement learning, etc. However, PIML requires real-world physical processes, and we must have some knowledge about them; otherwise, it would turn into pure statistical learning. The existing works on PIML can be categorized into two classes: using PIML to solve scientific problems and incorporating physical priors to solve machine learning problems. 

 The field of physics-informed machine learning (PIML) has witnessed significant progress in addressing scientific problems that rely on accurate physical laws, often formulated by differential equations. PIML can be classified into two main categories, namely, ``neural simulation'' and ``inverse problems'' related to physical systems~\cite{karniadakis2021physics}. The neural simulation focuses on predicting or forecasting the states of physical systems using physical knowledge and available data. Examples of forward problems include solving PDE systems, predicting molecular properties, and forecasting future weather patterns. In contrast, inverse problems aim to identify a physical system that satisfies the given data or constraints. Examples of inverse problems include scientific discovery of PDEs from data and optimal control of PDE systems. The remarkable advancements in PIML have enabled the development of accurate models and efficient algorithms that combine physical knowledge and machine learning. This integration has opened up new opportunities for interdisciplinary research, enabling insights into complex problems across various fields such as computational biology, geophysics and environmental science~\cite{willard2022integrating}, etc. PIML has the potential to revolutionize scientific discovery and technological innovation. Figure \ref{timeline} shows a chronological summary of recent work proposed in this area. The ongoing research in this field continues to push the boundaries of what is possible.




Incorporating physical knowledge into machine learning models can significantly enhance their effectiveness, simplicity, and robustness. For instance, PIML can improve the efficiency and robustness of robots' design~\cite{bjelonic2023learning}. In computer vision, PIML can improve object detection and recognition and increase models' robustness to environmental changes~\cite{piloto2022intuitive}. PIML can also improve natural language processing models' ability to generate and comprehend text in numerous disciplines, and it can enhance the accuracy and efficiency of reinforcement learning models by integrating physical knowledge~\cite{ramesh2022physics}. By incorporating physical knowledge, PIML can overcome the limitations of traditional machine learning algorithms, which typically require large amounts of data to learn. Nevertheless, representing physical knowledge as physical priors in various domains, where symmetry and intuitive physical constraints prevail, can be more challenging than representing them as partial differential equations. Despite these challenges, the integration of PIML in AI has significant potential to enhance the performance and robustness of AI systems in various fields.




\section{Neural Simulation}
\label{sec:neural simulation}

Using neural network based methods for simulating physical systems governed by PDEs/ODEs/SDEs (named \emph{neural simulation}) is a fruitful and active research domain in physics-informed machine learning. 
In this section, we first list notations and background knowledge used in the paper. Neural simulation mainly consists of two parts, i.e. solving a single PDEs/ODEs using neural networks (named \emph{neural solver}) and learning solution maps of parametric PDEs/ODEs (named \emph{neural operator}). Then we will summarize problems, methods, theory and challenges for \emph{neural solver} and \emph{neural operator} in detail. 

\subsection{Challenges of Traditional ODEs/PDEs Solvers}
Numerical methods are the main traditional solvers for ODEs/PDEs. These methods convert \textit{continuous} differential equations (original ODEs/PDEs or their equivalents) into \textit{discrete} systems of linear equations. Then, the equations are solved on (regular or irregular) meshes. For ODEs, the finite difference methods (FDM) \cite{causon2010introductory} are the most important ones, of which the Runge–Kutta method \cite{butcher1996history} is most representative. The FDM replaces the derivatives in the equations with numerical differences which are evaluated on meshes. For PDEs, in addition to FDM (usually only applicable to geometrically regular PDEs), the finite volume methods (FVM) \cite{eymard2000finite} and the finite element methods (FEM) \cite{felippa2004introduction} are also commonly used mesh-based methods. Such methods consider the integral form equivalent to the original PDEs, and follow the idea of numerical integration to transform the original equations into a system of linear equations. In addition, in recent years, meshless methods (such as spectral methods \cite{bernardi1997spectral}, which are based on the series expansion) have been developed and become powerful solvers for PDEs.

Traditional solvers for ODEs/PDEs are relatively mature, and are of high precision and good stability with complete theoretical foundations. However, we have to point out some of the bottlenecks that severely limit their application. First, traditional solvers suffer from the ``curse of dimensionality''. Supposing that the number of grid nodes is $n$. A crude estimate of the time complexity is given by $\mathcal{O}(dn^r)$ for most traditional solvers \cite{xue2020amortized}, where $d\ge 1$ is the constant and $r$ generally satisfies that $r\approx 3$. Computational cost increases dramatically when the dimensionality of the problem becomes very high, making the computation time of the problem unacceptable. What is more, for nonlinear and geometrically complex PDEs, $d$ is far larger than $1$ and the cost is even worse (for many practical geometrically complex problems, although the dimension is only $3$ or $4$, the computation time can take weeks or even months). Second, traditional solvers have difficulty in incorporating data from experiments and cannot handle situations where the governing equations are (partially) unknown (such as inverse design, described in Section~\ref{sec_inverse_design}). This is because the theoretical basis of the traditional solvers requires the PDEs to be known; otherwise, no meaningful solution will be obtained. Further, these methods are usually not learning-based and cannot incorporate data, which makes it difficult to generalize them to new scenarios. 

Although traditional solvers are still the most widely used at present, they face serious challenges. This provides an opportunity for neural network-based methods. First, neural networks have the potential to resist the ``curse of dimensionality''. In many application scenarios, the high-dimensional data can be well approximated by a much lower-dimensional manifold. \emph{With the help of generalizability, we believe they have the potential to learn such a lower-dimensional mapping and handle high-dimensional problems efficiently; we take the success of neural networks in computer vision \cite{yoo2015deep} as an example. Second, it is easy to incorporate data for neural networks, implicitly enabling knowledge extraction to  enhance prediction results. A simple way is to include the supervised data losses into the loss function and directly train the neural network with some gradient descent algorithm like SGD and Adam \cite{kingma2014adam}.}

\subsection{Neural Solver}
\label{sec_neural_solver}


\subsubsection{Problem Formulation}

This problem aims to solve a single physical system using (partially) known
physical laws and available data. Assume the system is governed by the
ODEs/PDEs in Eq.~\eqref{pde1}.
We also
might have a dataset containing state variables collected
by sensors at some given points $\mathcal{D}= \{ u (\tmmathbf{x}_i) \}_{i = 1, \ldots n}$. \ Our
goal is to solve and represent the state variables of the system $u
(\tmmathbf{x})$. If we use neural networks with weights $w \in W$ to
parameterize the state variables, then
\begin{equation}
  \min_{w \in W}  \| u_w (\tmmathbf{x}) - \tilde{u} (\tmmathbf{x}) \|,
\end{equation}
where $\tilde{u} (\tmmathbf{x})$ is the ground truth state variable. 

The
problem is to use neural networks to represent and solve the state of the
physical system if the physical laws are completely known, to replace
traditional methods like FEMs and FVMs. We call the methods for solving this
problem "neural solvers." There are two potential advantages to use neural methods which might revolutionize numerical simulation in the future. First, the ability and flexibility of neural networks to integrate data and knowledge provide a scalable framework for handling problems with imperfect knowledge or limited data. Second, neural networks, as a novel function representation tool, are shown to be more effective for representing high-dimensional functions, which offers a promising direction for solving high-dimensional PDEs. However, there are still many drawbacks involving computational efficiency, accuracy, and convergence problems for existing neural solvers compared with numerical solvers such as FEM, which has been studied for decades. Thus, how to develop a scalable, efficient, and accurate neural solver is a fundamental challenge in the field of physics-informed machine learning.

In this section, we introduce methods based on neural networks that are
able to incorporate (partially) known physical knowledge (PDEs) for
simulating and solving a physical system. The most representative approach along
these lines is Physics-Informed Neural Networks (PINNs) \cite{raissi2019physics}. First, we introduce the
basic ideas and framework of PINNs. Then, we present different variants of
PINNs that improve PINNs from different viewpoints such as architectures, loss
functions, speed and memory cost, etc. Finally, we propose several unresolved
challenges in the field of neural solvers.
\cite{qi2017pointnet}
\subsubsection{Framework of Physics-Informed Neural Networks }\label{sec_pinn_intro}

PINNs are proposed by \cite{raissi2019physics}, which is the first work that incorporates
physical knowledge (PDEs) into the architecture of neural networks to solve
forward and inverse problems of PDEs. It is a flexible neural network method
that can incorporate PDE constraints into the data-driven learning paradigm.
Suppose there is a system that obeys the PDEs of Equation \eqref{pde1}  and a dataset $\{ u (\tmmathbf{x}_i) \}_{i = 1, \ldots N}$. Then, it is possible to construct
a neural network $u_w (\tmmathbf{x})$ and train it with the following loss
functions as 
\begin{equation}
\begin{split}
  \mathcal{L}= \frac{\lambda_r}{| \Omega |} \int_{\Omega} \| \mathcal{F} (u_w
  ; \theta) (\tmmathbf{x}) \|^2 \mathd \tmmathbf{x}+ \frac{\lambda_i}{|
  \Omega_0 |} \int_{\Omega_0} \| \mathcal{I} (u_w ; \theta) (\tmmathbf{x})
  \|^2 \mathd \tmmathbf{x} \\+ \frac{\lambda_b}{| \partial \Omega |}
  \int_{\partial \Omega} \| \mathcal{B} (u_w ; \theta) (\tmmathbf{x}) \|^2
  \mathd \tmmathbf{x}+ \frac{\lambda_d}{N} \sum_{i = 1}^N \|u_w
  (\tmmathbf{x}_i) - u (\tmmathbf{x}_i)\|^2,  \label{pinn1}
\end{split}
\end{equation}
in which the term $\int_{\Omega}  \| \mathcal{F} (u_w ; \theta) (\tmmathbf{x})
\|^2 \mathd \tmmathbf{x}$ is the (PDE) residual loss that forces the PINNs
$u_w$ to satisfy the PDE constraints; $\int_{\Omega_0} \| \mathcal{I} (u_w ;
\theta) (\tmmathbf{x}) \|^2 \mathd \tmmathbf{x}$ and $\int_{\partial \Omega}
\| \mathcal{B} (u_w ; \theta) (\tmmathbf{x}) \|^2 \mathd \tmmathbf{x}$ are
respectively the initial condition loss and boundary condition loss that force the
PINNs to satisfy the initial condition and boundary condition; $\frac{1}{N}
\sum_{i = 1}^N \|u_w (\tmmathbf{x}_i) - u (\tmmathbf{x}_i)\|^2$ is the regular
data loss in data-driven machine learning that tries to fit the dataset. 

For simplicity of notation, we denote,
\begin{eqnarray}
  l_r & \triangleq & \| \mathcal{F} (u_w ; \theta) (\tmmathbf{x}) \|^2, \\
  l_i & \triangleq & \| \mathcal{I} (u_w ; \theta) (\tmmathbf{x}) \|^2, \\
  l_b & \triangleq & \| \mathcal{B} (u_w ; \theta) (\tmmathbf{x}) \|^2, \\
  l_d & \triangleq & \|u_w (\tmmathbf{x}_i) - u (\tmmathbf{x}_i)  \|^2, 
\end{eqnarray}
and $\Omega_r \triangleq \Omega, \Omega_i \triangleq \Omega_0$, $\Omega_b \triangleq \partial \Omega$, 
$\Omega_d \triangleq\mathcal{D}_d$ so that these losses can be written in a unified
form as
\begin{equation}
  \mathcal{L}_k = \frac{\lambda_k}{| \Omega_k |} \int_{\Omega_k} l_k \mathd
  \tmmathbf{x}, k \in \{ r, i, b, d \} .
\end{equation}

Because the losses of PINNs are flexible and scalable, we can simply omit the
corresponding loss terms if there are no available data or initial/boundary
constraints. \ The learning rate or the weights of these losses can be tuned
by setting the hyperparameters $\lambda_r$, $\lambda_i$, $\lambda_b$ and
$\lambda_d$. In order to compute Equation \eqref{pinn1}, we need to evaluate
several integral terms that involve the high-order derivatives computation of
$u_w (\tmmathbf{x})$. PINNs uses the automatic differentiation of the
computation graph to calculate these derivative terms. Then, it uses
Monte-Carlo sampling to approximate the integral using a set of collocation
points. We use $\mathcal{D}_r$, $\mathcal{D}_i$, $\mathcal{D}_b$ and
${\mathcal{D}_d} $ to represent the dataset of collocation points. We
denote $N_r$, $N_i$, $N_b$ and $N_d$ as the amount of data. Then, the loss function
can be approximated by
\begin{equation}
\begin{split}
  \mathcal{L}= \frac{\lambda_r}{N_r} \sum_{i = 1}^{N_r} \| \mathcal{F} (u_w ;
  \theta) (\tmmathbf{x}_i) \|^2 + \frac{\lambda_i}{N_i} \sum_{i = 1}^{N_i} \|
  \mathcal{I} (u_w ; \theta) (\tmmathbf{x}_i) \|^2 \\ + \frac{\lambda_b}{N_b}
  \sum_{i = 1}^{N_b} \| \mathcal{B} (u_w ; \theta) (\tmmathbf{x}_i) \|^2 +
  \frac{\lambda_d}{N_d} \sum_{i = 1}^N \|u_w (\tmmathbf{x}_i) - u
  (\tmmathbf{x}_i)\|^2 \label{pinn2} .
\end{split}
\end{equation}
Because of the use of automatic differentiation, Equation \eqref{pinn2} is tractable and
can be efficiently trained using first-order methods like SGD and second-order optimizers like L-BFGS. \



\subsubsection{PINN Variants}

\begin{table*}[!t]
\centering
\begin{tabular}{c|c|c|c}
\hline
\multirow{15}{*}{Neural Solver} & Method & Description & Representatives \\ \cline{2-4} 
 & \multirow{3}{*}{Loss Reweighting} & Grad Norm & GradientPathologiesPINNs
\cite{wang2021understanding} \\
 &  & NTK Reweighting & PINNsNTK\cite{wang2022and}  \\
 &  & Variance Reweighting & Inverse-Dirichlet PINNs\cite{maddu2022inverse} \\ \cline{2-4} 
 & \multirow{3}{*}{Novel Optimization Targets} & Numerical Differentiation & DGM \cite{sirignano2018dgm}, CAN-PINN \cite{chiu2022can}, cvPINNs \cite{patel2022thermodynamically} \\
 &  & Variantional Formulation & vPINN \cite{kharazmi2019variational}, hp-PINN\cite{kharazmi2021hp}, VarNet\cite{khodayi2020varnet}, WAN \cite{zang2020weak} \\
 &  & Regularization &  gPINNs \cite{yu2022gradient}, Sobolev Training \cite{son2021sobolev} \\ \cline{2-4} 
 & \multirow{6}{*}{Novel Architectures} & Adaptive Activation & LAAF-PINNs\cite{jagtap2020adaptive, jagtap2020locally}, SReLU\cite{liu2020multi} \\
 &  & Feature Preprocessing & Fourier Embedding \cite{wang2021eigenvector}, Prior Dictionary Embedding \cite{peng2020accelerating} \\
 &  & Boundary Encoding & TFC-based \cite{leake2020deep}, CENN \cite{wang2021cenn}, PFNN \cite{sheng2021pfnn}, HCNet \cite{liu2022unified}  \\
 &  & Sequential Architecture & PhyCRNet\cite{ren2022phycrnet}, PhyLSTM \cite{zhang2020physics} AR-DenseED\cite{geneva2020modeling}, HNN \cite{greydanus2019hamiltonian}, HGN \cite{lutter2019deep} \\
 &  & Convolutional Architecture & PhyGeoNet \cite{gao2021phygeonet}, PhyCRNet \cite{ren2022phycrnet}, PPNN \cite{liu2022predicting} \\
 &  & Domain Decomposition & XPINNs \cite{jagtap2020extended}, cPINNs \cite{jagtap2020conservative}, FBPINNs\cite{moseley2021finite}, Shukla et al. \cite{shukla2021parallel} \\ \cline{2-4} 
 & \multirow{2}{*}{Other Learning Paradigms} & Transfer Learning &  Desai et al. \cite{desai2021one}, MF-PIDNN \cite{chakraborty2021transfer} \\
 &  & Meta-Learning & Psaros et al.\cite{psaros2022meta}, NRPINNs\cite{liu2022novel}  \\ \hline
\end{tabular}
\caption{An overview of variants of PINNs. Variants of PINNs include loss reweighting, novel optimization targets, novel architectures and other techniques such as meta-learning.}
\label{pinns}
\end{table*}

Although PINNs is a concise and flexible framework for solving forward and
inverse problems of PDEs, there are many limitations and much room for
improvement. Roughly speaking, all variants of PINNs focus on developing better optimization targets and neural architectures to improve the performance of PINNs. Here, we briefly summarize the limitations that are addressed by the current variants of PINNs. 
\begin{itemize}
    \item  Different loss terms in PINNs might
have very different convergence speeds; more seriously, these losses might conflict with
each other. To resolve this problem, many variants of PINNs
have proposed different learning rate annealing methods from different perspectives. Some studies have borrowed ideas from traditional multi-task
learning \cite{zhang2021survey}. There are also studies that invent new reweighting schemes, inspired by
theoretical analysis or empirical discoveries of PINNs \cite{wang2021understanding, wang2022and}.

\item PINNs directly penalizes a simple weighted average the PDE residual losses and
initial/boundary condition losses, which might be sub-optimal \cite{kharazmi2019variational, yu2022gradient} for optimization
and training on complex PDEs. Some work has attempted to adopt different loss
functions for optimization that have better convergence and generalization
ability\cite{kharazmi2019variational, yu2018deep, zang2020weak, chiu2022can}. Other work has proposed adding more regularization terms for training PINNs \cite{yu2022gradient, son2021sobolev}.
There is another line of papers that combines the variational formulation with
residual loss of PINNs \cite{yu2018deep, khodayi2020varnet, kharazmi2019variational}.

\item  Many physical systems exhibit extremely
complicated multi-scale and chaotic behaviors, such as shock waves, phase transition,
and turbulence. For these complex phenomena, it is difficult or inefficient to
represent the system using a single MLP architecture. To resolve this
challenge, many studies have proposed specific neural architectures for solving some
PDEs. Some work \cite{zhang2020physics, ren2022phycrnet} has proposed incorporating LSTMs/RNNs, which are more suitable for
processing sequential data into PINNs, to solve time-dependent problems involved in
reducing errors accumulated over a long period of time. Other work \cite{ren2022phycrnet, gao2021phygeonet, liu2022predicting} has proposed mesh-
based representation and has used the architecture of CNNs. To further deal with the complexity
brought about by complex geometric shapes in many practical applications, some work
has designed neural networks using hard constraints for encoding initial/boundary
conditions. Domain decomposition \cite{jagtap2021extended, moseley2021finite} and feature preprocessing techniques \cite{wang2021eigenvector} are 
proposed to handle multi-scale and large-scale problems. Another line of work has proposed to utilize other learning paradigms such as transfer learning \cite{desai2021one, chakraborty2021transfer} and meta-learning \cite{psaros2022meta, liu2022novel} to improve the performance of PINNs.

\end{itemize}

In summary, we use Table (\ref{pinns}) to give the big picture of these variants of PINNs.


\subsubsection{Loss Re-Weighting and Data Re-Sampling}
\label{loss reweight}
A physical system dominated by PDEs usually simultaneously satisfies multiple constraints, such as PDEs and initial and boundary conditions. If we directly optimize it by adding these losses together, as shown in Equation \eqref{pinn1}, there arises a problem. The scale and convergence speed of different losses might be completely different, so that the optimization process might be dominated by some losses, which might converge slowly or converge to wrong solutions  \cite{zhang2021survey}. Existing methods for resolving this problem can be categorized into two classes. One is to re-weight different losses to balance the training process and accelerate the convergence speed. The other is to re-sample data (collocation points) to boost the optimization.  

\textbf{Loss re-weighting.}
Many different studies have proposed loss re-weighting or adaptive learning rate annealing methods by analyzing the training dynamics of PINNs from different perspectives or using different assumptions. \cite{wang2021understanding} is the most famous work that shows that the gradients when training PINNs might be pathological, i.e., the loss of PDE residual is much larger than the boundary condition loss for high frequency functions. The training process is then dominated by the PDE loss, making it difficult to converge to a solution that satisfies boundary conditions. This study also introduced a simple method to mitigate the loss imbalance by re-weighting the learning rates. 
Let $\mathcal{L}_r (w)$ and $\mathcal{L}_i (w)$ respectively be the loss of PDE residual and
other loss terms, i.e., initial/boundary conditions. It computes the update
$\hat{\lambda}_i$ using the following equations at the $n$-th iteration as
\begin{equation}
  \hat{\lambda}_i = \frac{\max \{ \nabla_w \mathcal{L}_r (w_n) \}}{\overline{|
  \nabla_w \mathcal{L}_i (w_n) |}} .
\end{equation}
Then, the learning rate $\lambda_i$ is updated by
\begin{equation}
  \lambda_i \leftarrow (1 - \alpha) \lambda_i + \alpha \hat{\lambda}_i,
  \label{lrmomentum}
\end{equation}
where $\alpha$ is a momentum hyperparameter controlling the update of the
learning rates. Further, \cite{wang2022and} rigorously analyzes the training of PINNs on a Poisson equation using the theory of Neural Tangeting Kernel \cite{jacot2018neural}. It proves that PINNs has a spectral bias, in that it prefers learning low-frequency functions, and therefore high frequency components are hard to converge. Based on this observation, it designs a learning rate annealing method based on NTKs. 

For PDEs with Dirichlet conditions,
they sample a dataset of collocation points from $\Omega$ and $\partial \Omega$,
i.e.$\{ \tmmathbf{x}_i \} \subset \Omega$ and $\{ \tmmathbf{x}_i^b \} \subset
\partial \Omega$. The neural tanget kernel of PINNs is defined by
\begin{equation}
  \tmmathbf{K}= \left(\begin{array}{cc}
    \tmmathbf{K}_{b b} & \tmmathbf{K}_{b r}\\
    \tmmathbf{K}_{b r} & \tmmathbf{K}_{r r}
  \end{array}\right),
\end{equation}
where $\tmmathbf{K}_{u u}$, $\tmmathbf{K}_{u r}$ and $\tmmathbf{K}_{r r}$ are submatrices of the NTK
defined by,
\begin{eqnarray}
  (\tmmathbf{K}_{b b})_{i, j} & = & \left\langle \frac{\mathd u_w
  (\tmmathbf{x}_i^b)}{\mathd w}, \frac{\mathd u_w (\tmmathbf{x}_j^b)}{\mathd
  w} \right\rangle, \\
  (\tmmathbf{K}_{b r})_{i, j} & = & \left\langle \frac{\mathd u_w
  (\tmmathbf{x}_i^b)}{\mathd w}, \frac{\mathd \mathcal{F} (u_w)
  (\tmmathbf{x}_j)}{\mathd w} \right\rangle, \\
  (\tmmathbf{K}_{r r})_{i, j} & = & \left\langle \frac{\mathd \mathcal{F}
  (u_w) (\tmmathbf{x}_i)}{\mathd w}, \frac{\mathd \mathcal{F} (u_w)
  (\tmmathbf{x}_j)}{\mathd w} \right\rangle . 
\end{eqnarray}
The convergence speed of PINNs is decided by the eigenvalues of
$\tmmathbf{K}$. To balance the optimization of PDE residual and boundary
condition losses, we use the trace of the neural tanget kernel to tune the
learning rates $\lambda_b$ and $\lambda_r$ for $\mathcal{L}_b$ and
$\mathcal{L}_r$,
\begin{eqnarray}
  \lambda_b & = & \frac{\tmop{Tr} (\tmmathbf{K})}{\tmop{Tr} (\tmmathbf{K}_{b
  b})}, \\
  \lambda_r & = & \frac{\tmop{Tr} (\tmmathbf{K})}{\tmop{Tr} (\tmmathbf{K}_{r
  r})} . 
\end{eqnarray}
This learning rate annealing scheme is shown to be efficient solving systems
containing multiple frequencies like wave equations.

Another study, \cite{maddu2022inverse}, proposed to use gradient variance to balance the training of PINNs,
\begin{equation}
  \hat{\lambda}_i = \frac{\max_k \{ \tmop{Var} [\nabla_w \mathcal{L}_k (w)]
  \}}{\tmop{Var} [\nabla_w \mathcal{L}_i (w)]} .
\end{equation}
It also uses the momentum update with parameter $\alpha$,
\begin{equation}
  \lambda_i \leftarrow (1 - \alpha) \lambda_i + \alpha \hat{\lambda}_i .
\end{equation}
This approach is called Inverse-Dirichlet Weighting. Experiments show that it alleviates gradients vanishing and catastrophic forgetting in multi-scale modeling.

\cite{leiteritz2021surrogate,van2022optimally} propose to use the characteristic quantities $M_i$ defined as follows,
\begin{equation}
  M_i [u] \approx \frac{\| u_i \|^2_2}{| \Omega |}
\end{equation}
Then, the learning rate for each loss is determined by
\begin{equation}
  \lambda_i = \left( \frac{\sum_k M_k [u]}{M_i [u]} \right)^{- 1} .
\end{equation}
The idea of this method is to approximate the optimal loss weighting under the assumption that error could be uniformly bounded. \cite{leiteritz2021surrogate} also uses a soft penalty method to incorporate learning from data of different levels of fidelity.

To ensure causality, \cite{wang2022respecting} propose to set the learning rates for training PINNs on time-dependent problems to decay with time. Let $\mathcal{L} (t_k, w)$ be the losses at time $t_k$. Then, the total loss is,
\begin{equation}
  \mathcal{L} (w) = \sum_k \lambda_k \mathcal{L} (t_k, w)
\end{equation}
And the weights $\lambda_k$ are
\begin{equation}
  \lambda_i = \exp \left( - \varepsilon \sum_k^{i - 1} \mathcal{L} (t_k, w)
  \right) .
\end{equation}

Besides heuristic methods, \cite{psaros2022meta} attempts to learn optimal weights from data using meta-learning. \cite{rohrhofer2021pareto} investigate the properties of the Pareto front between data losses and physical regularization. \cite{liu2021dual,mcclenny2020self} model the tuning of loss weights as a problem of finding saddle points in a min-max formulation. \cite{liu2021dual} solves the min-max problem using the Dual-Dimer method. \cite{mcclenny2020self} shows connections between the min-max problem and a PDE Constrained Optimization (PDECO) using a penalty method.
Though there are many methods for tuning weights of loss functions, there is no fair and comprehensive benchmark to compare these methods.  

\textbf{Data Re-Sampling.} Another set of methods to handle the imbalance learning process is to re-sample collocation points adaptively. One simple strategy is to sample quasi-random points or a low-discrepancy sequence of points from the geometric domain \cite{das2022state}. This sampling strategy is model-agnostic and only depends on the geometric shape. Representative sampling methods include Sobel sequence \cite{sobol1967distribution}, Latin hypercube sampling \cite{mckay2000comparison}, Halton sequence \cite{berblinger1991monte}, Hammersley sampling \cite{wong1997sampling}, Faure sampling \cite{faure1982discrepance} and so on \cite{daw2022rethinking, wu2023comprehensive}.  

Besides these model-agnostic sampling strategies, another intuitive idea is to sample collocation points from areas with higher error. Thus, we could put more effort into optimizing losses in these areas. Some approaches have designed adaptive sampling strategies based on this idea. Along these lines, PDE residual loss of a vanilla PINNs could viewed as an expectation over a
probability distribution as
\begin{equation}
  \mathcal{L}_r =\mathbb{E}_{\tmmathbf{x} \sim p} [l_r]
  =\mathbb{E}_{\tmmathbf{x} \sim p} [\| \mathcal{F} (u) (\tmmathbf{x}) \|^2],
\end{equation}
and the initial/boundary losses could be described in the same manner. Here, $p$ is
a uniform distribution defined on $\Omega$. In \cite{nabian2021efficient}, the author proposed to sample collocation
points with importance sampling,
\[ \mathcal{L}_r =\mathbb{E}_{\tmmathbf{x} \sim q} \left[ \frac{p
   (\tmmathbf{x})}{q (\tmmathbf{x})} \| \mathcal{F} (u) (\tmmathbf{x}) \|^2
   \right] . \]
Choosing a better probability distribution might accelerate training of PINNs, because
it uses the following distribution to sample a mini-batch of $M$ collocation
points from a dataset of $N$ points uniformly selected ($M < N$),
\begin{equation}
  q (\tmmathbf{x}_i) = \frac{\| \nabla_w l_r (w, \tmmathbf{x}_i) \|}{\sum_j \|
  \nabla_w l_r (w, \tmmathbf{x}_j) \|} \approx \frac{l_r (w,
  \tmmathbf{x}_i)}{\sum_j l_r (w, \tmmathbf{x}_j)}, 1 \leqslant i \leqslant N.
  \label{is}
\end{equation}
However, this requires the evaluation of residuals in a large dataset, which is
inefficient. Therefore, the study proposes using a piece-wise constant approximation to
the loss function to accelerate sampling from the distribution. Note that
Equation \eqref{is} approximates the norm of gradients, with the loss itself similar to \cite{katharopoulos2017biased}.

Further, \cite{tang2021deep} view the losses as a probability distribution and use a generative model to sample from this distribution. Thus, areas with higher residuals contain more collocation points for optimization. Specifically, the distribution of residuals is
\begin{equation}
  q_r (\tmmathbf{x}) = \frac{1}{Z} l_r (\tmmathbf{x}) = \frac{\| \mathcal{F}
  (u) (\tmmathbf{x}) \|^2}{\int_{\Omega} \| \mathcal{F} (u) (\tmmathbf{x})
  \|^2 \mathd \tmmathbf{x}} .
\end{equation}
Sampling from this distribution is not trivial and the authors propose to use a
flow-based generative model \cite{dinh2016density} to sample from the distribution. Similarly, \cite{gu2021selectnet} uses self-paced learning that gradually modifies the sampling strategy from uniform sampling to residual-based sampling.

\subsubsection{Novel Optimization Objectives}
In this section, we describe variants of PINNs that adopt different optimization objectives.
Although various loss re-weighting and data re-sampling methods accelerate convergence of PINNs for some problems, these methods only serve as a trick, since they only allocate different weights for losses but do not modify the losses themselves. 
There is another strand of research that has proposed to train PINNs with novel objective functions rather than weighted summation of residuals. Some studies combine numerical differentiation into PINNs' training process. Some propose to adopt or incorporate variational (or weak) formulation inspired by Finite Element Methods (FEM) instead of PDE residuals. Other approaches propose adding more regularization terms to accelerate training of PINNs.

\textbf{Incorporating Numerical Differentiation.} Vanilla PINNs use automatic differentiation to calculate higher-order derivatives of a neural network with respect to input variables (spatial and temporal coordinates). This method is accurate because we can analytically calculate the derivatives with respect to each layer using backpropagation.
DGM \cite{sirignano2018dgm} points out that computing higher-order derivatives is computationally expensive for high-dimensional problems such as high-dimensional Hamilton-Jacobian-Bellman (HJB) equations \cite{bellman1966dynamic}, which are widely used in control theory and reinforcement learning. This approach proposes to use Monte-Carlo methods to approximate second-order derivatives. Suppose the sum of the second-order derivatives in $\mathcal{L}_r$ is
$\frac{1}{2} \sum_{i, j}^d \rho_{i, j} \sigma_i (x) \sigma_j (x)
\frac{\partial^2 f}{\partial x_i \partial x_j} (t, x ; w)$. Assume $(\rho_{i,
j})_{i, j = 1}^d$ is a positive definite matrix, and define $\sigma (x) =
(\sigma_1 (x), \ldots \sigma_d (x))$. There are many PDEs corresponding to this
case, such as the HJB equation and Fokker-Planck equation. We have the following
equation,
\begin{equation}
\small
\begin{split}
  \sum_{i, j}^d \rho_{i, j} \sigma_i (x) \sigma_j (x) \frac{\partial^2
  f}{\partial x_i \partial x_j} (t, x ; \theta) =  \lim_{\Delta \rightarrow
  0^+} \mathbb{E} \left[\right. \sum^d_i \frac{\sigma_i (x) W_{\Delta}^i}{\Delta}  \\ 
  \left( \frac{\partial f}{\partial x_i} (t, x + \sigma (x) W_{\Delta} ; w) -
  \frac{\partial f}{\partial x_i} (t, x ; w) \right) \left.\right],
\end{split}
\end{equation}
where $W_t \in \mathbb{R}^d$ is a Brownian motion and we choose $\Delta > 0$
as the step size. This reduces the computational complexity from $O (d^2 N)$ to
$O (d N)$. 

 CAN-PINN\cite{chiu2022can} shows that PINNs using automatic differentiation might need a large number of collocation points for training. CAN-PINN uses carefully designed numerical differentiation schemes to replace some terms in automatic differentiation. Specifically, upwind schemes and central schemes \cite{morton2005numerical} are adopted in convection terms, replacing automatic differentiation. 
 
 Control volume PINNs (cvPINNs) \cite{patel2022thermodynamically} borrow the idea of traditional finite volume methods to solve hyperbolic PDEs. This approach partitions the domain into several cells and the PDE losses of hyperbolic conservation laws are transformed into an integral over these cells. Nonlocal PINNs\cite{haghighat2021nonlocal} uses a Peridynamic Differential Operator, which is a numerical method
incorporating long-range interactions, and removes spatial derivatives in the governing equations.

\textbf{Variational formulation.} In traditional FEM solvers, variational (or weak) formulation is an essential tool that reduces the smoothness requirements for choosing basis functions. In variational formulation, the PDEs are multiplied by a set of test functions and transformed into an equivalent form using integrals by parts, as introduced before. The derivative order of this equivalent form is lower than the original PDEs. Although PINNs with smooth activation functions are infinitely differentiable, many studies have shown that there might be potential benefits from adopting the variational (or weak) formulation. In the theory of FEM analysis, solving a PDEs in variational form is equivalent to minimizing an energy function. While this functional form is different from the optimization target of vanilla PINNs, the optimal solution is exactly the same.

 For example, consider a system satisfying the following Poisson's equation
with natural boundary conditions over the boundary:
\begin{eqnarray}
  \Delta u & = & f (\tmmathbf{x}), x \in \Omega, \\
  \frac{\partial u}{\partial n} & = & 0, x \in \partial \Omega . 
\end{eqnarray}
If we use PINNs to solve this problem, we use a neural network $u_w$ to
represent the solution and minimize the following objective:
\begin{equation}
  \mathcal{L} (w) = \frac{\lambda_r}{| \Omega |} \int_{\Omega} \| \Delta u_w -
  f (\tmmathbf{x}) \|^2 \mathd \tmmathbf{x}+ \frac{\lambda_b}{| \partial
  \Omega |} \int_{\partial \Omega} \left\| \frac{\partial u_w}{\partial n}
  \right\|^2 \mathd \tmmathbf{x}.
\end{equation}
The Deep Ritz Method (DRM)\cite{yu2018deep} proposes incorporating the variational
formulation into training the neural networks. Specifically, the objective
function using the variational formulation for this problem is
\begin{equation}
  \mathcal{J} (w) = \int_{\Omega} \left( \frac{1}{2} | \nabla u_w
  (\tmmathbf{x}) |^2 - f (\tmmathbf{x}) u_w (\tmmathbf{x}) \right) \mathd
  \tmmathbf{x}.
\end{equation}
Note that this objective function only involves the first-order derivatives of
$u (\tmmathbf{x})$; thus, we do not need to calculate high-order derivatives.
Additionally, the variational formulation naturally absorbs the natural boundary
conditions, so we do not need to add more penalty terms. This objective
function could also be minimized using gradient descent on weights $w$ similar
to PINNs. If the system satisfies Dirichlet boundary conditions, i.e., if
\begin{equation}
  u (\tmmathbf{x}) = g (\tmmathbf{x}), x \in \partial \Omega,
\end{equation}
we would still need to add a constraint to enforce this type of boundary conditions:
\begin{equation}
\begin{split}
  \mathcal{J} (w) = \int_{\Omega} \left( \frac{1}{2} | \nabla u_w
  (\tmmathbf{x}) |^2 - f (\tmmathbf{x}) u_w (\tmmathbf{x}) \right) \mathd
  \tmmathbf{x}+\\ \lambda_b \int_{\partial \Omega} (u (\tmmathbf{x}) - g
  (\tmmathbf{x}))^2 \mathd \tmmathbf{x}.
  \end{split}
\end{equation}
In fact, the DRM method was proposed even before PINNs. However, DRM is only
available for self-adjoint differential operators, thus limiting its applications.
What's more, \cite{lu2021machine} shows that the fast rate generalization bound of DRM is suboptimal on elliptic PDEs.

Further, in VPINNs \cite{kharazmi2019variational}, the authors propose to develop a Petrov-Galerkin formulation \cite{reddy2019introduction} for training PINNs on more general PDEs. VPINNs consider a broader type of PDEs,
\begin{eqnarray}
  \mathcal{F} (u) (\tmmathbf{x}) & = & 0, x \in \Omega, \\
  u (\tmmathbf{x}) & = & g (\tmmathbf{x}), x \in \partial \Omega . 
\end{eqnarray}
It first chooses a (finite) set of test functions $v (\tmmathbf{x}) \in V_K$,
$N_b$ points from the boundary, and constructs the following loss functions,
\begin{equation}
  \mathcal{J} (w) = \frac{1}{K} \sum_{k = 1}^K | \langle \mathcal{F} (u_w), v
  \rangle_{\Omega} |^2 + \lambda_b \frac{1}{N_b} \sum_{i = 1}^{N_b} | u_w
  (\tmmathbf{x}_i) - g (\tmmathbf{x}_i) |^2 .
\end{equation}
The interior product denotes an integral over the geometric domain,
\begin{equation}
  \langle \mathcal{F} (u_w), v \rangle_{\Omega} = \int_{\Omega} \langle
  \mathcal{F} (u_w) (\tmmathbf{x}), v (\tmmathbf{x}) \rangle \mathd
  \tmmathbf{x}.
\end{equation}
The key of VPINNs is to properly choose test functions according to different
problems. In applications, sine and polynomial functions are good candidates
for test functions. As a special case, if we use a delta function $v
(\tmmathbf{x}, \tmmathbf{x}_0) = \delta (\tmmathbf{x}-\tmmathbf{x}_0)$ as the test
function and $\tmmathbf{x}_0$ as the collocation points, VPINNs are the same
with vanilla PINNs.

In subsequent work, VarNet \cite{khodayi2020varnet} has proposed to take piecewise linear functions as test functions, so that it is more parallelizable and easier to compute the inner products between test functions and neural networks. hp-VPINNs\cite{kharazmi2021hp} proposes to partition the domain into several subdomains and then solve PDEs in these subdomains using variational formulation. The partitioning technique is also called domain decomposition, which will be introduced in detail in subsection \ref{pinn-architecture}. Similar work, such as CENN \cite{wang2021cenn} and D3M\cite{li2019d3m}, also adopt domain decomposition-based variational formulation as loss functions, but they employ other tricks like multi-scale features \cite{wang2021eigenvector} or multi-scale neural networks.
PFNN \cite{sheng2021pfnn} constructs two neural networks and use one of them to enforce essential boundary conditions. Then they use the second neural network to learn from the variational formulation, similar to DGM. By encoding the boundary condition first, it avoids the penalty terms and does not need to tune the weights $\lambda_b$ for them. The boundary encoding technique will be introduced in subsection \ref{pinn-architecture}.

The selection of test functions is crucial for variational formulation-based PINNs. The studies mentioned above chose test functions from a specific function class such as sine or polynomial functions, using priors about the problem. Besides these heuristically chosen test functions, there is another work called Weak Adversarial Networks (WAN)\cite{zang2020weak} that models the training using variational formulation as a min-max problem. Specifically, if the PDEs are strictly satisfied, then for any test function $v \in V$,
\begin{equation}
  \langle \mathcal{F} (u), v \rangle_{\Omega} = 0. \label{wan1}
\end{equation}
Instead of selecting many test functions from a predefined set, WAN chooses the
worst case test function to measure the mismatch of current solution $u_w$. We define the norm of a test function $v$ as $\| v \|_{\Omega} = \sqrt{\langle v,
v \rangle_{\Omega}}$. For problems with natural boundary conditions, we define an operator norm of $\mathcal{F}$ as
follows,
\begin{equation}
  \| \mathcal{F} (u) \|_{\tmop{op}} \assign \max \left\{ \frac{\langle
  \mathcal{F} (u), v \rangle_{\Omega}}{\| v \|_{\Omega}} : v \in H^1_0, v \neq
  0 \right\} .
\end{equation}
If $u$ is the solution of variational formulation in Equation \eqref{wan1}, the
operator norm should be 0. From this perspective, minimizing the operator norm equals solving the variational formulation of PDEs. Then training PINNs is to minimize
the following objective,
\begin{equation}
  \min_{u \in H^1} \| \mathcal{F} (u) \|_{\tmop{op}}^2 .
\end{equation}
In fact,this is a min-max problem like Generative Adversarial Network \cite{goodfellow2014generative}. If we represent solutions and test
functions with neural networks parameterized with $w$ and $\theta$, we have,
\begin{equation}
  \min_w \max_{\theta} \frac{\left| \left\langle \mathcal{F} \left( {u_w} 
  \right), v_{\theta} \right\rangle_{\Omega} \right|^2}{\| v_{\theta}
  \|^2_{\Omega}} .
\end{equation}
This is exactly a loss function for optimizing a GAN and we can use existing
techniques for training GANs to optimize it. For problems with other boundary conditions like Dirichlet/Robin boundary conditions, regularization terms including boundary condition losses should be included when defining the operator norm. \cite{zang2020weak,bao2020numerical} discuss training details and other applications of Weak Adversarial Networks.

Variational (or weak) formulation of PDEs is widely used in Finite Element Method. Such formulation is also shown to be effective for training PINNs in many situations. Many studies have paid attention to selecting appropriate test functions and loss formulations, as introduced before. However, variational form is not the only equivalent form of PDEs. There are other works adopting different formulations of PDEs. For example, BINet\cite{lin2021binet} combines boundary integral equation methods with neural networks for solving PDEs.
In summary, combining other equivalent formulations of PDEs inspired by traditional numerical PDE solvers and PINNs' training is an important topic. Empirical or theoretical analysis on which formulation benefits the training of PINNs has still been largely unexplored.

\textbf{Regularization terms.}

Regularization is an important and simple trick that can boost the training or the generalization ability of machine learning models in many practical applications. In computer vision and machine learning, many regularization terms are proposed according to their effect on the neural networks. For example, $L$-2 regularization \cite{hoerl1970ridge}  can mitigate the overfitting of the model. $L$-1 regularization \cite{tibshirani1996regression} is used to extract sparse features. There are other regularization approaches such as label smoothing \cite{muller2019does} and knowledge distillation \cite{hinton2015distilling}. These methods are called explicit regularization because they add new loss terms that directly modify the gradients computation. Despite existing regularization methods that might also be useful to PINNs, there are novel regularization terms specifically designed for PINNs.

A representative example of these new regularization methods is called
gradient-enhanced training \cite{yu2022gradient}, or Soblev training\cite{son2021sobolev,maddu2022inverse}. The motivation of
gradient-enhanced training is to incorporate higher order derivatives for PDEs
as regularization terms. Since PDEs are a set of identical relations, we can
calculate any order of derivatives of it. Denote $\mathcal{D}_i^k =
\frac{\partial^k}{\partial x_i^k}$ to be the operator of $k$-th order
derivatives for variable $x_i$. Then for all $k, i$, we have
\begin{equation}
  \mathcal{D}^k_i \mathcal{F} (u) (\tmmathbf{x}) = \frac{\partial^k}{\partial
  x^k_i} \mathcal{F} (u) (\tmmathbf{x}) = 0.
\end{equation}
Gradient enhanced training (or Sobleb training) adds regularization
terms based on the derivatives of PDE residuals. Suppose we choose a set of
indexes $K = \{ k_t \}_{t = 1, \ldots m}$ and $I = \{ i_t \}_{t = 1 \ldots m}$,
where $k_t, i_t \in \mathbb{N}_+$ and $1 \leqslant i_t \leqslant d$. Then, the
gradient enhanced regularization is
\begin{equation}
  \mathcal{L}_{\tmop{reg}} = \sum_{k, i \in K, I} \sum_{\tmmathbf{x}_j \in
  \mathcal{D}_r} \lambda_{k, i} \| \mathcal{D}^k_i \mathcal{F} (u)
  (\tmmathbf{x}_j) \|^2 .
\end{equation}
Here, $\mathcal{D}_r = \{ \tmmathbf{x}_j \}$ are the collocation points to
evaluate these regularization terms based on higher order derivatives of PDE residuals. Experiments show that in some situations these regularization terms enable the PINNs to train more quickly and accurately. However, choosing the index sets $K$ and $I$ as the heuristic decision for gradient enhanced PINNs.

\subsubsection{Novel Neural Architectures}\label{sec_pinn_arch}
\label{pinn-architecture}
In this subsection, we introduce variants of PINNs with novel neural architectures for specific problems. Developing proper architectures of neural networks with strong generalization ability is a crucial challenge in machine learning. Although the multi-layer perceptron (MLPs) is a general architecture with the capacity to fit any function, its ability to generalize to many domain-specific problems is suboptimal since it lacks appropriate inductive biases \cite{mitchell1980need}. To incorporate the priors about the data into the model structure, different neural architectures are proposed. For example, for image or grid data, convolutional neural networks (CNNs)\cite{krizhevsky2012imagenet} are proposed because they can extract information from local structures of these types of data. RNN\cite{rumelhart1985learning}, LSTM\cite{hochreiter1997long}
 and transformers \cite{vaswani2017attention}, which have strong ability to model temporal dependency, are proposed to recognize and generate sequential data such as text, audio and time series. Graph neural networks\cite{kipf2016semi} are proposed to extract local node features and global graph features on irregular graph data.
 
 In vanilla PINNs, multi-layer perceptron (MLPs) has been adopted for solving general PDEs, and has achieved remarkable success. However, the architecture of MLP has many drawbacks when solving some domain-specific and complex PDE systems. Therefore, many variants of PINNs have been developed  to improve on architectures for the purpose of adapting to these domain-specific problems. These studies can roughly be divided into several classes. First, the selection of activation functions is noteworthy and many studies have proposed adaptive activation functions for PINNs to deal with the multi-scale structure of physical systems. Second, some work has investigated to embed input spatial-temporal coordinates. These studies propose different feature preprocessing layers, such as Fourier features, to enable learning of different frequencies. Third, architectures like CNNs and LSTMs can be used in PINNs for specific problems. For example, PINNs using convolutional architecture is able to output the whole solution field in one pass rather than the value on a single point. In addition, sequential neural architectures like LSTM can be used to accelerate solving time-dependent PDEs. Fourth, hard boundary constraints can be enforced for some problems with the help of an additional neural network that trains only on boundary condition losses. These methods separate the training of PDEs and initial/boundary conditions in order to avoid the loss imbalance issue. Finally, domain decomposition is proposed to solve large-scale problems. Its purpose is to partition the geometric domain into several subdomains and train a PINNs on each domain to reduce the training difficulty.

\textbf{Activation Functions.}

The nonlinear activation functions play an important role in the expressive power of neural networks. For deep neural networks, ReLU \cite{nair2010rectified}, Sigmoid, Tanh, and Sine \cite{sitzmann2020implicit,wong2021learning} are the most commonly used.
We often need to calculate higher-order derivatives. Therefore, only smooth activation function can be used in PINNs. The Swish activation function is used as a smoothed approximation of ReLU. It is defined as $Swish(x)=x\cdot Sigmoid(\beta x)$, where $\beta$ is a hyperparameter. Despite using existing activation functions, some studies \cite{jagtap2020adaptive,jagtap2020locally} have proposed adaptive activation functions for PINNs to deal with multi-scale physical systems and the gradient vanishing problem. In \cite{jagtap2020adaptive}, the authors propose to use $\sigma(n a\cdot x)$, where $\sigma(\cdot)$ is an activation function, $a$ is a learnable weight and $n$ is  a positive integer hyperparameter to scale the inputs. \cite{jagtap2020locally} further extends the adaptive activation in \cite{jagtap2020adaptive} to two types: layer-wise adaptive activation and neuron-wise adaptive activation. The layer-wise adaptive activation learns one $a$ for each layer. The neuron-wise adaptive activation learns $a$ for each output neuron. It also proposes a special regularization term called the slope recovery term to increase the slope of activation functions. Suppose $\{a_k:1\leq k \leq K\}$ is the set of parameters of adaptive activation functions. The slope recovery regularization term is
\begin{equation}
  \mathcal{J}= \lambda_{reg}\frac{1}{\frac{1}{K} \sum_{k = 1}^K \exp (a_k)},
\end{equation}
where $\lambda_{reg}$ is a hyperparameter.

\cite{liu2020multi} proposes two novel activation functions with compact support,
defined as
\begin{equation}
  \tmop{SReLU} (x) = x_+ (1 - x)_+, \label{srelu}
\end{equation}
\begin{equation}
  \phi (x) = x^2_+ - 3 (x - 1)^2_+ + 3 (x - 2)^2_+ - (x - 3)^2_+, \label{phi}
\end{equation}
where $x_+ = \max \{ x, 0 \} = \tmop{ReLU} (x)$. These two activation functions
look like the RBF kernel but they are compactly supported.

\textbf{Feature Preprocessing (Embedding).}
Feature preprocessing is a basic tool before we feed data into neural networks. Data whitening is an essential preprocessing method widely used in image preprocessing. It normalizes the data to zero means and unit variance. A good feature preprocessing or embedding method might accelerate the training of neural networks. For many practical multi-scale physical systems, we face the challenge that the scale and magnitude are completely different for different parts of the system. For example, for a wave propagation problem in two mediums, the wave length is about $10^3$ times shorter in solid material than in air. It will not make any difference if we directly apply simple normalization to the input coordinates. For these problems with a sharp variation in space or time, the solution usually contains multiple distinct frequencies. \cite{tancik2020fourier} provides a feature embedding method called Fourier features, which was first used in scene representation \cite{mildenhall2020nerf}. 
Suppose $\tmmathbf{x} \in \mathbb{R}^d$ is the input coordinates, and
$\tmmathbf{b}_i \in \mathbb{R}^d$ are scale parameters. Then the Fourier feature
embeds the input coordinates using the following equation:
\begin{equation}
\begin{split}
  \gamma (\tmmathbf{x}) = (\sin (2 \pi \tmmathbf{b}_1^T \cdummy \tmmathbf{x}),
  \cos (2 \pi \tmmathbf{b}_1^T \cdummy \tmmathbf{x}), \ldots,\\ \sin (2 \pi
  \tmmathbf{b}_m^T \cdummy \tmmathbf{x}), \cos (2 \pi \tmmathbf{b}_m^T \cdummy
  \tmmathbf{x})) .
\end{split}
\end{equation}
It embeds the low-dimensional coordinates into high dimensions. The selection
of scale parameters $b_i$ plays a crucial role in Fourier feature embedding.
For instance, in NERF \cite{mildenhall2020nerf}, it uses a geometric series and maps each spatial coordinate
separately:
\begin{equation}
\begin{split}
  \gamma (x_i) = (\sin (2^0 \pi x_i), \cos (2^0 \pi x_i), \ldots \\ \sin (2^{L -
  1} \pi x_i), \cos (2^{L - 1} \pi x_i)) .
\end{split}
\end{equation}
We see that, for large $L$, $\sin (2^{L - 1} \pi x_i)$ changed dramatically even
if $x_i$ varies only a little. This naturally has the effect of scaling the
input. More detailed analysis \cite{tancik2020fourier} based on the theory of the Neural Tangeting Kernel
(NTK) shows that Fourier features make  it easier for the neural networks to learn
high-frequency functions, which mitigates spectral biases. \cite{wang2021eigenvector} further
extends this analysis to the training of PINNs. This approach proposes to sample the
scale parameter $\tmmathbf{b}_i$ from a Gaussian distribution, i.e.,
\begin{equation}
  \tmmathbf{b}_i \sim \mathcal{N} (0, \sigma^2),
\end{equation}
where $\sigma$ is a hyperparameter. It also uses two independent Fourier
features networks to embed the spatial coordinates and temporal coordinates,
respectively. \cite{sitzmann2020implicit,wong2021learning} propose to use sine as the activation functions for neural
networks and the corresponding initialization scheme for weights
$\tmmathbf{b}_i$. \cite{liu2020multi} proposes multi-scale feature embedding based on
the SReLU and $\phi (\cdummy)$ activation functions introduced in Equation
\ref{srelu} and \ref{phi}. We simply use $\sigma$ to denote one of them; it
embeds the coordinates using the following equation:
\begin{equation}
  \gamma (x_i) = \sigma (n x_i), n = 1, \ldots L.
\end{equation}
This formulation can be viewed as both an adaptive activation function and
multi-scale feature embedding. \cite{peng2020accelerating} generalizes the functions used for
feature preprocessing as a prior dictionary. The dictionary includes
trigonometric functions, locally supported functions or learnable functions.
The prior dictionary is flexible; it is chosen based on prior knowledge about the
problem.

We have introduced different activation functions and feature preprocessing layers in PINNs. In the next several subsections, we will introduce variants of PINNs that adopt different network architectures.

\textbf{Multiple NNs and Boundary Encoding.}

The vanilla PINNs inputs the spatial-temporal coordinates and outputs the state variable, which is usually a vector for high-dimensional PDEs. These high-dimensional problems are multi-task learning problems; therefore, vanilla PINNs can be viewed as 
parameter sharing for all tasks. This might lead to suboptimal performance due to the capacity limit of a single MLP. To achieve better accuracy, some studies \cite{haghighat2021physics,lin2021seamless,mojgani2022lagrangian} propose using multiple MLPs without sharing parameters to separately output each component of the state variable. As a simple approach to avoiding parameter sharing, multiple NNs are also used in decomposing the problem into several easier subproblems. First, multiple NNs are used to output intermediate variables and reduce the order of PDEs/ODEs. \cite{rao2020physics} proposes using variable substitution and outputs several intermediate variables using different branches. This method can reduce the order of the PDEs and provides many advantages in practice. Second, multiple NNs with postprocessing layers are used to encode boundary and initial conditions with hard constraints, which will be described in the next several paragraphs. Third, multiple NNs are used in domain decomposition, which will be presented in a later subsection, since it is an independent and comprehensive technique to improve performance of PINNs and is associated with a rich body of literature.

An important technique for many variants of PINNs is to adopt (multiple) NNs with post-processing layers for encoding boundary/initial conditions with hard constraints. In the previous subsection \ref{loss reweight}, we see that balancing losses between PDE residuals and boundary/initial conditions is critical for PINNs. In addition to using adaptive schemes for loss reweighting, another approach is to hard constraint the NNs to satisfy one of them and learn the other one. However, encoding PDEs with hard constraints is only feasible for several simple PDEs with a general solution. 
For example, for the following one-dimensional wave equation,
\begin{equation}
  \frac{\partial^2 u}{\partial t^2} - c^2 \frac{\partial^2 u}{\partial x^2} =
  0,
\end{equation}
it has a general solution,
\begin{equation}
  u = f (x - c t) + g (x + c t) .
\end{equation}
For this equation, we could encode PDE with hard constraints by using NNs to represent $f_w$ and $g_w$. Then, we could train the NNs by fitting $f_w$ and
$g_w$ on boundary/initial conditions. However, most PDEs do not have an analytical general solution. For this reason, most studies focus on encoding boundary/initial conditions with hard constraints rather than PDEs. These studies can be traced back more than two decades \cite{lagaris1998artificial,mcfall2009artificial}. 
For simple boundary conditions like $\Omega = [0, L]$ and $u (0) = u (L) =
0$, we can simply construct a hypothesis space satisfying the constraints
and do not need to use multiple NNs with carefully designed architectures.
Specifically, we can add a simple post-processing layer \cite{lagaris1998artificial,mcfall2009artificial} after the neural
networks $u_w$ 
\begin{equation}
  u_w' = u_w \cdummy x (L - x) .
\end{equation}
This simple method can be extended to Dirichlet boundary conditions on
rectangle domains. If the boundary condition is periodic, we can construct a
neural network \cite{lu2021physics} that outputs a vector $u_w = (u_1, \ldots u_n, v_1, \ldots v_n)$
and represent the solution on the basis of Fourier,
\begin{equation} 
  u_w = \sum_{k = 1}^n \left( u_k (x) \sin \frac{2 \pi k x}{L} + v_k (x) \cos
  \frac{2 \pi k x}{L} \right) .
\end{equation}
These methods are further generalized and unified in the Theory of Functional Connections(TFC) \cite{leake2020deep,schiassi2021extreme}.

For a simple geometric problem, it is possible to design handcrafted post-processing layers for hard constraining boundary/initial conditions. However, they fail to handle problems on a general, irregular
domain. Encoding general boundary conditions, including boundary conditions of arbitrary forms, on irregular domains is still an unresolved problem, despite successful attempts \cite{sun2020surrogate,gao2022physics} for Dirichlet boundary conditions. It is worth noting that recent work \cite{liu2022unified} proposes a unified framework for encoding the three most widely used boundary conditions, i.e., Dirichlet, Neumann, and Robin boundary conditions, on geometrically complex domains, which significantly improves the applicability of hard-constraint methods.

Specifically, as a abstract framework of hard-constraint methods, we decompose the problem into several parts, i.e., a PDE
losses part and a boundary/initial condition part. Then, multiple NNs are used to
solve them separately. The resulting solution is a combination of them. Suppose
the system satisfies the Dirichlet boundary conditions: 
\begin{eqnarray}
  \mathcal{F} (u) (\tmmathbf{x}) & = & 0, x \in \Omega, \\
  u (\tmmathbf{x}) & = & g (\tmmathbf{x}), x \in \partial \Omega . 
\end{eqnarray}
We decompose the solution of the problem into two parts if such decomposition
exists:
\begin{equation}
  u (\tmmathbf{x}) = v (\tmmathbf{x}) + D (\tmmathbf{x}) y (\tmmathbf{x}) .
\end{equation}
Here we first train a neural network $v_w (\tmmathbf{x})$ to satisfy boundary conditions only:
\begin{equation}
  \mathcal{L}_b = \int_{\partial \Omega} | v (\tmmathbf{x}) - g (\tmmathbf{x})
  |^2 \mathd \tmmathbf{x}.
\end{equation}
Then, the function $D (\tmmathbf{x})$ is the key to separating the training of
PDE residuals and boundary/initial conditions. It should be smooth and
vanishes on the boundary, i.e.,
\begin{equation}
  D (\tmmathbf{x}) = 0, x \in \partial \Omega .
\end{equation}
However, it is usually difficult to choose a $D (\tmmathbf{x})$ that is
smooth everywhere for a general domain $\Omega$. Since we train a neural
network only on a set of collocation points, we fit a smooth $D_{w'}
(\tmmathbf{x})$ with a neural network to approximate the following distance
function \cite{berg2018unified}:
\begin{equation}
  d (\tmmathbf{x}) = \min_{\tmmathbf{x}_b \in \partial \Omega} \|
  \tmmathbf{x}-\tmmathbf{x}_b \|_2 .
\end{equation}
We can also train $D_{w'} (\tmmathbf{x})$ on collocation points. Then, the
final step is to train $y (\tmmathbf{x})$ in domain $\Omega$ with only PDE
residuals because the boundary condition is naturally satisfied by using this
decomposition. This method can be extended to initial conditions as
well\cite{sun2020surrogate}.

Many studies that followed the original work have proposed better variants of the distance
function $D (\tmmathbf{x})$. For instance, CENN\cite{wang2021cenn} constructs an (approximation of) $D
(\tmmathbf{x})$ through a linear combination of a radical basis function,
\begin{equation}
  D (\tmmathbf{x}) = \sum_{i = 1}^n w_i \phi (- \| \tmmathbf{x}-\tmmathbf{x}_i
  \|),
\end{equation}
where $\phi (x) = \exp (- \gamma r^2)$ is a radical basis function with
hyperparameter $\gamma$. $\{ \tmmathbf{x}_i : \tmmathbf{x}_i \in \Omega
\}$ is a dataset of collocation points. $y_i = \min_{\tmmathbf{x} \in
\partial \Omega} \| \tmmathbf{x}_i -\tmmathbf{x} \|$ is the distance of these
collocation points to the boundary, which can be precomputed. We can solve
$w_i$ using the following linear equation:
\begin{equation}
\footnotesize
  \left(\begin{array}{ccc}
    \phi (\| \tmmathbf{x}_1 -\tmmathbf{x}_1 \|) & \ldots & \phi (\|
    \tmmathbf{x}_1 -\tmmathbf{x}_n \|)\\
    \vdots & \ddots & \vdots\\
    \phi (\| \tmmathbf{x}_n -\tmmathbf{x}_1 \|) & \ldots & \phi (\|
    \tmmathbf{x}_n -\tmmathbf{x}_n \|)
  \end{array}\right) \cdummy \left(\begin{array}{c}
    w_1\\
    \vdots\\
    w_n
  \end{array}\right) = \left(\begin{array}{c}
    y_1\\
    \vdots\\
    y_n
  \end{array}\right) .
\end{equation}
The meaning of $D (\tmmathbf{x})$ is to interpolate a distance function with a
dataset of collocation points with radical basis functions.

For PFNN, \cite{sheng2021pfnn}, proposes a novel distance function $D
(\tmmathbf{x})$ for multiple complex boundaries. It divides the boundary $\partial
\Omega$ into several segments $\{ \gamma_i : 1 \leqslant i \leqslant K \}$ and
constructs a $D (\tmmathbf{x})$ based on these segments. For a given $\gamma_k$
and a non-neighbor segment $\gamma_{k_0}$, it defines a spline function $l_k$
to satisfy the following property:
\begin{equation}
  \text{} \left\{\begin{array}{ll}
    l_k (\tmmathbf{x}) = 0, & \tmmathbf{x} \in \gamma_k\\
    l_k (\tmmathbf{x}) = 1, & \tmmathbf{x} \in \gamma_{k_0}\\
    0 \leqslant l_k (\tmmathbf{x}) \leqslant 1, & \tmop{otherwise}
  \end{array}\right.
\end{equation}
It defines a type of indicator function that vanishes only on a certain
segment of the boundary. Then, we define the overall $l (\tmmathbf{x})$,
\begin{equation}
  l (\tmmathbf{x}) = \prod_{k = 1}^K 1 - (1 - l_k (\tmmathbf{x}))^{\mu},
\end{equation}
where $\mu \geqslant 1$ is a hyperparameter. Finally, we define $D
(\tmmathbf{x})$ as follows:
\begin{equation}
  D (\tmmathbf{x}) = \frac{l (\tmmathbf{x})}{\max_{\tmmathbf{x} \in \Omega} l
  (\tmmathbf{x})} .
\end{equation}
In practice, $l_k (\tmmathbf{x})$ is constructed by a combination of a radical basis
function and a linear function; because this is complicated, we omit the
details here. We see that the $D (\tmmathbf{x})$ is then smooth and vanishes
on all boundary segments.

\textbf{Sequential Neural Architecture.}
A large amount of work in machine learning is about specific network architectures to process sequential data such as text, audio, and time series. By now, there are many famous architectures for sequential data recognition, including Recurrent Neural Networks (RNN)\cite{rumelhart1985learning}, Long-Short Term Memory network (LSTM)\cite{hochreiter1997long}, Gated Recurrent Unit (GRU)\cite{chung2014empirical}, Transformer \cite{vaswani2017attention} and so on. In the field of physics, many real physical systems are time-dependent; therefore, future states rely on the past states of systems. These systems can be naturally modeled as sequential data. Along this line, many studies \cite{zhang2020physics,ren2022phycrnet,geneva2020modeling} propose to combine these neural architectures to train PINNs. A typical example is the following time-dependent PDEs:
\begin{equation}
  \frac{\partial u}{\partial t} + F \left( u, \frac{\partial u}{\partial x_1},
  \ldots, \frac{\partial u}{\partial x_d}, \ldots ; \theta \right) = 0.
\end{equation}
Vanilla PINNs builds a neural network that inputs $u (x_1, \ldots x_d, t)$
and updates the model using the PDE residual. If we adopt a sequential neural
architecture to solve the problem, we first discretize $t \in [0, T]$ into
several $n$ time slots $\{ t_i : t_i = i \Delta t, 1 \leqslant i \leqslant n
\}$. Then, we use numerical differentiation to approximate the derivatives
$\frac{\partial u}{\partial t}$. The loss is then constructed by:
\begin{equation}
  \mathcal{L}_{\tmop{reg}} = \left\| \frac{u_{i + 1} - u_i}{\mathLaplace t} -
  F \left( u_i, \frac{\partial u_i}{\partial x_1}, \ldots \frac{\partial
  u}{\partial x_d}, \ldots, \theta \right) \right\|^2 .
\end{equation}
Here, $u_i$ is the output of the neural networks at time $t_i$. In many studies \cite{zhang2020physics,geneva2020modeling},
LSTMs are used to represent the solution $u_i$. We see that, by using sequential
architecture, we can transform the problem into a set of time-independent PDEs. 

As well as work using LSTMs to solve general time-dependent problems, another line of work proposes a sequential architecture combining numerical differentiation to solve a specific class of systems governed by Newton's laws. In physics, solving or identifying dynamic systems governed by Newton's laws (or Hamiltonian, Lagrangian equations) is a fundamental issue. It has a wide range of applications in physics, robotics, mechanical engineering and molecular dynamics. There is a great deal of work designing specific neural architectures that naturally obey Hamiltonian equations and Lagrangian equations \cite{cranmer2020lagrangian,lutter2019deep,greydanus2019hamiltonian,toth2019hamiltonian}. 

Hamiltonian equations are a class of basic and concise first-order equations to describe temporal evolution of physical systems. For Hamiltonian systems, states are $(\tmmathbf{q}, \tmmathbf{p})$, where
$\tmmathbf{q}$ represents the coordinates and $\tmmathbf{p}$ represents the momentum of the
system. Hamiltonian neural networks (HNN) \cite{greydanus2019hamiltonian,bertalan2019learning} represent the Hamiltonian with a
neural network $\mathcal{H}_w (\tmmathbf{q}, \tmmathbf{p})$. The evolution of
the system is determined by
\begin{eqnarray}
  \frac{\mathd \tmmathbf{q}}{\mathd t} \approx \frac{\tmmathbf{q} (t + \Delta
  t) -\tmmathbf{q} (t)}{\Delta t} & = & \frac{\partial \mathcal{H}_w}{\partial
  \tmmathbf{p}}, \\
  \frac{\mathd \tmmathbf{p}}{\mathd t} \approx \frac{\tmmathbf{p} (t + \Delta
  t) -\tmmathbf{p} (t)}{\Delta t} & = & - \frac{\partial
  \mathcal{H}_w}{\partial \tmmathbf{q}} . 
\end{eqnarray}
By using numerical differentiation, Hamiltonian systems naturally evolve. We can learn the Hamiltonian from data using the
following residual:
\begin{equation}
  \mathcal{L}_r = \left\| \frac{\mathd \tmmathbf{q}}{\mathd t} -
  \frac{\partial \mathcal{H}_w}{\partial \tmmathbf{p}} \right\|^2_2 + \left\|
  \frac{\mathd \tmmathbf{q}}{\mathd t} + \frac{\partial
  \mathcal{H}_w}{\partial \tmmathbf{q}} \right\|^2_2 .
\end{equation}
Some work proposes advanced integrators or improved architecture \cite{zhu2020deep,zhong2019symplectic,sanchez2019hamiltonian,jin2020sympnets} for more accurate prediction. HGN \cite{toth2019hamiltonian} combines generative models such as variational auto-encoders (VAE) \cite{kingma2013auto} and Hamiltonian neural networks to model time-dependent systems with uncertainty.

\textbf{Convolutional Architectures.}
Convolutional neural networks are widely used in image processing and computer vision. Convolution utilizes the local dependency of pixels on image data to extract semantic information. In the field of numerical computing, certain convolutional kernels can be viewed as a numerical approximation of differential operators. Many studies \cite{zhang2020physics,gao2021phygeonet,ren2022phycrnet,chen2021theory} exploit this connection between convolutional kernels and (spatial) differential operators to develop convolutional neural architectures for physics-informed machine learning. 
Specifically, a one-dimensional Laplace operator can be approximated
(discretized) by
\begin{equation}
  D_1 \approx \frac{1}{h^2} \left(\begin{array}{ccc}
    1 & - 2 & 1
  \end{array}\right) .
\end{equation}
Similarly, a two-dimensional Laplace operator can be approximated by
\begin{equation}
  D_2 \approx \frac{1}{h^2} \left(\begin{array}{ccc}
    0 & 1 & 0\\
    1 & - 4 & 1\\
    0 & 1 & 0
  \end{array}\right) \approx \frac{1}{4 h^2} \left(\begin{array}{ccc}
    1 & 2 & 1\\
    2 & - 12 & 2\\
    1 & 2 & 1
  \end{array}\right) .
\end{equation}
The former convolutional kernel is called a five-point stencil; the
latter is called a nine-point stencil and has a higher approximation
order. Similarly, we can define convolutional kernels to approximate
a Laplace operator in higher dimensions. By discretizing state variables $u$ and
applying these discretized convolutional operators to them, we can
approximate the function of Laplace operators. Suppose we discretize a
two-dimensional state variable $u (x, y)$ based on a mesh (grid) $U = (u_{i
j})_{1 \leqslant i, j \leqslant n}$. Then we have
\begin{equation}
  \Delta u (x, y) \approx D_2 \ast U,
\end{equation}
where $\ast$ denotes the (discretized) convolution operation. We also can numerically
represent other differential operators by using different convolutional
kernels. By discretizing states and differential operators in spatial
dimensions, we can naturally use convolutional neural architectures to solve
PDEs or learn from data. Another advantage of discretization is that the
Dirichlet boundary condition and initial condition can be easily satisfied
by assigning boundary/initial points to given values. These convolutional neural architectures are usually jointly used with recurrent architecture like LSTMs as a  Conv-LSTM network for learning spatial-temporal dynamic systems \cite{ren2022phycrnet,zhang2020physics}. \cite{liu2022predicting} proposes a novel Conv-ResNet based architecture with a PDE preserving part and a learnable part for solving forward and inverse problems. It also introduces a U-Net architecture \cite{ronneberger2015u} with an encoder and a decoder to extract multiple-resolution features.

However, a limitation of vanilla CNN architecture is that it can only be used in a regular grid. For problems with complex irregular geometric domains, new methods need to be developed. \cite{gao2021phygeonet} proposes a parameterized coordinate transformation from the irregular physical domain to a regular reference domain,
\begin{equation}
  \tmmathbf{x}=\mathcal{G} (\tmmathbf{\xi}), \tmmathbf{\xi}=\mathcal{G}^{- 1}
  (\tmmathbf{x}),
\end{equation}
where $\tmmathbf{x} \in \Omega_p$ is the irregular physical domain and
$\tmmathbf{\xi} \in \Omega_r$ is the regular reference domain. This map needs
to be a bijection to ensure its reversibility. Then, we need to transform the
PDEs into the reference domain by using a theorem of variable substitution,
\begin{equation}
  \frac{\partial}{\partial x_i} = \sum_j \frac{\partial}{\partial \xi_j}
  \frac{\partial \xi_j}{\partial x_i} . \label{tvs}
\end{equation}
For a higher-order differentiable operator, we can simply apply the
discretized version of Equation \eqref{tvs} to avoid a complicated theoretical
derivation. Finding an analytical mapping of $\mathcal{G}$ is impossible as a
practical matter; a feasible solution is to calculate and store the mapping
and its inverse numerically \cite{thompson1974automatic}.  Besides using a coordinate transformation from an irregular domain to a regular reference domain, there are some studies that use graph networks \cite{sanchez2020learning,pfaff2020learning,gao2022physics} to learn (next-timestep) simulation results from data with PDE as inductive biases. \cite{han2022predicting} improves the performance of the graph network based architecture by introducing attention layers over the temporal dimension. \cite{jiang2022phygnnet, lotzsch2022learning} also adopts graph neural networks to improve operator learning.

\textbf{Domain Decomposition.}
Domain decomposition is a basic and effective framework to improve the performance of PINNs on large-scale problems or multi-scale problems. It partitions a domain into many subdomains and solves an easier problem in each subdomain using smaller subnetworks. To ensure the consistency and continuity of the whole solution, additional loss terms are imposed at the interfaces between these subdomains. This is a general framework for improving PINNs and many techniques introduced in previous sections can also be combined with domain decomposition. 

Suppose we have a domain $\Omega$ and it is decomposed into many subdomains
$\Omega = \bigcup_{k = 1}^K \Omega^k$. The PDEs in each subdomain are
\begin{eqnarray}
  \mathcal{F}^k (u ; \theta^k) (\tmmathbf{x}) & = & 0, x \in \Omega^k, \\
  \mathcal{B}^k (u ; \theta^k) (\tmmathbf{x}) & = & 0, x \in \partial
  \Omega^k, \\
  \mathcal{I}^k (u ; \theta^k) (\tmmathbf{x}) & = & 0, x \in \Omega_0^k . 
\end{eqnarray}
In each domain, we sample datasets of collocation points $\mathcal{D}_r^k = \{
\tmmathbf{x}_i^k : \tmmathbf{x}_i \in \Omega^k \}$ and boundary/initial
collocation points $\mathcal{D}_b^k = \{ \tmmathbf{x}_i^k : \tmmathbf{x}_i^k
\in \partial \Omega^k \}, \mathcal{D}_i^k = \{ \tmmathbf{x}_i^k :
\tmmathbf{x}_i^k \in \Omega^k_0 \}$. Note that mathematically subdomains could
be disjoint. However, in practice these subdomains need to have overlapping
area to allow the subnetworks training on these subdomains to communicate
with each other. This is necessary to ensure consistency of the solution. We call these
overlapping areas the interface and we denote $\{ I^m : I^m \subset \Omega, 1
\leqslant m \leqslant M \}$ to be the set of these interfaces. We can then
sample collocation points from interfaces $\mathcal{D}_I^m = \{
\tmmathbf{x}^m_i : \tmmathbf{x}_i^m \in I^m \} .$ For simplicity of
notation, we use $u^+$ and $u^-$ to denote two subnetworks' learning on two
adjacent subdomains for any interface. In each subdomain $\Omega^k$, we
parameterize the state variables $u$ with a (small) neural network $u_k
(\tmmathbf{x})$ parameterized by $w_k$. Denote all weights \ to be $w = (w_1,
\ldots w_K)$. The general training losses for PINNs with domain decomposition
are
\begin{equation}
  \mathcal{L}= \sum_{k = 1}^K (\lambda_r^k \mathcal{L}_r^k + \lambda_b^k
  \mathcal{L}_b^k + \lambda_i^k \mathcal{L}_i^k) + \sum_{m = 1}^M \lambda_I^m
  \mathcal{L}_I^m .
\end{equation}
Here, $\mathcal{L}_r^k, \mathcal{L}_b^k, \mathcal{L}_i^k$ are subdomain losses
for each subdomain $\Omega_k$. $\mathcal{L}_I^m$ is called the interface loss.
Subdomain losses can be optimized independently for each subnetwork.
However, because interface losses are communication losses between these subnetworks, they should be optimized together as a message passing from one subnetwork
to another. There are different approaches for the decision choice of interface losses and \ subdomain losses. Here, we will introduce several representative
studies adopting domain decomposition techniques.

cPINNs\cite{jagtap2020conservative} considers systems obeying conservation laws:
\begin{equation}
  \frac{\partial u}{\partial t} + \nabla \cdummy f (u, u_x, u_{x x}, \ldots) =
  0.
\end{equation}
It decomposes the whole domain into the subdomains mention above and designs the
following interface loss:
\begin{equation}
\begin{split}
      \mathcal{L}_I^m = \frac{1}{N_m} \sum_{i = 1}^{N_m} ( \lambda_{\tmop{Avg}} | u^- (\tmmathbf{x}^m_i) -
  u^+ (\tmmathbf{x}^m_i) |^2 + \\\lambda_{\tmop{flux}} |
  f (u^- (\tmmathbf{x}^m_i)) \cdummy \tmmathbf{n}- f (u^+ (\tmmathbf{x}_i^m))
  \cdummy \tmmathbf{n} |^2 ) .
\end{split}
\end{equation}
The first loss term penalizes the difference of two subnetworks on the
interface and the second loss term enforces the flux through the interface to be the same
. For inverse problems, learned parameters for different subdomains
should also be penalized,
\begin{equation}
  \mathcal{L}_{\tmop{reg}} = \frac{1}{N_m} \sum_{i = 1}^{N_m} (\theta^+
  (\tmmathbf{x}^m_i) - \theta^- (\tmmathbf{x}^m_i))^2 .
\end{equation}
However, penalizing the flux between interfaces is only feasible for systems
satisfying the conservative laws. To resolve the limitation, XPINNs \cite{jagtap2020extended,jagtap2021extended} further
proposes a generalized version of interface conditions. XPINNs proposes the
following interface conditions:
\begin{equation}
\small
\begin{split}
     \mathcal{L}_I^m = \frac{1}{N_m} \sum_{i = 1}^{N_m} \left(\right.  \lambda_{\tmop{Avg}}
  \| u^- (\tmmathbf{x}^m_i) - u^+ (\tmmathbf{x}^m_i) \|^2 + \\\lambda_F \left\|
  \mathcal{F}^- (u^- ; \theta^-) (\tmmathbf{x}^m_i) -\mathcal{F}^+ (u^+ ;
  \theta^+) \left( \tmmathbf{x}^m_i \right) \right\|^2 \left.\right) .
\end{split}
\end{equation}
The second term is different from cPINNs since it enforces the continuity of
general PDE residuals rather than flux continuity. It also is more flexible since
it allows the same code for both forward and inverse problems. It also makes it possible to add more domain-specific penalty terms in practical usage.

cPINNs and XPINNs are two basic methods using the domain decomposition framework. There are many methods that enhance domain decomposition with other techniques to improve performance.
For FBPINNs, \cite{moseley2021finite} proposes to  apply different weighted normalization layers to different subdomains to handle multi-scale problems. \cite{shukla2021parallel} scales cPINNs and XPINNs to problems with a larger scale with both optimized hardware and software implementation based on the MPI framework. This approach is able to train PINNs efficiently with multiple GPUs. \cite{meng2020ppinn} combines domain decomposition in temporal dimensions with a traditional ODE solver to boost accuracy. \cite{huang2022parallel} introduces adaptive weight balancing for interfaces based on the intersection of union (IoU). \cite{li2019d3m,kharazmi2021hp} combine variational formulation with domain decomposition. \cite{stiller2020large} proposes to use an architecture-gated mixture of experts (MoE)\cite{shazeer2017outrageously}, which can be viewed as a soft version of domain decomposition, since it does not explicitly divide subdomains and subnetworks.

\subsubsection{Open Challenges and Future Work.}

Though many attempts have been made to improve the convergence speed and accuracy of PINNs, there is still much room for improvement, which is left for future work. Here, we present several important topics that are far from being fully explored.
\begin{itemize}
    \item \textbf{Optimization Process}. The optimization method of PINNs can be improved in many respects. The training process of PINNs is significantly different from ordinary neural networks. Current optimizers and loss functions might not be optimal for PINNs. However, existing attempts are not ready, either theoretically or experimentally, to build a stable and effective neural solver.
    \item \textbf{Model Architecture} The neural network architecture, like other fields of deep learning, still needs more study. In the frontiers of deep learning, there are many novel architectures, such as normalization layers, and transformer architectures have been proposed which are shown to be superior in multiple domains. However, the application of these architectures in the field of physics-informed machine learning is far from completely explored. 
    \item \textbf{Solving High Dimensional Problems.}
    High dimensional PDEs like HJB equations and Schrödinger equations play a key role in science and engineering but they are notoriously difficult to solve due to the curse of dimensionality. The efficiency of neural networks representing high dimensional functions provides a promising advantage of neural solvers using physics-informed machine learning \cite{yu2018deep, han2018solving}. This is an open challenge for solving high dimensional PDEs with a wide range of applications such as quantum mechanics, molecular dynamics and control theory using neural networks.
    
\end{itemize}

\begin{table*}[]
\centering
\begin{tabular}{l|l|l}
\hline
Category \& Formulation &
  Rrepresentative &
  Description \\ \hline
\multirow{6}{*}{\begin{tabular}[c]{@{}l@{}}\textbf{Direct Methods}\\ \\ $G_w (\theta) (\tmmathbf{x})=b_0 + \sum_{k=1}^p b_k(\theta) t_k(\tmmathbf{x})$\end{tabular}} &
  DeepONet \cite{lu2021learning} &
  \begin{tabular}[c]{@{}l@{}}Parameterize $b_k$ and $t_k$ with neural networks,\\ which are trained with supervised data.\end{tabular} \\ \cline{2-3} 
 &
  \begin{tabular}[c]{@{}l@{}}Physics-informed DeepONet \\ \cite{wang2021learning}\end{tabular} &
  \begin{tabular}[c]{@{}l@{}}Train DeepONet with a combination of data and \\ physics-informed losses.\end{tabular} \\ \cline{2-3} 
 &
  \begin{tabular}[c]{@{}l@{}}Improved Architectures\\ for DeepONet \cite{wang2022improved,lu2022comprehensive}\end{tabular} &
  \begin{tabular}[c]{@{}l@{}}Including modified network structures (see Eq.~\eqref{eq_deeponet_structure}),\\ input transformation ($\bm{x} \mapsto \left(\bm{x},\sin(\bm{x}),\cos(\bm{x}),\dots\right)$),\\ POD-DeepONet (see Eq.~\eqref{eq_pod_deeponet}), \\ and output transformation (see Eq.~\eqref{eq_deeponet_output} and Eq.~\eqref{eq_deeponet_dirichlet}).\end{tabular} \\ \cline{2-3} 
 &
  \begin{tabular}[c]{@{}l@{}}Multiple-input DeepONet \\ \cite{jin2022mionet}\end{tabular} &
  \begin{tabular}[c]{@{}l@{}}A variant of DeepONet taking multiple \\ various parameters as input,\\ i.e., $\tilde{G}\colon \Theta_1\times \Theta_2 \times \cdots \times \Theta_n \rightarrow Y$.\end{tabular} \\ \cline{2-3} 
 &
  \begin{tabular}[c]{@{}l@{}}Pre-trained DeepONet for\\ Multi-physics \cite{mao2021deepm,cai2021deepm}\end{tabular} &
  \begin{tabular}[c]{@{}l@{}}Model a multi-physics system with several \\ pre-trained DeepONets serving as building blocks.\end{tabular} \\ \cline{2-3} 
 &
  Other Variants &
  \begin{tabular}[c]{@{}l@{}}Including Bayesian DeepONet \cite{lin2021accelerated}, \\ multi-fidelity DeepONet \cite{howard2022multifidelity}, \\ and MultiAuto-DeepONet \cite{zhang2022multiauto}.\end{tabular} \\ \hline
\multirow{2}{*}{\begin{tabular}[c]{@{}l@{}}\textbf{Green's Function Learning}\\ \\ $G_w(\theta)(\bm{x}) = \int_{\Omega} \mathcal{G}(\bm{x},\bm{y}) \theta(\bm{y}) \mathd \bm{y} + u_{\mathrm{homo}}(\bm{x})$,\\ where $\theta$ is a function $\theta = v(\bm{x})$\end{tabular}} &
  \begin{tabular}[c]{@{}l@{}}Methods for \\ Linear Operators \cite{zhang2021mod,boulle2022data}\end{tabular} &
  \begin{tabular}[c]{@{}l@{}}Parameterize $\mathcal{G}$ and $u_{\mathrm{homo}}$ with neural networks,\\ which are trained with supervised data \\ (and possibly physics-informed losses).\end{tabular} \\ \cline{2-3} 
 &
  \begin{tabular}[c]{@{}l@{}}Methods for \\ Nonlinear Operators \cite{gin2021deepgreen}\end{tabular} &
  \begin{tabular}[c]{@{}l@{}}Discretize the PDEs and use trainable mappings \\ to linearize the target operator,\\ where Green's function formula is subsequently \\ applied to construct the approximation.\end{tabular} \\ \hline
\multirow{3}{*}{\begin{tabular}[c]{@{}l@{}}\textbf{Grid-based Operator Learning}\\ \\ $G_w(\theta) = \{ u(\bm{x}_i) \}_{i=1}^N$, where $ \{ u(\bm{x}_i) \}_{i=1}^N$ and \\ $\theta = \{v(\bm{x}_i)\}_{i=1}^N$ are discretizations of input and \\ output functions in some \textbf{grids}\end{tabular}} &
  \begin{tabular}[c]{@{}l@{}}Convolutional \\ Neural Network \cite{gao2021phygeonet,khara2021field}\end{tabular} &
  \begin{tabular}[c]{@{}l@{}}A convolutional neural network is utilized to \\ approximate such an image-to-image mapping,\\ where the loss function is based on supervised \\ data (and possibly physics-informed losses).\end{tabular} \\ \cline{2-3} 
 &
  \begin{tabular}[c]{@{}l@{}}Fourier Neural Operator \\ \cite{li2020fourier} \end{tabular} &
  \begin{tabular}[c]{@{}l@{}}Several Fourier convolutional kernels are \\ incorporated into the network structure,\\ to better learn the features in the frequency domain.\end{tabular} \\ \cline{2-3} 
 &
  \begin{tabular}[c]{@{}l@{}}Neural Operator with \\ Attention Mechanism\\ \cite{cao2021choose,li2022transformer,kissas2022learning}\end{tabular} &
  \begin{tabular}[c]{@{}l@{}}The attention mechanism is introduced to \\ the design of the network structure,\\ to improve the abstraction ability of the model.\end{tabular} \\ \hline
\multirow{3}{*}{\begin{tabular}[c]{@{}l@{}}\textbf{Graph-based Operator Learning}\\ \\ $G_w(\theta) = \{ u(\bm{x}_i) \}_{i=1}^N$, where $ \{ u(\bm{x}_i) \}_{i=1}^N$ and \\ $\theta = \{v(\bm{x}_i)\}_{i=1}^N$ are discretizations of input and \\ output functions in some \textbf{graphs}\end{tabular}} &
  \begin{tabular}[c]{@{}l@{}}Graph Kernel Network \cite{li2020neural}\end{tabular} &
  \begin{tabular}[c]{@{}l@{}}A graph kernel network is employed to \\ learn such a graph-based mapping.\end{tabular} \\ \cline{2-3} 
 &
  \begin{tabular}[c]{@{}l@{}}Multipole Graph \\ Neural Operator \cite{li2020multipole}\end{tabular} &
  \begin{tabular}[c]{@{}l@{}}The graph kernel is decomposed into \\ several multi-level sub-kernels, \\ to capture multi-level neighboring interactions.\end{tabular} \\ \cline{2-3} 
 &
  \begin{tabular}[c]{@{}l@{}}Graph Neural Opeartor with \\ Autogressive Methods \cite{brandstetter2022message}\end{tabular} &
  Extend graph neural operators to time-dependent PDEs. \\ \hline
\end{tabular}
\caption{A brief summary of the methods in the neural operator.}
\label{tab:summary_no}
\end{table*}

\subsection{Neural Operator}
\label{sec_neural_operator}

In this section, we will first formally give the goal of neural operator, followed by revisiting several important methods (and their variants) in this field. These methods can be broadly classified into four categories, including the direct methods represented by the DeepONet~\cite{lu2021learning}, Green's function learning, grid-based operator learning (which is similar to approximating image-to-image mappings), and graph-based operator learning. A brief summary is provided in Table~\ref{tab:summary_no}. Finally, in Section~\ref{sec_neural_operator_challenge}, we mark the open challenges and future work in this field.

\subsubsection{Problem Formulation}
\label{sec_formulation_neural_operator}

The goal of a neural operator is to approximate a latent operator, that is, a mapping between the (vector of) parameters and the state variables (with neural networks). A neural operator solves a class of differential equations that map given parameters or
control functions $\theta \in \Theta$ to its solutions (i.e., state variables). The
physical laws in Equation~\eqref{pde1} are (partially) known and we might have a
dataset of $\mathcal{D}= \{ \tilde{G} (\theta_i) (\tmmathbf{x}_j) \}_{1 \leqslant i
\leqslant N_1, 1 \leqslant j \leqslant N_2}$ of different parameter $\theta$
and collocation points $\tmmathbf{x}$, where $\tilde{G}$ is the latent operator. Mathematically, the goal can be formalized as
\begin{equation}
  \min_{w \in W}  \| G_w (\theta) (\tmmathbf{x}) - \tilde{G} (\theta)
  (\tmmathbf{x}) \|,
\end{equation}
where $G : \Theta \times \Omega \rightarrow \mathbb{R}^m$ is the neural operator (a neural network with weights $w$) and
$\tilde{G}$ is the ground truth. 

For any $\theta \in \Theta$,
$\tilde{G} (\theta) (\cdummy)$ is the solution to the governing equations in
Equation~\eqref{pde1}. The difference between a neural solver and a neural operator
is that the goal of the neural operator is to learn a surrogate model representing ODEs/PDEs for all $\theta \in \Theta$ rather than just solving an instance of the physical system. The appearance of neural operators might revolutionize the surrogate modeling that is widely used in science and engineering. The key advantage of a neural operator is the strong generalization ability and large model capacity. Similar to pre-trained models in CV and NLP, a large pre-trained neural operator could be employed as a surrogate for the expensive traditional
ODEs/PDEs' solver or digital twins for a real physical system. From the perspective of algorithms, designing specialized model architectures and pre-training methods for neural operators are important open problems for physics-informed machine learning. From the perspective of applications, finding important application scenarios or downstream tasks for neural operators will be a challenge that requires interdisciplinary collaboration.


\subsubsection{Direct Methods}
\label{sec_operator_deeponet}


Based on the Universal Approximation Theorem of Operators \cite{chen1995universal}, the direct methods parameterize the mapping by a neural network which takes both the parameters $\theta$ and the coordinates $\tmmathbf{x}$ as its inputs. DeepONet \cite{lu2021learning} is one of the most famous representatives. In the following text, we will briefly introduce this method and its relevant variants. 

\textbf{DeepONets.} We now present the architecture of DeepONets as follows,
\begin{equation}\label{eq_deeponet}
    G_w (\theta) (\tmmathbf{x})=b_0 + \sum_{k=1}^p b_k(\theta) t_k(\tmmathbf{x}),
\end{equation}
where $G$ is the neural operator instantiated as DeepONet with learnable parameters $w$, $b_0\in \mathbb{R}$ is a learnable bias, and the branch network $(b_1,\dots,b_p)$ as well as the trunk network $(t_1,\dots,t_p)$ are two neural networks. The networks take the parameters $\theta$ and the coordinates $\tmmathbf{x}$ as the inputs, and output a vector of width $p$. We note that DeepONet does not specify the architectures of the branch and trunk networks, which can be FNNs, ResNets, or other architectures. Besides, if $\theta$ is an infinite-dimensional vector (e.g., a function in an abstract Hilbert space), we may need to represent it with another finite-dimensional vector since the width of the neural layer cannot be infinite. For instance, if $\theta$ is a function $f(x), x\in\mathbb{R}$, we may represent it in terms of the first $n$ Fourier coefficients of $f$ or the values at a set of given points $[f(x_1), \dots,f(x_n)]^\top$.

Given the dataset $\mathcal{D}= \{ \tilde{G} (\theta_i) (\tmmathbf{x}_j) \}_{1 \leqslant i
\leqslant N_1, 1 \leqslant j \leqslant N_2}$, we can train the DeepONet with the following supervised loss function:
\begin{equation}\label{eq_operator_loss}
    \mathcal{L} = \frac{1}{N_1N_2}\sum_{i=1}^{N_1} \sum_{j=1}^{N_2} \big\| G_w (\theta_i) (\tmmathbf{x}_j) - \tilde{G}(\theta_i)(\tmmathbf{x}_j) \big\|^2.
\end{equation}
Hereinafter, we refer to Equation~\eqref{eq_operator_loss} as the operator loss function $\mathcal{L}_{\text{operator}}$. It is worth noting that, in Equation~\eqref{eq_deeponet} the output of DeepONet is a scalar; however, the solution to the PDEs (i.e., $\tilde{G}(\theta)(\tmmathbf{x})$) can be a vector of dimension higher than 1. To bridge this gap, we can split the outputs of the branch and trunk networks and apply Equation~\eqref{eq_deeponet} to each group separately to generate the components of the resulting vector. For example, supposing that $\tilde{G}(\theta)(\tmmathbf{x})$ is a 2-dimensional vector and $p=100$; then, we can use the following ansatz,
\begin{equation}
\begin{split}
    G_w (\theta) (\tmmathbf{x})=\tmmathbf{b}_0 + \bigg[\sum_{k=1}^{50} b_k(\theta) t_k(\tmmathbf{x}),
    \sum_{k=51}^{100} b_k(\theta) t_k(\tmmathbf{x}), \bigg]^\top,
\end{split}
\end{equation}
where the bias $\tmmathbf{b}_0\in \mathbb{R}^2$ also becomes a 2-dimensional vector. More discussion on such multiple-output DeepONets can be found in the paper \cite{lu2022comprehensive}. 

DeepONet is a simple but effective neural operator. With the help of the generalizability of neural networks, DeepONet is able to learn latent operators from the data, not just the solution to a certain instance of the PDEs. In this way, for any instance in a class of parameterized PDEs, only a single forward pass of DeepONet is needed to obtain its solution. This is something that neural solvers like PINNs cannot do. Unlike numerical methods such as FEM, both the input and output of DeepONet are mesh-independent and thus it is more flexible and less sensitive to the increase in dimensionality. However, DeepONet still has some limitations. For example, since DeepONet is purely data-driven, the requirement of the training data is usually relatively large, especially for some complex PDEs. The generation of these data (by numerical simulations or experiments) is often expensive, thus greatly limiting its application. Therefore, some variants of DeepONet have been proposed to solve these problems, which are described below.

\textbf{Physics-informed DeepONets.}
As mentioned above, a severe problem with DeepONet is that training data acquisition is sometimes too expensive. A straightforward way to overcome this defect is to incorporate the idea of PINNs \cite{raissi2019physics} and add the physics-informed loss $\mathcal{L}_{\mathrm{physics}}$ into the loss function. Such a method \cite{wang2021learning} can reduce the data requirements of DeepONet (for some simple PDEs, the labeled data are not even needed) since the physics-informed loss does not require any labeled data, and only some points in the domain $\Omega$ is necessary (that is to say, we only need to sample $\{\theta_i\}$ and $\{\tmmathbf{x}_j\}$ and do not have to evaluate $\{\tilde{G} (\theta_i) (\tmmathbf{x}_j)\}$). The loss function of this method can be expressed as
\begin{equation}
    \mathcal{L} = \mathcal{L}_{\mathrm{operator}} + \mathcal{L}_{\mathrm{physics}},
\label{eq_pideeponet}
\end{equation}
where $\mathcal{L}_{\mathrm{operator}}$ is defined in Equation~\eqref{eq_operator_loss} and $\mathcal{L}_{\mathrm{physics}}$ is given by,
\begin{equation}\label{eq_physics_loss_in_pideeponet}
\begin{split}
  \mathcal{L}_{\mathrm{physics}}= \frac{1}{N_1}\sum_{i=1}^{N_1}\bigg( \frac{\lambda_r}{N_r} \sum_{j = 1}^{N_r} \| \mathcal{F} (u_w ;
  \theta_i) (\tmmathbf{x}_j) \|^2 \\
  + \frac{\lambda_i}{N_i} \sum_{j = 1}^{N_i} \|
  \mathcal{I} (u_w ; \theta_i) (\tmmathbf{x}_j) \|^2  + \frac{\lambda_b}{N_b}
  \sum_{j = 1}^{N_b} \| \mathcal{B} (u_w ; \theta_i) (\tmmathbf{x}_j) \|^2\bigg),
\end{split}
\end{equation}
where $u_w=G_w(\theta_i)$ is the approximation of the solution to PDEs under parameters $\theta_i$, $N_r$, $N_i$, $N_b$ are, respectively, the number of points sampled inside the domain $\Omega$, at the initial time, and on the boundary, and $\lambda_r$, $\lambda_i$, $\lambda_b$ are the corresponding weights of losses. Recalling the loss function of PINNs in Equation~\eqref{pinn2}, we find that Equation~\eqref{eq_physics_loss_in_pideeponet} is similar, except for additional parameterization of $\theta_i$ and the absence of the fourth term (i.e., the regular data loss whose role has been fulfilled by $\mathcal{L}_{\mathrm{operator}}$ here).

Physics-informed DeepONet directly combines the approaches of PINNs and DeepONet, which alleviates the problems of both DeepONet's large data demand and PINNs' poor approximation of the solution to complex PDEs. This idea has been applied to solve other parametric systems in addition to parametric PDEs, such as a specific class of eigenvalue problems \cite{kovacs2022conditional}. In addition, many variants based on physics-informed DeepONet have been proposed, such as the variant for long-term simulation of dynamic systems \cite{wang2021long}. While physics-informed DeepONet has the advantages of physics-informed and data-driven learning, it is much too simple to combine the two learning methods by merging loss functions, where the weights of the losses can be difficult to specify. Similar to PINNs, some loss re-weighting and data re-sampling algorithms have been proposed to address this difficulty \cite{wang2022improved,kontolati2022influence}. Future work includes finding a more efficient way to combine physical prior knowledge with available data.



\textbf{Improved Architectures for DeepONets.}
The architecture plays an important role in deep learning, including operator learning. Recently, researchers have proposed several improved architectures for DeepONets. These can be divided into two types: novel network structures and pre-processing \& post-processing techniques.

Wang \textit{et. al.} \cite{wang2022improved} propose a novel modified structure for DeepONets, which incorporates encoders. The forward pass of the modified structure can be formulated as:
\begin{equation}\label{eq_deeponet_structure}
\begin{aligned}
&\bm{U}= \phi(\bm{W}_{\theta} \theta + \bm{b}_{\theta}), \bm{V}=\phi(\bm{W}_{x} \bm{x} + \bm{b}_{x}),\\
&\bm{H}^{(1)}_{\theta} = \phi(\bm{W}_{\theta}^{(1)} \theta + \bm{b}_{\theta}^{(1)}), \bm{H}^{(1)}_{x} = \phi(\bm{W}_{x}^{(1)} \bm{x} + \bm{b}_{x}^{(1)}),\\
&\bm{Z}^{(l)}_{\theta} = \phi(\bm{W}_{\theta}^{(l)} \bm{H}_{\theta}^{(l)} + \bm{b}_{\theta}^{(l)}), \bm{Z}^{(l)}_{x} = \phi(\bm{W}_{x}^{(l)} \bm{H}_{x}^{(l)} + \bm{b}_{x}^{(l)}),\\
&\bm{H}_{\theta}^{(l+1)}=(1-\bm{Z}_{\theta}^{(l)})\odot \bm{U} + \bm{Z}_{\theta}^{(l)}\odot \bm{V},\\
&\bm{H}_{x}^{(l+1)}=(1-\bm{Z}_{x}^{(l)})\odot \bm{U} + \bm{Z}_{x}^{(l)}\odot \bm{V},\\
&\bm{H}_{\theta}^{(L)}=\phi(\bm{W}_{\theta}^{(L)} \bm{H}_{\theta}^{(L-1)} + \bm{b}^{(L)}_{\theta}),\\ &\bm{H}_{x}^{(L)}=\phi(\bm{W}_{x}^{(L)} \bm{H}_{x}^{(L-1)} + \bm{b}^{(L)}_{x}),\\
&G_w(\theta)(\bm{x})=\left\langle \bm{H}_{\theta}^{(L)}, \bm{H}_{x}^{(L)} \right\rangle,
\end{aligned}
\end{equation}
where $l=1,\dots,L-1$, $\phi$ denotes a given activation function; $\odot$ denotes the element-wise multiplication; $\{ \bm{W}^{(i)}_{\theta}, \bm{b}^{(i)}_{\theta} \}_{i=1}^{L+1}$ and $\{ \bm{W}^{(i)}_{x}, \bm{b}^{(i)}_{x} \}_{i=1}^{L+1}$ are, respectively, the learnable weights and biases of the branch and trunk networks; and $\langle \cdummy \rangle$ denotes the inner product. The modified structure utilizes two encoders to embed the inputs $\theta$ and $\bm{x}$ into two feature vectors $U$ and $V$, respectively, in high-dimensional latent spaces. $U$ and $V$ are then added element-wise and fed into the hidden layers $\bm{H}_{\theta}^{(l)}$ and $\bm{H}_{x}^{(l)}$. In contrast to vanilla DeepONet (see Equation~\eqref{eq_deeponet}), the messages of $\theta$ and $\bm{x}$ are merged before going through each hidden layer instead of merging them just before output, which improves the ability to abstract nonlinear features. In addition to the structure introduced above, other studies have considered the Auto-Decoder structure \cite{huang2021meta}. 

Pre-processing refers to techniques that process the input before putting it into the neural network. Feature expansion \cite{lu2022comprehensive} is one of the frequently used techniques. For example, for the PDEs with oscillating solutions, a harmonic feature expansion \cite{di2021deeponet} can be applied to the input $\bm{x}$ before entering into the trunk network,
\begin{equation}
    \bm{x} \mapsto \left(\bm{x},  \sin(\bm{x}),  \cos(\bm{x}), \sin(2\bm{x}), \cos(2\bm{x}),\dots  \right),
\end{equation}
where we assume $\bm{x}$ is a 1-dimensional vector. With carefully designed feature expansion or feature mapping, we can pass more valuable information to the neural network, allowing the model to better approximate the underlying operator. Similarly, we can also pre-compute the proper orthogonal decomposition (POD) modes of the state variables $u(\bm{x})$ from the dataset (after zero-mean normalization) and replace the trunk network with them (POD-DeepONet \cite{lu2022comprehensive}),
\begin{equation}\label{eq_pod_deeponet}
    G_w (\theta) (\tmmathbf{x})=\phi_0(\bm{x}) + \sum_{k=1}^p b_k(\theta) \phi_k(\tmmathbf{x}),
\end{equation}
where $\phi_0(\bm{x})$ is the mean function of $u(\bm{x})$, i.e, $\phi_0(\bm{x})=\mathbb{E}_{\theta} [\tilde{G}(\theta)(\bm{x})]$, $\{\phi_1(\bm{x}),\dots,\phi_p(\bm{x})\}$ are the POD modes of $u(\bm{x})$.

In contrast, post-processing refers to techniques that process the output of the neural network before generating the approximation. For example, to stabilize the training process, we may rescale the output of the DeepONet to achieve a unit variance,
\begin{equation}\label{eq_deeponet_output}
    G_w (\theta) (\tmmathbf{x})=\frac{1}{\sqrt{\mathrm{Var}[\sum_{k=1}^p b_k(\theta) t_k(\tmmathbf{x})}]}\left[b_0 + \sum_{k=1}^p b_k(\theta) t_k(\tmmathbf{x}) \right],
\end{equation}
where the variance $\mathrm{Var}[\sum_{k=1}^p b_k(\theta) t_k(\tmmathbf{x})]$ depends on the specific initialization methods of the neural network. We refer to the paper \cite{lu2022comprehensive} for a detailed discussion. Another example is the hard-constraint boundary conditions which have drawn much attention in PINNs (see Section~\ref{sec_pinn_arch}, \textbf{Multiple NNs and Boundary Encoding.}). We consider the following Dirichlet BC,
\begin{equation}\label{eq_deeponet_dirichlet}
    \mathcal{B}(u;\theta)(\bm{x})\triangleq u(\bm{x}) = g(\bm{x}), x \in \partial\Omega,
\end{equation}
where we note again that $\bm{x} = (x, t)$, $x$ and $t$ are the spatial and temporal coordinates, respectively. To enforce the above BC, we can construct our anstaz as follows,
\begin{equation}
    G_w (\theta) (\tmmathbf{x}) = g(\bm{x}) + l(x) \mathcal{N}(\theta)(\bm{x}),
\end{equation}
where $\mathcal{N}(\theta)(\bm{x})$ is the output of the original DeepONet (see Equation~\eqref{eq_deeponet}), and $l(x)$ is a smooth distance function which satisfies:
\begin{equation}
\begin{cases}
l(x)=0& \text{if}\ x\in\partial\Omega,\\
l(x)>0& \text{otherwise}.
\end{cases}
\end{equation}
We note that, if $g(\bm{x})$ is not defined in total $\Omega$, we may have to extend its definition smoothly. As for other types of BCs such as Periodic BCs \cite{dong2021method}, preprocessing techniques may be also utilized to enforce the BCs.

\cite{mao2020physics}
\textbf{Multiple-input DeepONets.}
As we discussed above, the DeepONet is designed for the operator whose input space $\Theta$ is a single Banach space, that is, the input $\theta$ can only be a vector (or a function) but not multiple vectors (or a vector-value function) which are defined on different spaces. To extend DeepONet to learn a operator with the input of multiple vectors (i.e., a multiple-input operator), a new theory of universal approximation needs to be proven and a new network architecture needs to be designed. The paper \cite{jin2022mionet} gives the answers to both of them. The authors first prove the theory of universal approximation of neural networks for a multiple-input operator, which can be described as,
\begin{equation}\label{eq_multiple_input_operator}
    \tilde{G}\colon \Theta_1\times \Theta_2 \times \cdots \times \Theta_n \rightarrow Y,
\end{equation}
where $ \Theta_1,\dots,\Theta_n $ are $n$ (different) input spaces, and $Y$ is the target space. In the context of the neural operator (see Section~\ref{sec_formulation_neural_operator}), we have $Y=\Omega \rightarrow \mathbb{R}^m$ and Equation~\eqref{eq_multiple_input_operator} can be rewritten as,
\begin{equation}
    \tilde{G}\colon \Theta_1\times \Theta_2 \times \cdots \times \Theta_n\times \Omega \rightarrow \mathbb{R}^m.
\end{equation}

Motivated by the newly proven theory, the authors propose the extension of DeepONet, MIONet, for the multiple-input operator,
\begin{equation}
    G_w(\tmmathbf{\theta})(\tmmathbf{x}) = b_0 + \sum_{k=1}^p b_k^1(\theta_1) \cdots b_k^n(\theta_n) t_k(\tmmathbf{x}),
\end{equation}
where $\theta_i \in \Theta_i$, $i=1,\dots,n$ are $n$ input vectors, $\tmmathbf{\theta}=[\theta_1,\dots,\theta_n]^\top$, and $(b^i_1,\dots,b^i_p)$, $i=1,\dots,n$ are $n$ independent branch networks.

Multiple-input DeepONet (MIONet) is very useful when we want to learn a latent operator which takes more than two functions as its inputs. For example, we consider the following ODE system, 
\begin{eqnarray}
  \frac{\mathd u_1}{\mathd t} &=& u_2(t),\\
  \frac{\mathd u_2}{\mathd t} &=& -f_1(t)\sin(u_1(t)) +f_2(t),
\end{eqnarray}
where $t\in(0,1]$ and the initial condition is $u_1(0)=u_2(0)=0$. Our operator of interest is given by,
\begin{equation}
    \tilde{G}\colon (f_1,f_2)\mapsto u_1.
\end{equation}
Here, the solution $u_1$ depends on both $f_1$ and $f_2$, conforming to the definition of the multiple-input operator. We can employ the MIONet to learn such operators. 

\textbf{Pre-trained DeepONets for multi-physics.}

We note that DeepONets learn a mapping between functions, which can be used as pre-trained models for fast inference. As proposed in \cite{mao2021deepm} and \cite{cai2021deepm} (DeepM\&Mnets), multiple pre-trained DeepONets are used as building blocks for multi-physical systems where there are multiple state variables and PDEs. We consider the following example of electroconvection,
\begin{eqnarray}
\frac{\partial \bm{u}}{\partial t} &=& -\nabla p + \nabla^2 \bm{u} + \bm{f},\\
\nabla \cdot \bm{u} &=& 0,\\
-2\epsilon^2 \nabla^2 \phi &=& c^+ - c^-,\\
\frac{\partial c^{\pm}}{\partial t} &=& -\nabla\cdot (c^{\pm}\bm{u} - \nabla c^{\pm} \mp c^{\pm} \nabla \phi),
\end{eqnarray}
where $\bm{u}(\bm{x})$ is the velocity, $p$ is the pressure, $\phi(\bm{x})$ is the electric potential, $c^+(\bm{x})$ and $c^-(\bm{x})$ are, respectively, the cation and anion concentrations, $\bm{f}(\bm{x})$ is the electrostatic body force, and $\epsilon$ is the Debye length. 

We first define two classes of operators. The first class is to map the electric potential to other state variables:
\begin{equation}
    G_{\diamond}\colon \phi \mapsto \diamond, \diamond = \bm{u}, p, c^{\pm}.
\end{equation}
The second class is to map the concentrations to the electric potential:
\begin{equation}
    G_{\phi}\colon (c^+, c^-) \mapsto \phi.
\end{equation}
Without loss of clarity, we confuse the notations of the ground truth operator and the corresponding neural operator (i.e., DeepONet). Then, we can pretrain DeepONets for those operators with proper datasets. For example, we can train $G_{\bm{u}}$ on the dataset $\{ G_{\bm{u}}(\phi_i)(\bm{x}_j) \}_{1 \leqslant i \leqslant N_1, 1 \leqslant j \leqslant N_2}$, where $\{ \phi_i \}_{i=1}^{N_1}$ are sampled in the Hilbert space according to a certain distribution. After pretraining all the DeepONets, we finally train our ansatz, a neural network $\mathcal{N}\colon \bm{x} \mapsto (\hat{\bm{u}}, \hat{p}, \hat{c}^{\pm}, \hat{\phi})$ with the dataset $\{ (\bm{u}(\bm{x}_i), p(\bm{x}_i), c^{\pm}(\bm{x}_i), \phi(\bm{x}_i)) \}_{i=1}^{N_d}$. The loss function is given by 
\begin{eqnarray}
\mathcal{L} &=& \lambda_{\mathrm{data}} \mathcal{L}_{\mathrm{data}} + \lambda_{\mathrm{op}} (\mathcal{L}_{\mathrm{op1}} + \lambda_{\mathrm{op2}}),\\
\mathcal{L}_{\mathrm{data}} &=&  \sum_{\diamond \in \{ \bm{u}, p, c^{\pm}, \phi \}} \frac{1}{N_d}\sum_{i=1}^{N_d} \left(  \hat{\diamond}(\bm{x}_i) - \diamond(\bm{x}_i)  \right)^2,\\
\mathcal{L}_{\mathrm{op1}} &=&  \sum_{\diamond \in \{ \bm{u}, p, c^{\pm} \}} \frac{1}{N_{op}}\sum_{i=1}^{N_{op}} \left(  \hat{\diamond}(\bm{x}_i) - G_{\diamond}(\hat{\phi})(\bm{x}_i)  \right)^2,\\
\mathcal{L}_{\mathrm{op2}} &=&  \frac{1}{N_{op}}\sum_{i=1}^{N_{op}} \left(  \hat{\phi}(\bm{x}_i) - G_{\phi}(\hat{c}^+, \hat{c}^-)(\bm{x}_i)  \right)^2,
\end{eqnarray}
where $\mathcal{L}_{\mathrm{data}}$ measures the data mismatch and $\mathcal{L}_{\mathrm{op1}}, \mathcal{L}_{\mathrm{op2}}$ measure the deviation between the neural network and the pre-trained DeepONets. We note that the pre-trained DeepONets stay fixed during the training process. Moreover, if we need to estimate only some of the state variables, we can make the other state variables the hidden outputs. We refer to \cite{mao2021deepm} and \cite{cai2021deepm} for relevant details.

DeepM\&Mnets are very suitable for performing "physically meaningful" interpolation on discrete observation data points, where the physical priors are embedded in pre-trained DeepONets. Since we have decoupled the training of DeepONets and that of the ansatz, incorporating the estimations of DeepONets into the loss function does not bring too much computational overhead to the training of the latter. However, the premise is that we need to train these DeepONets on another dataset in advance, where the acquisition of the dataset and the training of DeepONets are often very time-consuming. This is also a fundamental drawback of applying DeepONets as building blocks to other neural models.


\textbf{Other Variants.}
Besides the important variants of DeepONet introduced above, many other variants have been proposed to solve problems in different domains, such as Bayesian DeepONets for training with noise and uncertainty estimation \cite{lin2021accelerated}, multi-fidelity DeepONets for training with multi-fidelity data \cite{howard2022multifidelity,lu2022multifidelity}, and MultiAuto-DeepONets for high-dimensional stochastic problems and multi-resolution inputs \cite{zhang2022multiauto}.

\subsubsection{Green's Function Learning}
In this subsection, we first review the concept of Green's function and the goal of Green's function learning. Then we introduce the methods of Green's function learning for linear and nonlinear operators, respectively.

\textbf{Green's function.} The method of Green's function is a basic approach for manually solving a class of linear PDEs such as Poisson's equation. It can be described as
\begin{eqnarray}\label{eq_linear_pde}
\mathcal{F}_L (u) &=& f, \bm{x}\in \Omega,\label{eq_linear_pde_1}\\
\mathcal{B}_L (u) &=& g, \bm{x} \in \partial\Omega,\label{eq_linear_pde_2}
\end{eqnarray}
where $\mathcal{F}_L$ and $\mathcal{B}_L$ are two linear operators, $f(\bm{x})$ is the forcing term, and $g(\bm{x})$ is the constraint function on the boundary $\partial \Omega$. Green's function $\mathcal{G}(\bm{x},\bm{y})$ corresponding to the above boundary value problem is implicitly defined as follows,
\begin{eqnarray}
    \mathcal{F}_L \left(\mathcal{G}(\bm{x},\bm{y}) \right) &=& \delta(\bm{y} - \bm{x}), \bm{x},\bm{y}\in \Omega,\\
    \mathcal{B}_L \left(\mathcal{G}(\bm{x},\bm{y}) \right) &=& 0, \bm{x}\in \partial\Omega,
\end{eqnarray}
where the inputs of $\mathcal{F}_L$ and $\mathcal{B}_L$ are both the function $\bm{x}\mapsto \mathcal{G}(\bm{x}, \bm{y})$ for fixed $\bm{y}$ and $\delta(\cdummy)$ is the Dirac delta function. From the superposition principle, we can construct the solution to the boundary value problem defined by Equation~\eqref{eq_linear_pde_1} and \eqref{eq_linear_pde_2} as
\begin{equation}\label{eq_green_function}
    u(\bm{x}) = \int_{\Omega} \mathcal{G}(\bm{x},\bm{y}) f(\bm{y}) \mathd \bm{y} + u_{\mathrm{homo}}(\bm{x}),
\end{equation}
where $u_{\mathrm{homo}}$ is the homogeneous solution which satisfies $\mathcal{F}_L (u_{\mathrm{homo}}(\bm{x})) = 0,\bm{x} \in \Omega$ and $\mathcal{B}_L (u_{\mathrm{homo}}(\bm{x})) = g,\bm{x} \in \partial\Omega$. 

However, for complicated PDEs, the analytical expression of Green's function $\mathcal{G}(\bm{x},\bm{y})$ may be hard to solve. To tackle this challenge, we may approximate Green's function $\mathcal{G}(\bm{x},\bm{y})$ with neural networks, which is the original intention of Green's function learning. To be formal, Green's function learning hopes to learn an operator $\tilde{G}\colon f\mapsto u$ from the forcing term $f$ to the solution $u$ using the structure of the Green's function (see Equation~\eqref{eq_green_function}). Green's function learning can be considered as a subclass of the neural operators, where the parameter $\theta$ is restricted to be the forcing term $f$. Several methods have been proposed for Green's function learning, which will be presented in the following.

\textbf{Green's function learning for linear operators.}

As for linear $\mathcal{F}_L$ and $\mathcal{B}_L$, we can directly utilize the format given in Equation~\eqref{eq_green_function} to construct our ansatz as proposed in \cite{zhang2021mod} and \cite{boulle2022data}. Specifically, we parameterize the Green's function $\mathcal{G}(\bm{x},\bm{y})$ and the homogeneous solution $u_{\mathrm{homo}}$ with two neural networks which are trained in a supervised learning manner on a dataset $\mathcal{D}= \{ \tilde{G} (f_i) (\tmmathbf{x}_j) \}_{1 \leqslant i \leqslant N_1, 1 \leqslant j \leqslant N_2}$. In addition, physics-informed losses corresponding to the PDEs in Equation~\eqref{eq_linear_pde_1} and \eqref{eq_linear_pde_2} can be incorporated in the loss function \cite{zhang2021mod}.

Compared with other neural operator methods, Green's function learning has the following advantages. First, the structure of Green's function (see Equation~\eqref{eq_green_function}) contains more priors about the physical system, which makes the training of neural networks more data-efficient. Second, it is mathematically easier to approximate Green's function than the latent operator $\tilde{G}$ \cite{boulle2022data}. Third, the structure of Green's function can be employed flexibly, since many physical or mathematical properties (such as the symmetry of Green's function) can be encoded into the network architecture to improve accuracy. However, Green's function learning also has some limitations. On the one hand, such methods are limited to a special class of PDEs as in Equation~\eqref{eq_linear_pde_1} and \eqref{eq_linear_pde_2}. On the other hand, the input dimension of Green's function is twice the spatial dimension, which makes it infeasible to apply many grid-based methods.

\textbf{Green's function learning for nonlinear operators.}

We recall that the format of Green's function given in Equation~\eqref{eq_green_function} is only available for linear operators which satisfy the superposition principle. Necessary processing techniques must be undertaken when handing nonlinear boundary value problems of the form
\begin{equation}
    \mathcal{F}_N (u) = f, \bm{x}\in \Omega,
\end{equation}
where $\mathcal{F}_N$ is a nonlinear operator and a linear boundary condition is imposed as in Equation~\eqref{eq_linear_pde_2}. For example, in DeepGreen \cite{gin2021deepgreen} we first discretize the boundary value problem to obtain that
\begin{equation}
    \bm{F}_N[ \bm{u} ] = \bm{f},
\end{equation}
where $\bm{F}_N$, $\bm{u}$, $\bm{f}$ are spatial discretizations of $\mathcal{F}_N$, $u$, $f$, respectively. Then we use two mappings, $\bm{\psi}$ and $\bm{\phi}$, which are parameterized by autoencoder networks, to transform $\bm{u}$ and $\bm{f}$,
\begin{eqnarray}
\bm{v} &=& \bm{\psi}(\bm{u}),\\
\bm{h} &=& \bm{\phi}(\bm{f}),
\end{eqnarray}
where $\bm{v}$ and $\bm{h}$ satisfy the following:
\begin{equation}
    \bm{F}'_L[ \bm{v} ] = \bm{h},
\end{equation}
for some linear operator $\bm{F}'_L$ in the latent space. Finally, we can apply the structure of Green's function to this linearized boundary value problem. It is noted that we learn the two mappings, $\bm{\psi}$ and $\bm{\phi}$, from a dataset $\mathcal{D}=\{ (\bm{f}_i, \bm{u}_i) \}_{i=1}^N$ and do not specify $\bm{F}'_L$, whose linearity is enforced by a linear superposition loss.

DeepGreen has successfully extended Green's function learning to nonlinear boundary value problems. Nevertheless, we should point out that there is a no theoretically rigorous underpinning for the existence of $\bm{\psi}$ or for $\bm{\phi}$, and the linearity of $\bm{F}'_L$ does not strictly hold. Besides, DeepGreen is inherently a grid-based method that may be prohibitive for high-dimensional problems. 



\subsubsection{Grid-based Operator Learning}\label{sec_grid_operator_learning}
Besides the direct methods such as DeepONet, we can also formalize the latent operator as a grid-based mapping, that is,
\begin{equation}\label{eq_grid_based_operator}
    \tilde{G}\colon \theta \mapsto \{ u(\bm{x}_i) \}_{i=1}^N,
\end{equation}
where $u$ is the solution to PDEs under parameters $\theta$, and $\{ \bm{x}_i \}_{i=1}^N$ is a set of $N$ query coordinates (i.e., a grid of points) which is usually predefined. We call the methods to learn the operator of such a formalization \textit{grid-based operator learning} methods. If $\theta$ is (some representation of) a function and shares the same grid with $u$, Equation~\eqref{eq_grid_based_operator} can be equivalently rewritten as
\begin{equation}\label{eq_grid_based_operator_2}
    \tilde{G}\colon \theta = \{ v(\bm{x}_i) \}_{i=1}^N \mapsto \{ u(\bm{x}_i) \}_{i=1}^N,
\end{equation}
where $v$ is the input function. Moreover, if the points are uniformly distributed (i.e., a regular gird), we can replace the notations of $\{ v(\bm{x}_i) \}_{i=1}^N$ and $\{ u(\bm{x}_i) \}_{i=1}^N$ with tensors $\bm{X}$ and $\bm{Y}$, respectively. Therefore, such operators are also called \textit{image-to-image mappings}.

\textbf{Convolutional neural networks.}
The convolutional neural network \cite{o2015introduction} is a well-known and powerful model to learn image-to-image mapping in the field of computer vision. Here, we can also utilize convolutional architecture, such as the U-Net architecture \cite{ronneberger2015u}, to approximate an operator which can be formalized as image-to-image mapping:
\begin{equation}
    \tilde{G}(\theta) \approx G_w(\theta) = \mathrm{CNN}_w(\theta),
\end{equation}
where the output size of the convolutional neural network $\mathrm{CNN}$ is the same as the size of the regular grid. 

According to the connection between numerical differential operators and convolutional kernels (see Section~\ref{sec_pinn_arch}, \textbf{Convolutional Architectures}), physics-informed learning methods already have been developed for convolutional neural networks \cite{gao2021phygeonet,khara2021field}, which do not require any labeled data. Moreover, \cite{winovich2019convpde} applies a Bayesian framework on convolutional neural networks, facilitating pointwise uncertainty quantification.

With the help of the convolution operation, convolutional neural networks can effectively extract segmentation behind the training ``images'' and learn a low-rank representation of the physical laws. However, such architectures  suffer from the ``curse of dimensionality'' due to the dependency of the grid and hardly utilize the frequency information of the input, which is sometimes very important for the input function (we note that the input is usually a very smooth function but not a real ``image''). In addition, the architectures are only applicable for regular grids. Although there are methods which intend to map an irregular domain to a regular one (see Section~\ref{sec_pinn_arch}, \textbf{Convolutional Architectures}), they are still very inflexible for geometrically complicated PDEs, which is exactly what graph-based methods are meant to solve (we discuss graph-based methods in Section~\ref{sec_graph_operator_learning}) 


\begin{figure*}[t!]
\begin{center}
\centering
\captionsetup{justification=centering}
\includegraphics[width=.8\textwidth]{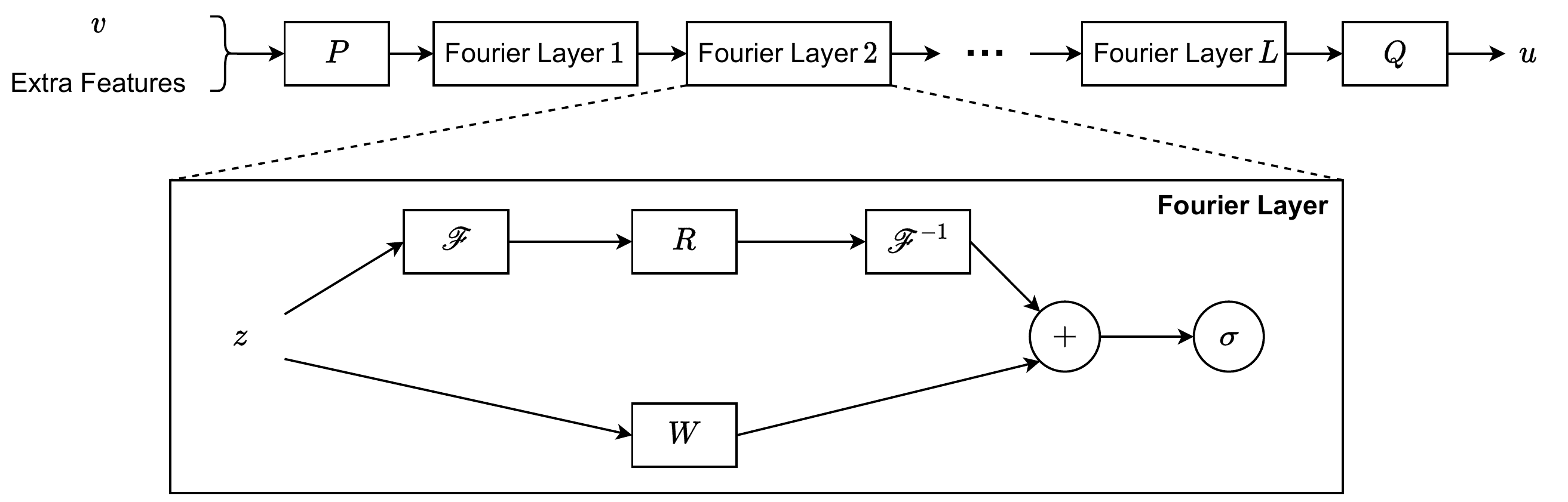}
\caption{The Architecture of FNOs.}
\label{fig_fno}
\end{center}
\end{figure*}

\textbf{Fourier neural operators.}
The Fourier neural operator (FNO) \cite{li2020fourier} is another architecture for learning image-to-image mappings, which considers the features of the input function in both spatial and frequency domains via the Fourier transformation. The architecture of FNOs is presented in Figure~\ref{fig_fno}. In the FNO, we first apply a local transformation $P\colon \mathbb{R}\rightarrow \mathbb{R}^{d_z}$ on the input function $v$ (which is represented by the function values on a regular grid $\{ \bm{x}_i \}_{i=1}^N$) as well as some extra features (if needed),
\begin{equation}
    z_0(\bm{x}_i) = P(v(\bm{x}_i))\in\mathbb{R}^{d_z}, i=1,\dots,N, 
\end{equation}
where the transformation $P$ is usually parameterized by a shallow fully-connected neural network. It is noted that the output $\{z_0 (\bm{x}_i)\}_{i=1}^N$ lives on the same grid as $v$, which can be viewed as an image with $d_z$ channels. In the next step, we iteratively apply $L$ Fourier layers on $z_0$:
\begin{equation}
    z_0(\bm{x}_i)\mapsto z_1(\bm{x}_i)\mapsto \cdots \mapsto z_L(\bm{x}_i),i=1,\dots,N,
\end{equation}
where $z_j\in \mathbb{R}^{d_z}, j=1,\dots,L$. The Fourier layer is defined as
\begin{equation}
    z_{l+1} = \sigma \left( \mathscr{F}^{-1} \left( \bm{R}_l \cdot \mathscr{F}(z_l) \right) + \bm{W}_l \cdot z_l + \bm{b}_l \right),
\end{equation}
where $l=0,\dots,L-1$, $\sigma$ is an activation function; $\bm{b}$ is a bias; $R_l$ and $W_l$ are, respectively, the weight tensors in the frequency and spatial domain; $\mathscr{F}$ is the operator representing the Fast Fourier Transformation (FFT); and $\mathscr{F}^{-1}$ is its inverse. Finally, another local transformation $Q\colon \mathbb{R}^{d_z}\rightarrow \mathbb{R}$ is used to project $z_L$ back to the domain of the output,
\begin{equation}
    u(\bm{x}_i) = Q(z_0(\bm{x}_i))\in\mathbb{R}, i=1,\dots,N. 
\end{equation}
Here, we assume the input and output functions $v$ and $u$ are scalar-value functions. We also note that FNOs can be easily extended to the scenarios of vector-value functions with multi-channel versions of $P$ and $Q$. 


Using the Fourier transformation, FNOs can effectively abstract the features of the input function in the frequency domain, which makes their experimental performance significantly better than other architectures such as U-Net. Therefore, neural operators combined with the Fourier transformations have become a paradigm, and a lot of work is devoted to improving or applying this approach. For instance, \cite{lu2022comprehensive} extends FNOs to geometrically complex cases and cases where the input and output functions are defined on different domains. \cite{grady2022towards} has designed a distributed version of FNOs for large-scale problems.
Other relevant work includes introducing the wavelet transform \cite{chun2010tutorial} into operator learning \cite{gupta2021multiwavelet,tripura2022wavelet}, improved FNO for irregular geometries \cite{li2022fourier}, and so on \cite{tran2021factorized,you2022learning,rahman2022u}. In addition, FNOs recently have been applied for weather forecasting \cite{pathak2022fourcastnet}. However, we must emphasize that FNOs are still grid-based and cannot overcome the ``curse of dimensionality''.


\textbf{Neural operators with the attention mechanism.}

The attention mechanism is a famous and powerful tool for natural language processing, computer vision, and many machine learning tasks \cite{niu2021review}. The vanilla attention mechanism can be formulated as
\begin{equation}\label{eq_att}
    \bm{z}_i = \sum_{j=1}^n \alpha_{ij}\bm{v}_j, \alpha_{ij}=\frac{\exp{\left(h(\bm{q}_i,\bm{k}_j)\right)}}{\sum_{s=1}^n \exp{\left(h(\bm{q}_i,\bm{k}_s)\right)}},
\end{equation}
where $\{\bm{q}_j\}_{j=1}^n$, $\{\bm{k}_j\}_{j=1}^n$, $\{\bm{v}_j\}_{j=1}^n$ are the query vectors, key vectors, and value vectors respectively; $\bm{z}_i$ is the output corresponding to $\bm{q}_i$; and $\alpha_{ij}$ is the attention weight. The weight function $h(\cdummy)$ is usually chosen as a scaled dot-product \cite{vaswani2017attention}. In this way, Equation~\eqref{eq_att} can be rewritten in matrix fashion:
\begin{equation}
    \bm{Z} = \mathrm{softmax}\left( \frac{\bm{Q}\bm{K}^{\top}}{\sqrt{d}} \right) \bm{V}
\end{equation}
where $\bm{q}_i,\bm{k}_i,\bm{v}_i,\bm{z}_i\in \mathbb{R}^d$ are the $i$-th row vectors of the matrices $\bm{Q},\bm{K},\bm{V},\bm{Z} \in \mathbb{R}^{n\times d}$ respectively.

In \cite{cao2021choose}, Cao \textit{et. al.} have introduced the attention mechanism into the architecture of neural operators. They employ CNN-like architectures to extract features from the input function (which is represented by its values on a regular grid, as a matrix $\bm{X}$) and incorporate the attention mechanism in hidden layers. Specifically, the matrices $\bm{Q},\bm{K},\bm{V}$ are parameterized as follows:
\begin{equation}
    \bm{Q}\triangleq \bm{y}_{\mathrm{in}} \bm{W}_Q, \bm{K}\triangleq \bm{y}_{\mathrm{in}} \bm{W}_K, \bm{V}\triangleq \bm{y}_{\mathrm{in}} \bm{W}_V,
\end{equation}
where $\bm{y}_{\mathrm{in}}\in \mathbb{R}^{n\times d}$ is the input feature embedding, and $\bm{W}_Q$, $\bm{W}_K$, $\bm{W}_V\in \mathbb{R}^{d\times d}$ are learnable weights. The attention head that maps a feature embedding $\bm{y}_{\mathrm{in}}\in \mathbb{R}^{n\times d}$ to another feature embedding $\bm{y}_{\mathrm{out}}\in \mathbb{R}^{n\times d}$ is given by
\begin{equation}
    \bm{y}_{\mathrm{out}} = \mathrm{Ln} \left(\bm{y}_{\mathrm{in}} + \mathrm{Attn}(\bm{y}_{\mathrm{in}}) + g\left( \mathrm{Ln} \left(\bm{y}_{\mathrm{in}} + \mathrm{Attn}(\bm{y}_{\mathrm{in}}) \right) \right) \right),
\end{equation}
where $\mathrm{Ln}(\cdummy)$ is the layer normalization, $g(\cdummy)$ is parameterized as a neural network, and $\mathrm{Attn}$ is the attention layer, which can be one of the followings,
\begin{eqnarray}
\text{(Fourier-type attention)}\quad \mathrm{Attn}(\bm{y}_{\mathrm{in}})\triangleq\left(\widetilde{\bm{Q}} \widetilde{\bm{K}}^{\top} \right)
\bm{V} / n,\\
\text{(Galerkin-type attention)}\quad
\mathrm{Attn}(\bm{y}_{\mathrm{in}})\triangleq \bm{Q} \left(\widetilde{\bm{K}}^{\top} 
\widetilde{\bm{V}}\right) / n,
\end{eqnarray}
where $\widetilde{\cdummy}$ denotes learnable non-batch-based normalization. Here, we can find the softmax function does not show up, reducing much computational effort for high-dimensional inputs.

One of the limitations of the aforementioned attention mechanism is that the input and output feature embedding $\bm{y}_{\mathrm{in}}$ and $\bm{y}_{\mathrm{out}}$ literally live on the same grid. To address this limitation and allow for arbitrary query locations, \cite{li2022transformer} proposed the cross-attention mechanism, where the query matrix $\bm{Q}$ is encoded from the query locations instead of the input feature embedding $\bm{y}_{\mathrm{in}}$. Another study \cite{kissas2022learning} has developed the kernel-couple attention mechanism to better model the correlations between the query locations, which can be described as:
\begin{equation}
    G_w(\theta)(\bm{x}) = \sum_{i=1}^n \mathrm{softmax}\left(  \int_{\Omega} \kappa(\bm{x},\bm{x}') g(\bm{x}')  \mathd \bm{x'} \right)_i \odot \bm{e}_i(\theta),
\end{equation}
where $\kappa\colon \Omega \times \Omega \rightarrow \mathbb{R}$ is a kernel function, $g$ is a score function (which can be parameterized as a neural network), and $\bm{e}_i$ is the input feature encoder. We remark that the above formalization is much like a variant of DeepONet but with the advanced attention mechanism.

By introducing the attention mechanism, such models as described above can better model complicated properties of the latent operator, which are beneficial for real-world physical systems. In the future, big models incorporating attention mechanisms will be proposed, just like BERT \cite{devlin2018bert} in natural language processing. However, such models should have a larger number of parameters and will require more training samples. Also, samples obtained from physical systems are often very expensive. How to address this contradiction will be the key to popularizing big models in the field of neural operators.


\subsubsection{Graph-based Operator Learning}\label{sec_graph_operator_learning}
Graphs have gained much popularity in the community of science and engineering, as an expressive structure for modeling interactions between individuals and discretizing continuous space. In particular, the standard output of the numerical PDE solver (e.g., FEM) is a triangle mesh which is a specific type of graph. Therefore, it is natural to model the grid $\{ \bm{x}_i \}_{i=1}^N$ (see Section~\ref{sec_grid_operator_learning}) as a graph $\mathcal{G}=(\mathcal{V}, \mathcal{E})$ with nodes $\bm{x}_i\in \mathcal{V}$, edges $e_{ij}\in \mathcal{E}$, and node features, including the input and output functions $\{ v(\bm{x}_i) \}_{i=1}^N$ and $\{ u(\bm{x}_i) \}_{i=1}^N$. Our goal is to learn the latent operator $\tilde{G}$ in Equation~\eqref{eq_grid_based_operator_2} defined on the graph $\mathcal{G}$ in a data-driven manner.

\textbf{Graph neural operators.}
Inspired by the format of Green's function (see Equation~\eqref{eq_green_function}), Li \textit{et. al.} have introduced a graph kernel network into operator learning \cite{li2020neural}. The model can be described as
\begin{eqnarray}
z_0(\bm{x}) &=& \bm{P}\left(\bm{x}, v(\bm{x}), v_{\epsilon}(\bm{x}), \nabla v_{\epsilon}(\bm{x}) \right) + p,\\
z_{t+1}(\bm{x}) &=& \sigma\Big(  \bm{W}z_t(\bm{x}) + \int_{\Omega} \kappa_{\phi} \big( \bm{x}, \bm{y}, \nonumber\\
&& v(\bm{x}), v(\bm{y}) \big) z_t(\bm{y}) \nu_{\bm{x}}( \mathd \bm{y}) \Big),\label{eq_gnn_operator_2}\\
u(\bm{x}) &=& \bm{Q} z_T(\bm{x}) + q,
\end{eqnarray}
where $t=0,\dots,T-1$ denote the indexes of hidden feature embeddings, $\bm{P}\in\mathbb{R}^{n\times 2(d+1)}$ (we assume $\bm{x}\in\mathbb{R}^d$ is a pure spatial coordinate), $p,z_0,\dots,z_T\in \mathbb{R}^n$, $\bm{Q}\in\mathbb{R}^{1\times n}$, and $q\in \mathbb{R}$. In addition, $\sigma$ is an activation function, $\nu_{\bm{x}}$ is a fixed Borel measure for $\forall \bm{x} \in \Omega$, and $v_{\epsilon}$ is a Gaussian smoothed version of the input function $v$. The learnable weights include $\bm{W}\in \mathbb{R}^{n\times n}$ and the parameters $\phi$ of the kernel $\kappa_{\phi}\colon \mathbb{R}^{2(d+1)}\rightarrow \mathbb{R}^{n\times n}$. With the help of message-passing architectures, we can approximate the integral in Equation~\eqref{eq_gnn_operator_2} as,
\begin{eqnarray}\label{eq_gnn_operator_kernel}
    z_{t+1}(\bm{x}_i)&\approx& \sigma\Big(  \bm{W}z_t(\bm{x}_i) + \frac{1}{| \mathcal{N}(i) |} \sum_{\bm{x}_j \in \mathcal{N}(i)} \nonumber\\
    &&\kappa_{\phi} \left( \bm{x}_i, \bm{x}_j, v(\bm{x}_i), v(\bm{x}_j) \right) z_t(\bm{x}_j) \Big),
\end{eqnarray}
where $\mathcal{N}(i)$ is the neighborhood of the node $\bm{x}_i$ and $\kappa_{\phi}$ is parameterized by a neural network. Here, it is noted that the input $\bm{x}_i$ is located on the graph but is not an arbitrary coordinate in the implementation. 

Based on the above framework, Li \textit{et. al.} later proposed the multipole graph neural operator \cite{li2020multipole}, which decomposes the kernel into multi-level sub-kernels to capture short-to-long-range interactions instead of only neighboring interactions (see Equation~\eqref{eq_gnn_operator_kernel}) with linear complexity in the node numbers. Compared with the vanilla graph neural operator, such a model can better gather global information and thus has better generalizability. Recently, another study \cite{brandstetter2022message} combined the graph neural operator with autoregressive methods to facilitate learning operators defined by time-dependent PDEs. Further, an encoder-decoder mechanism was incorporated into the network architecture to boost the expression ability. 

Graph neural operators are good at dealing with inputs of the format of the graph which can allow for unstructured discretization and model the interactions between nodes via edge features (which are very useful for some physical systems like multi-particle systems). However, we also notice that processing dense graphs can be computationally unfavorable for such models.  



\subsubsection{Open Challenges and Future Work}\label{sec_neural_operator_challenge}
As previously mentioned, there has been fruitful work along with impressive success in this field of the neural operator. Moreover, some of these achievements have found their stages in the vast land of applications in science and engineering, which will be discussed later in Section~\ref{sec:application}. Still, the neural operator is a young and fast-growing field, where many open challenges remain to be solved. Some of them are noted as follows.
\begin{itemize}
    \item \textbf{Incorporating physical priors.} Introducing physical priors (including fully or partially known governing equations, physical intuition, conservation, and symmetry, etc.) into training can effectively improve the generalizability of the model and reduce training data demands. Nowadays, the most popular method is to employ the framework of physics-informed learning (see Section~\ref{sec_pinn_intro}). However, the speed and accuracy of physics-informed learning are far inferior to traditional numerical methods such as the FEM in solving PDEs \cite{krishnapriyan2021characterizing}. Future work includes theoretically improving the framework of physics-informed learning or pursuing schemes that can effectively combine physical priors with data-driven approaches.
    \item \textbf{Reducing the cost of gathering datasets.} The basic goal of the neural operator is to learn a mapping from the parameter space to the solution space via data-driven methods. To this end, a considerable number of samples are often required for a dataset. The samples are generated by numerical methods or experiments, both of which are very expensive, especially when the dimension of the parameter $\theta$ is high and the problem domain $\Omega$ is geometrically complex. This greatly limits the application of neural operators. How to deal with the expense of data generation will become one of the major challenges in the future.
    \item \textbf{Developing large pre-trained models.} At present, the main neural operator methods are all based on small models, and can only solve a specific type of parameterized physical system in one training. In the future, it may become a trend to develop pretrained large models that can be reused by various physical systems (after fine tuning), as in other machine learning fields such as computer vision and natural language processing. In this way, the model only needs to be trained once and can be employed in many other physical problems, reducing the time overhead of neural operator training and data generation.
    \item \textbf{Modeling real-world physical systems.} Many numerical testing experiments for neural operators are based on idealized physical systems which are far from real-world physical systems. Modeling real-world physical systems will bring more challenges, including geometrically complex problems, high-dimensional problems (such as optimal control problems \cite{luus1990application}), chaotic problems (where the resulting solution is very sensitive to the initial conditions \cite{williams1997chaos}), long-term prediction (which needs to deal with the accumulated error over time), etc. More research in this area will be the key to applying neural operators to practical problems.
\end{itemize}

\subsection{Theory}

In this subsection, we introduce some preliminary explorations of theoretical justification for physics-informed machine learning, especially for PINNs and DeepONet. First, we introduce the expression ability of neural networks, like DeepONet, for approximating operators. Then we present some work about the convergence of PINNs. Furthermore, we introduce some analyses of approximation and generalization errors. Finally, we mark the open challenges and future work in this field.


\subsubsection{Expression Ability}
In statistical machine learning, a hypothesis is a mapping of features onto labels, and the hypothesis set is the set of hypothesis~\cite{mohri2018foundations}. Different algorithms will choose different hypothesis sets and hope to find the optimal hypothesis of the set. For example, the hypothesis set of linear regression is all linear mappings. In most machine learning tasks, our goal is to find a proper hypothesis in the hypothesis space through some algorithm. Since the hypothesis set is the subset of the set of all possible mappings and the optimal mapping may be not in the hypothesis, the expression ability of the hypothesis space determines the optimal hypothesis, of which the analysis is significant.

It is well known that multi-layer neural networks are universal approximators, i.e., they can approximate any measurable function to arbitrary accuracy~\cite{hornik1989multilayer}. There is a similar and more interesting result for approximating operators, i.e., one-layer neural networks can approximate any operator, which is a mapping from a space of functions to another space of functions, to arbitrary accuracy~\cite{chen1995universal}. Based on this result, current work~\cite{lu2021learning} points out that DeepONet~\cite{lu2021learning} with branch net as well as trunk net, can approximate any operator. We repeat this conclusion as follows:
\begin{theorem}[Universal Approximation Theorem for DeepONet~\cite{chen1995universal,lu2021learning}]
\label{thm_don_uat}
Suppose $X$ is a Banach space, $K_1\subseteq X, K_2\subseteq\mathbb{R}^d,V\subseteq C(K_1)$ are compact sets, here $C(K_1)$ represents the set of all continuous functions in $K_1$. Let $G:V\rightarrow C(K_2)$ be a nonlinear continuous operator, i.e., for any function $u\in V$, $G(u)\in C(K_2)$, then for $\forall y\in K_2\subseteq \mathbb{R}^d$, $G(u)(y)\in\mathbb{R}$.

Then for $\forall\epsilon>0$, there $\exists n,p,m\in\mathbb{N}$, constants $c_i^k, \xi_{ij}^k, \theta_i^k, \zeta_k\in\mathbb{R}, w_k\in\mathbb{R}^d, x_j\in K_1, i =1,...,n,k=1,...,p,j=1,..,m$, satisfying that
\begin{equation}
\small
    \left|G(u)(y) - \sum_{k=1}^p \underbrace{\sum_{i=1}^n c_i^k \sigma\left(\sum_{j=1}^m \xi_{ij}^k u(x_j) + \theta_i^k\right)}_{branch} \underbrace{\sigma(w_k y + \zeta_k)}_{trunk} \right| \leq \epsilon,
\end{equation}
for $\forall u\in V$ and $y\in K_2$.
\end{theorem}
Theorem~\ref{thm_don_uat} indicates that the expressive ability of DeepONet is powerful enough to approximate any nonlinear continuous operator, which reveals the potential application of neural networks for learning operators. However, in Theorem~\ref{thm_don_uat}, we not only prove the existence of such a neural network but also the size of this neural network, which is significant for designing specific networks, and is not fully understood. 


Although these works~\cite{chen1995universal,lu2021learning} have revealed the powerful expression ability of one-layer neural networks, some follow-up work~\cite{yu2021arbitrary} points that wide, shallow neural networks may need exponentially many
neurons to obtain similar expression ability with deep, narrow ones. And we introduce their results as below
\begin{theorem}[\cite{yu2021arbitrary}]
\label{thm-deep-shalow}
Suppose $X$ is a Banach space, $K_1\subseteq X,V\subseteq C(K_1)$ are compact sets. Then, for $\forall n,k\in\mathbb{N}, n,k\ge 1$, $\exists G_k:V\rightarrow C([0,1]^n)$ as a nonlinear continuous operator satisfying that
\begin{itemize}
    \item There exists a ReLU neural network $\phi$ mapping from $[0,1]^n$ to $\mathbb{R}$ with depth $2k^3+8$ and width $\Theta(1)$ satisfying that $\phi(y) = G_k(u)(y)\forall u\in V, y\in[0,1]^n$.
    \item Let $m\ge 1$ be an integer and $\psi:[0,1]^{m+n}\rightarrow \mathbb{R}$ be a ReLU neural network of which the depth $\leq k$ and the total nodes $\leq 2^k$. Then for any $x_1, ..., x_m\in K_1, u\in V$, we have
    \begin{equation}
        \int_{[0,1]^d}\left|G_k(u)(y) - \psi(u(x_1), ..., u(x_m), y)\right| dy \ge \frac{1}{64}
    \end{equation}
\end{itemize}
\end{theorem}
Theorem~\ref{thm-deep-shalow} conducts a nonlinear continuous operator, which can be approximated by a deep, narrow neural network with depth $2k^3+8$ and width $\Theta(1)$. However, there are no wide, shallow neural networks with depth $\leq k$ and total nodes $\leq 2^k$ that can effectively approximate it. This result illustrates that deeper neural networks might be more efficient for approximating operators, to some degree. Furthermore, \cite{yu2021arbitrary} provides an upper bound of the width of the deeper neural networks for approximating operators. The results are as follows
\begin{theorem}[\cite{yu2021arbitrary}]
\label{thm-deep-width}
Assume that the activation function $\sigma$ satisfies some mild assumptions (details are in~\cite{yu2021arbitrary}), then for $\forall \epsilon>0$, $\exists F: \mathbb{R}^{m+n}\rightarrow \mathbb{R}$ is a $\sigma$-activated neural network with width at most $m+n+5$ satisfying
\begin{equation}
    |G(u)(y) - F(u(x_1), ..., u(x_m), y)|<\epsilon,
\end{equation}
for $\forall u, y$. Moreover, if we can construct a $\sigma$-activated neural network with width 3 and depth $L$ to approximate the mapping $(a,b)\rightarrow ab$ to any error, then we can prove that the network $F$ has depth $\mathcal{O}(M+N+L)$, here $M,N,m,\{x_i\}_{i=1}^m$ are the same as the notation of Theorem 5 in~\cite{chen1995universal}.
\end{theorem}
Theorem~\ref{thm-deep-width} provides a theoretical guarantee of deep neural networks for approximating operators with the upper bound of the width and the depth of the network. 


Also, the expression ability of other architectures is worth studying and there are also some relevant conclusions; for example, \cite{kovachki2021universal} provides a universal approximation theorem for FNO~\cite{li2020fourier}.

\subsubsection{Convergence}
In statistical machine learning, whether an algorithm converges and its convergence speed are important indexes to evaluate the algorithm. For example,~\cite{shamir2013stochastic} analyzes the convergence property of stochastic gradient descent. Since the properties of physics equations are relatively complicated, the convergence properties of physics-informed machine learning algorithms have not been well studied.

\cite{shin2020convergence} takes a first step to analyze the convergence of PINNs for solving time-independent PDEs, i.e.,
\begin{eqnarray}
\label{theory-1}
  \mathcal{F} (u ) (\tmmathbf{x}) & = & f(\tmmathbf{x}), \tmmathbf{x}\in\Omega  \\
  \mathcal{B} (u ) (\tmmathbf{x}) & = & g(\tmmathbf{x}), \tmmathbf{x}\in\partial\Omega, \nonumber
\end{eqnarray}
where $\mathcal{F}$ is the differential operator and $\mathcal{B}$ is the boundary condition. We assume that these PDEs have a unique classical solution $u$ and hope to approximate the solution of the PDEs with a set of training data, including residual data and initial/boundary data. Moreover, we denote the number of training set as the vector $\tmmathbf{m} = (m_r, m_b)$, where $m_r, m_b$ are the number of training points of residual data and initial/boundary data respectively. We assume that $\{(\tmmathbf{x}_r^i, f(\tmmathbf{x}_r^i))\}_{i=1}^{m_r}$ is the set of the residual data and $\{(\tmmathbf{x}_b^i, f(\tmmathbf{x}_b^i))\}_{i=1}^{m_b}$ is the set of the initial/boundary data, here $\tmmathbf{x}_r^i \in \Omega,i=1,2,...,m_r; \tmmathbf{x}_b^j \in \partial \Omega,j=1,2,...,m_b$. Given a set of neural networks $\mathcal{H}_{n}$ as our hypothesis set, then for any mapping $h\in\mathcal{H}_n$, we can define its empirical PINN loss via $\{(\tmmathbf{x}_r^i, f(\tmmathbf{x}_r^i))\}_{i=1}^{m_r}$ and $\{(\tmmathbf{x}_b^i, f(\tmmathbf{x}_b^i))\}_{i=1}^{m_b}$ as
\begin{equation}
\begin{split}
\label{pinn_loss_re}
    &\mathrm{Loss}_m (h;\bm{\lambda}, \bm{\lambda}^R)\\
    =& (\lambda_r\|\mathcal{F} (h) (\tmmathbf{x}_r) - f(\tmmathbf{x}_r) \|^2)\mathbb{I}_{\Omega}(\tmmathbf{x}_r) + \lambda_r^R R_r(h)\\
    +&(\lambda_b\|\mathcal{B} (h) (\tmmathbf{x}_b) - g(\tmmathbf{x}_b) \|^2)\mathbb{I}_{\partial\Omega}(\tmmathbf{x}_b) + \lambda_b^R R_b(h),
\end{split}
\end{equation}
here $\mathbb{I}$ is the indicated function, $R_r, R_b$ are regularization functions, and $\bm{\lambda} = (\lambda_r, \lambda_b), \bm{\lambda}^R = (\lambda_r^R, \lambda_b^R)$ are hyperparameters. Moreover, we can define the expected PINN loss as
\begin{equation}
    \mathrm{Loss} (h;\bm{\lambda}, \bm{\lambda}^R) = \mathbb{E} [\mathrm{Loss}_m (h;\bm{\lambda}, \bm{\lambda}^R)].
\end{equation}
Our goal is to 
minimize the expected PINN loss in the hypothesis set, i.e.,
\begin{equation}
    \min_{h\in\mathcal{H}_n} \mathrm{Loss} (h; \bm{\lambda}, \bm{\lambda}^R).
\end{equation}
However, in practice, it is difficult to calculate the expected PINN loss and we always use $\mathrm{Loss}_m (h;\bm{\lambda}, \bm{\lambda}^R)$ as an approximator, i.e.,
\begin{equation}
    \min_{h\in\mathcal{H}_n} \mathrm{Loss}_m (h; \bm{\lambda}, \bm{\lambda}^R).
\end{equation}
Consequently, it is significant to analyze the difference between it and the true solution.

Based on (\ref{pinn_loss_re}), if we take $\bm{\lambda}^R=0$, we can get the standard empirical PINN loss and the standard expected PINN loss as
\begin{equation}
\begin{split}
    \mathrm{Loss}_m^{\mathrm{PINN}}(h; \bm{\lambda}) &= \frac{\lambda_r}{m_r}\sum_{i=1}^{m_r}\|\mathcal{F} (h) (\tmmathbf{x}_r^i) - f(\tmmathbf{x}_r^i) \|^2\\
    &+ \frac{\lambda_b}{m_b}\sum_{i=1}^{m_b}\|\mathcal{B} (h) (\tmmathbf{x}_b^i) - g(\tmmathbf{x}_b^i) \|^2,\\
    \mathrm{Loss}^{\mathrm{PINN}}(h; \bm{\lambda}) &= \lambda_r \|\mathcal{F} (h) - f\|^2_{L^2}+ \lambda_b \|\mathcal{B} (h) - g\|^2_{L^2}.
\end{split}
\end{equation}
First, \cite{shin2020convergence} utilizes the regularized empirical loss for bounding the expected PINN loss. The result is shown below:
\begin{theorem}[\cite{shin2020convergence}]
\label{thm-pinn_loss}
Under some assumptions (details are in~\cite{shin2020convergence}), we set $m_r$ and $m_b$ be the number of points sampled from $\Omega$ and $\partial\Omega$ respectively. 

If $d\ge 2$, with probability at least $(1 - \sqrt{m_r}(1-1/\sqrt{m_r})^{m_r})(1 - \sqrt{m_b}(1-1/\sqrt{m_b})^{m_b})$, we can prove that
\begin{equation}
    \mathrm{Loss}^{\mathrm{PINN}} (h;\bm{\lambda}) \leq C_m \mathrm{Loss}_m (h;\bm{\lambda},\hat{\bm{\lambda}}_m^R) + C' (m_r^{-\frac{\alpha}{d}} + m_b^{-\frac{\alpha}{d-1}}),
\end{equation}
where $C_m$ and $C'$ are constants.

If $d = 1$, with probability at least $1 - \sqrt{m_r}(1-1/\sqrt{m_r})^{m_r}$, we can prove that
\begin{equation}
    \mathrm{Loss}^{\mathrm{PINN}} (h;\bm{\lambda}) \leq C_m \mathrm{Loss}_m (h;\bm{\lambda},\hat{\bm{\lambda}}_m^R) + C' m_r^{-\frac{\alpha}{d}},
\end{equation}
here $C_m$ and $C'$ are constants.
\end{theorem}
Furthermore, by setting the regular term $R_r(h) = \mathcal{F}[h], R_b(h) = \mathcal{F}[h]$, we can define the H\"older regularized empirical PINN loss as
\begin{equation}
\small
\begin{split}
    &\mathrm{Loss}_m (h;\bm{\lambda}, \bm{\lambda}^R)\\
    = &\left\{
    \begin{aligned}
    &\mathrm{Loss}_m^{\mathrm{PINN}} (h;\bm{\lambda})+\lambda_{r,m}^R [\mathcal{F}[h]]_{\alpha,U}^2+\lambda_{b,m}^R [\mathcal{B}[h]]_{\alpha,\Gamma}^2,\ \text{if}\ d\ge 2\\
    &\mathrm{Loss}_m^{\mathrm{PINN}} (h;\bm{\lambda})+\lambda_{r,m}^R [\mathcal{F}[h]]_{\alpha,U}^2,\ \text{if}\ d=1
    \end{aligned}
    \right.
\end{split}
\end{equation}
Based on Theorem~\ref{thm-pinn_loss},~\cite{shin2020convergence} minimizes a high-probability upper bound of the expected PINN loss by minimizing the H\"older regularized loss. Moreover,~\cite{shin2020convergence} further shows that the minimum of H\"older regularized empirical loss has a small expected PINN loss and will converge to the ground truth, i.e., 
\begin{theorem}[\cite{shin2020convergence}]
\label{thm_pinn_conve}
Under some assumptions (details are in~\cite{shin2020convergence}), we set $m_r$ and $m_b$ as the number of points sampled from $\Omega$ and $\partial\Omega$ respectively and set
\begin{equation}
    h_{m_r} = \mathop{\arg\min}_{h\in\mathcal{H}_{m_r}} \mathrm{Loss}_{m_r}(h;\bm{\lambda}, \bm{\lambda}^R)
\end{equation}
Then we have following results
\begin{itemize}
    \item With probability at least $(1 - \sqrt{m_r}(1-c_r/\sqrt{m_r})^{m_r})(1 - \sqrt{m_b}(1-c_b/\sqrt{m_b})^{m_b})$, we can prove that
\begin{equation}
    \mathrm{Loss}^{\mathrm{PINN}} (h_{m_r};\bm{\lambda}) = \mathcal{O}(m_r^{-\frac{\alpha}{d}})
\end{equation}
    \item With probability 1, we can prove that
\begin{equation}
    \lim_{m_r\rightarrow\infty} \mathcal{F}[h_{m_r}] = f, \lim_{m_r\rightarrow\infty} \mathcal{B}[h_{m_r}] = g.
\end{equation}
\end{itemize}
\end{theorem}
Theorem~\ref{thm_pinn_conve} propose a general convergence analyses for PINNs under some conditions. Moreover,~\cite{shin2020convergence} provides more detailed conclusions for linear elliptic PDEs and linear parabolic PDEs.

Besides PINNs, there is other work focusing on the convergence of other physics-informed machine learning methods-- for example,~\cite{dondl2021uniform} provides uniform convergence guarantees for the Deep Ritz Method for Nonlinear Problems and ~\cite{de2021convergence} provides convergence rates for learning linear operators from noisy data.


\subsubsection{Error Estimation}
In statistical machine learning, error estimation is significant for analyzing high or low performance of the model and guiding us to design better algorithms. Also, the error estimation of physics-informed machine learning is relatively preliminary and there is only a little research on this topic. 

A recent work~\cite{lanthaler2022error} analyzes the approximation and generalization errors of DeepONet~\cite{lu2021learning}. First,~\cite{lanthaler2022error} provides a more general form of DeepONet. Set $D\subseteq\mathbb{R}^d, U\subseteq\mathbb{R}^n$ are two compact sets. We consider three operators for building DeepONet:
\begin{itemize}
    \item \textbf{Encoder.} Given a set of points $\{x_i\}_{i=1}^m,x_i\in D$,~\cite{lanthaler2022error} first defines the Encoder $\mathcal{E}$ as
    \begin{equation}
        \mathcal{E}: C(D)\rightarrow \mathbb{R}^m,\quad \mathcal{E}(u) = (u(x_1), ..., u(x_m)),
    \end{equation}

    \item \textbf{Approximator.} Under the same set of points $\{x_i\}_{i=1}^m$,~\cite{lanthaler2022error} defines the approximator $\mathcal{A}$ as a neural network satisfying that
    \begin{equation}
        \mathcal{A}: \mathbb{R}^m \rightarrow \mathbb{R}^p,\quad \{u_j\}_{j=1}^m\rightarrow \{\mathcal{A}_k\}_{k=1}^p.
    \end{equation}

    \item \textbf{Reconstructor.} To define the Reconstructor $\mathcal{R}$,~\cite{lanthaler2022error} first uses a neural network $\boldsymbol{\tau}$ to denote a trunk net as
    \begin{equation}
        \boldsymbol{\tau}:\mathbb{R}^n\rightarrow\mathbb{R}^{p+1},\quad y=(y_1, ..., y_n)\rightarrow \{\tau_k(y)\}_{k=0}^p
    \end{equation}
    Then, based on $\boldsymbol{\tau}$,~\cite{lanthaler2022error} defines the Reconstructor $\mathcal{R}$ as
    \begin{equation}
    \begin{split}
        &\mathcal{R} = \mathcal{R}_{\boldsymbol{\tau}}: \mathbb{R}^p\rightarrow C(U),\\
        &\mathcal{R}_{\boldsymbol{\tau}}(\{\mathcal{A}_k\}_{k=1}^p) = \tau_0 + \sum_{k=1}^p \mathcal{A}_k \tau_k\in C(U),\\
        &\mathcal{R}_{\boldsymbol{\tau}}(\{\mathcal{A}_k\}_{k=1}^p)(y) = \tau_0(y)+ \sum_{k=1}^p \mathcal{A}_k \tau_k(y), \forall y\in U.
    \end{split}
    \end{equation}
\end{itemize}
Thus,~\cite{lanthaler2022error} can combine them to get the DeepONet $\mathcal{N}$ as
\begin{equation}
    \mathcal{N}: C(D)\rightarrow C(U),\quad \mathcal{N}(u) = (\mathcal{R}\circ\mathcal{A}\circ\mathcal{E})(u),
\end{equation}
where $\boldsymbol{\beta} = \mathcal{A}\circ\mathcal{E}$ is the branch net. For $\forall y\in U$, we have
\begin{equation}
\begin{split}
    \mathcal{N}(u)(y) &= (\mathcal{R}\circ\mathcal{A}\circ\mathcal{E})(u)(y)\\
    &=\tau_0(y)+ \sum_{k=1}^p \mathcal{A}_k \tau_k(y).
\end{split}
\end{equation}
First,~\cite{lanthaler2022error} analyzes the approximation error of DeepONet. Assume that the underlying operator is $\mathcal{G}:C(D)\rightarrow C(U)$ and there is a fixed  probability measure $\mu\in \mathcal{P}(X)$, we can naturally use the $L^2(\mu)$-norm of $\mathcal{G}$ and our DeepONet $\mathcal{N}$ to measure the approximation error of $\mathcal{N}$ as
\begin{equation}
\label{don_ea_error}
    \hat{\mathcal{E}} = \left(\int_{X}\int_{U} |\mathcal{G}(u)(y) - \mathcal{N}(u)(y)|^2 dy d\mu(u)\right)^{1/2}.
\end{equation}
However, since $\mathcal{N}$ consists of different operators, it is challenging to directly analyze $\hat{\mathcal{E}}$. To handle this problem, the main result of~\cite{lanthaler2022error} is to decompose $\hat{\mathcal{E}}$ into three parts: encoding error $\hat{\mathcal{E}}_{\mathcal{E}}$, approximation error $\hat{\mathcal{E}}_{\mathcal{A}}$ and reconstruction error $\hat{\mathcal{E}}_{\mathcal{R}}$.

To better analyze the encoder $\mathcal{E}$ and the reconstructor $\mathcal{R}$, we define the decoder $\mathcal{D}$ and the projector $\mathcal{P}$ satisfying: 
\begin{equation}
\begin{split}
    &\mathcal{E}\circ\mathcal{D} = \mathrm{Id}:\mathbb{R}^m\rightarrow\mathbb{R}^m,\\
    &\mathcal{D}\circ\mathcal{E} \approx \mathrm{Id}:X\rightarrow X,\\
    &\mathcal{P}\circ\mathcal{R} = \mathrm{Id}:\mathbb{R}^p\rightarrow \mathbb{R}^p,\\
    & \mathcal{R}\circ\mathcal{P} \approx \mathrm{Id}:Y\rightarrow Y.
\end{split}
\end{equation}
Then, we can define the encoding error $\hat{\mathcal{E}}_{\mathcal{E}}$, the approximation error $\hat{\mathcal{E}}_{\mathcal{A}}$ and the reconstruction error $\hat{\mathcal{E}}_{\mathcal{R}}$ respectively as below
\begin{equation}
\begin{split}
    &\hat{\mathcal{E}}_{\mathcal{E}} = \left(\int_X \|\mathcal{D}\circ\mathcal{E}(u) - u\|_X^2 d\mu(u)\right)^{1/2},\\
    &\hat{\mathcal{E}}_{\mathcal{A}} = \left(\int_{\mathbb{R}^m}\|\mathcal{A}(u) - \mathcal{P}\circ\mathcal{G}\circ\mathcal{D}(u)\|^2_{l^2(\mathbb{R}^p)} d(\mathcal{E}_{\#}\mu)(u)\right)^{1/2},\\
    &\hat{\mathcal{E}}_{\mathcal{R}} = \left(\int_{\mathbb{R}^m}\|\mathcal{R}\circ\mathcal{P}(u) - u\|^2_{L^2(U)} d(\mathcal{G}_{\#}\mu)(u)\right)^{1/2}.
\end{split}
\end{equation}
Furthermore,~\cite{lanthaler2022error} decomposes $\hat{\mathcal{E}}$ as below
\begin{theorem}[\cite{lanthaler2022error}]
\label{thm_deep_error}
Under some mild assumptions (details in~\cite{lanthaler2022error}), the error (\ref{don_ea_error}) can be bounded by
\begin{equation}
    \hat{\mathcal{E}} \leq \mathrm{Lip}_{\alpha}(\mathcal{G})\mathrm{Lip}(\mathcal{R}\circ\mathcal{P})(\hat{\mathcal{E}}_{\mathcal{E}})^{\alpha} + \mathrm{Lip(}\mathcal{R})\hat{\mathcal{E}}_{\mathcal{A}} + \hat{\mathcal{E}}_{\mathcal{R}},
\end{equation}
here $\mathcal{G}$ is $\alpha$-H$\ddot o$lder continuous and $\mathrm{Lip}_{\alpha},\mathrm{Lip}$ is defined for any mapping $\mathcal{F}:\mathcal{X}\rightarrow\mathcal{Y}$ between Banach spaces $\mathcal{X},\mathcal{Y}$:
\begin{equation}
\begin{split}
    \mathrm{Lip}_{\alpha}(F)=& \sup_{u,u' \in \mathcal{X}} \frac{\|\mathcal{F}(u) - \mathcal{F}(u')\|_{\mathcal{Y}}}{\|u-u'\|_{\mathcal{X}}^{\alpha}}, \\
    \mathrm{Lip}(F)=&\mathrm{Lip}_{1}(F).
\end{split}
\end{equation}
\end{theorem}
Based on Theorem~\ref{thm_deep_error},~\cite{lanthaler2022error} further provides more detailed analyses for bounding $\hat{\mathcal{E}}_{\mathcal{E}},\hat{\mathcal{E}}_{\mathcal{A}},\hat{\mathcal{E}}_{\mathcal{R}}$, which are helpful for bounding and analyzing $\hat{\mathcal{E}}$.

Besides approximation error,~\cite{lanthaler2022error} also provides analysis of the generalization error of DeepONet. Given the underlying operator $\mathcal{G}$, we hope to train the DeepONet $\mathcal{N}$ that minimizes the loss function
\begin{equation}
\label{don_ea_loss_1}
    \hat{\mathcal{L}}(\mathcal{N}) = \int_{L^2(D)} \int_U |\mathcal{G}(u)(y) - \mathcal{N}(u)(y)|^2 dy d\mu(u).
\end{equation}
However, we can not directly calculate $\hat{\mathcal{L}}$ and we always use empirical loss as a surrogate. To approximate it, we always sample
\begin{equation}
    U_1,U_2,...,U_N\sim \mu,\ Y_1,Y_2,...,Y_N\sim \mathrm{Unif}(U),
\end{equation}
and define the empirical loss as
\begin{equation}
\label{don_ea_loss_2}
    \hat{\mathcal{L}}_N(\mathcal{N}) = \frac{|U|}{N} \sum_{j=1}^N |\mathcal{G}(U_j)(Y_j) - \mathcal{N}(U_j)(Y_j)|^2.
\end{equation}
Assume that $\hat{\mathcal{N}},\hat{\mathcal{N}}_N$ be the minimizer of the loss~(\ref{don_ea_loss_1}) and the loss~(\ref{don_ea_loss_2}) respectively, i.e.
\begin{equation}
\begin{split}
    \hat{\mathcal{N}} = & \mathop{\arg\min}_{\mathcal{N}} \hat{\mathcal{L}}(\mathcal{N}),\\
    \hat{\mathcal{N}}_N = & \mathop{\arg\min}_{\mathcal{N}}  \hat{\mathcal{L}}_N(\mathcal{N}),
\end{split}
\end{equation}
we can define the generalization error as
\begin{equation}
    (\hat{\mathcal{E}}_{\mathrm{gen}})^2 = \hat{\mathcal{L}}(\hat{\mathcal{N}}_N) - \hat{\mathcal{L}}(\hat{\mathcal{N}}).
\end{equation}
Moreover,~\cite{lanthaler2022error} provides Theorem~\ref{thm_deep_gene} as below to bound the generalization error.
\begin{theorem}[\cite{lanthaler2022error}]
\label{thm_deep_gene}
Under some assumptions (details are in~\cite{lanthaler2022error}), we can bound the generalization error as
\begin{equation}
    \mathbb{E}\left[\left|\hat{\mathcal{L}}(\hat{\mathcal{N}}_N) - \hat{\mathcal{L}}(\hat{\mathcal{N}})\right|\right] \leq \frac{C}{\sqrt{N}} (1 + Cd_{\theta}\log(CB\sqrt{N})^{2\kappa+1/2}),
\end{equation}
here $C,d_{\theta},B,\kappa$ are constants.
\end{theorem}



Besides this work analyzing the error estimation of DeepONet, there is also some work focus on estimating the error of other architectures. For example,~\cite{de2021error} estimates the error of PINN for Linear Kolmogorov PDEs, and~\cite{kovachki2021universal} analyzes the error of FNO.

\subsubsection{Open Challenges and Future Work}
As previously mentioned, although physics-informed machine learning has received more and more attention, and although some representative algorithms, like PINNs and DeepONet, have shown encouraging performance, their theoretical properties have not been well explored. Since theoretical justification, including expression ability, convergence, and error estimation, is significant for guiding to design better algorithms, these fields are of great value in future research. Besides current progress as mentioned above, we mark some open problems of theoretical justification for physics-informed machine learning as follows,
\begin{itemize}
    \item \textbf{Expression Ability} The expression ability of neural networks for approximating operators has made great progress, but there are still some challenges and open problems. First, although some work~\cite{yu2021arbitrary} has discussed the expression ability of deep, narrow neural networks and wide, shallow networks, for approximating operators, why deep networks have better expression ability has not been well studied. Moreover, how to design more effective architecture to approximate operators with fewer nodes is significant for designing more stable and effective algorithms. Besides DeepONets, the expression ability of other architectures is also worth more in-depth analysis.
    \item \textbf{Convergence} The convergence of physics-informed machine learning algorithms is significant to evaluate their effectiveness. Unfortunately, the current research in this field is still very preliminary since analyzing the stability of PDEs itself is complicated. How to analyze the convergence of PINNs for different kinds of PDEs will be one of the major challenges in the future and can inspire us to design more efficient architectures and algorithms. Moreover, the convergence of other algorithms like DeepONets needs further exploration
    \item \textbf{Error Estimation} At present, some studies have preliminarily considered the errors of physics-informed machine learning algorithms, like DeepONet and FNO. There are two kinds of error estimation that have attracted more and more attention, i.e., approximation error and generalization error. Carefully analyzing approximation error is beneficial for designing more effective algorithms. Moreover, analyzing generalization error and improving the generalization of algorithms by using physics knowledge are also noteworthy directions for developing more general and stable algorithms. More research in this field will be the key to better understanding and combining physics knowledge and data.
\end{itemize}

\subsection{Application}
\label{sec:application}

Physics-informed machine learning is playing a more and more important role in various fields and solving some problems that cannot be accomplished by traditional methods. 
In this section, we briefly introduce some important applications of PIML in several fields, including fluid dynamics, material science, optimal control, and scientific discovery.

\subsubsection{Fluid Dynamics}
Fluids are one of the most difficult physical systems due to the high nonlinearity and mathematical complexity of the governing equations. An example is the Navier-Stokes equation, where chaos may occur under certain conditions. Therefore,  methods of physical information learning have been introduced in this field to solve many problems that are difficult for traditional methods. Applications mainly include predicting fluid dynamics (non-newtonian fluids \cite{mahmoudabadbozchelou2022nn}, high-speed flows \cite{mao2020physics}, multiscale flows \cite{lou2021physics}, multiphase flows \cite{wen2022u}, and multiscale bubble dynamics \cite{lin2021operator}) with/without data, simulating turbulence \cite{pang2020npinns,jin2021nsfnets}, design problems in the context of fluids \cite{zhang2020frequency}, and reconstructing high-precision flow data (super-resolution) \cite{gao2021super,li2022using}.


For instance, \cite{wen2022u} proposed U-FNO for solving parametric multiphase flow problems. U-FNO is designed based on the Fourier neural operator (FNO) and incorporates a U-Net structure to improve the representation ability in high-frequency information. Another study, \cite{cai2022physics}, reviewed physics-informed methods in fluid mechanics for seamlessly integrating data and showed the effectiveness of physics-informed neural networks (PINNs) in the inverse problems related to simulating several types of flows.

\subsubsection{Material Science}
In material science, researchers utilize physics-informed deep learning methods such as PINNs to model the optical, electrical, and mechanical properties of materials (nonhomogeneous materials \cite{zhang2020physics2}, metamaterials \cite{chen2020physics}, and elastic-viscoplastic materials \cite{arora2022physics}), as well as specific structure (e.g., surface cracks \cite{shukla2020physics}, fractures \cite{goswami2020transfer}, defects \cite{zhang2022analyses}, etc.) under the influence of external force or temperature.

For example, \cite{shukla2020physics} identified and characterized the surface-breaking cracks in a given metal plate by estimating the speed of sound inside with a PINN, which combined physical laws with the effective permittivity parameters of finite-size scattering systems comprised of many interacting multi-component nanoparticles and nanostructures. This approach can facilitate the designing new metamaterials with nanostructures. Another paper, \cite{lu2020extraction}, focused on extracting elastoplastic properties of materials such as metals and alloys from instrumented indentation data using a multi-fidelity deep learning method.



\subsubsection{Other Fields}
In addition to the above, physics-informed deep learning methods have important applications in many other fields, including heat transfer \cite{cai2021physics2,wang2021deep,bora2021neural}, waves \cite{clark2020deep,guan2021fourier,bihlo2022physics}, nuclear physics \cite{ivanov2020physics,schiassi2022physics}, traffic \cite{barreau2021physics}, electricity \& magnetics \cite{misyris2020physics,li2021physics2,chen2022physics}, and the following fields.

\textbf{Medicine.}
Physics-informed methods are used to model physical processes in the human system (e.g., blood flow \cite{kissas2020machine}, drug assimilation \cite{goswami2021study}), or the dynamics of a disease \cite{cavanagh2021physics}), and other relevant physical systems such as diagnostic ultrasound \cite{alkhadhr2021modeling}. For example, \cite{kharazmi2021identifiability} analyzed a number of epidemiological models through the lens of PINNs in the context of the spread of COVID-19. This paper studied the simulated results with realistic data and reported possible control measures. Graph neural networks based methods are used for molecular property prediction \cite{schutt2017schnet,satorras2021n,hao2020asgn,zhu2022unified} and molecular discovery \cite{jin2018junction,you2018graph, abueidda2020topology, gebauer2019symmetry, hoogeboom2022equivariant, bao2022equivariant}. 

\textbf{Geography.}
A new line of work has attempted to apply PINNs in several topics of geography, including climate \cite{kashinath2021physics, pathak2022fourcastnet}, geology \cite{zheng2020physics}, seismology \cite{karimpouli2020physics}, and pollution \cite{bukhari2022fractional}. For instance, \cite{soriano2021assessment} evaluated groundwater contamination from unconventional oil and gas development; the predictions brought many critical insights into the prevalence of contamination based on historical data.

\textbf{Industry.}
Physics-informed deep learning methods have also emerged as powerful tools in the industry. Examples include applying physics-informed methods in solving civil engineering problems \cite{vadyala2021review}, processing composites in smart manufacturing \cite{ramezankhani2022data}, and modeling metal additive manufacturing processes \cite{zhu2021machine}.


\section{Inverse Problem}
\label{sec_inverse_design}


In addition to using neural networks as a surrogate model for simulating
physical systems, there is another important and challenging task: to
optimize or discover unknown parameters of a physical system. This problem
is also called inverse problems (e.g. inverse design), and is widely used in many fields such as engineering~\cite{hoyer2019neural,kim2020designing,bi2020scalable}, design~\cite{sasaki2019topology,sun2019review}, fluid dynamics~\cite{raissi2020hidden}, etc.  Since inverse problem involves numerous scenarios and sub-problems~\cite{pilozzi2018machine,shahnas2018inverse,gorbachenko2016neural,xu2019neural}, we take inverse design, which is crucial in both academic research and industrial application, as a representative example and review methods that incorporate machine learning algorithms, especially neural networks in this section.
We first formalize the problem of inverse design and introduce the basic concepts, traditional methods, and challenges in Section~\ref{sec_inverse_design_bg}. 
Considering that the solution of inverse design usually involves multiple steps, such as the simulation of the physical system or process, the evaluation of the performance, and the representation of the configuration, we present methods according to their roles in the task of inverse design. Neural surrogate modeling of the physical system has received widespread attention and related research is introduced in Section~\ref{sec_inverse_design_surrogate}. Methods that focus on other parts of inverse design are introduced in Section~\ref{sec_inverse_design_general}. We further review methods for more general inverse problems beyond inverse design in Section~\ref{sec_inverse_problem}. Finally, in Section~\ref{sec_inverse_design_discussion} we discuss the remaining challenges and future work in this field.

\subsection{Problem Formulation}
\label{sec_inverse_design_bg}

Generally speaking, in an inverse design task, an optimal configuration of a physical system is sought to achieve the desired performance, while some given constraints, usually associated with physical properties, are satisfied. For example, both the shape optimization of the airfoil to minimize drag during flight, and the heater placement in an office to manage the temperature, are typical examples of inverse design. Considering that we have a collected
dataset of physical systems with different parameters $\mathcal{D}= \{ u
(\tmmathbf{x}_i ; \theta_j) \}_{1 \leqslant i \leqslant N_1, 1 \leqslant j
\leqslant N_2}$, the problem can be formalized as
\begin{equation}
\label{eq_inverse_design_form}
    \begin{split}
        \min_{\theta\in\Theta} \mathcal{J} (u (\tmmathbf{x}; \theta), \theta),\\
        \textit{s.t.\;} \mathcal{P}(u;\theta)(\tmmathbf{x})=0.
    \end{split}
\end{equation}
Here, $u (\tmmathbf{x}; \theta)$ are the state variables and $\mathcal{J}$ is the design objective of the physical system configured by parameters
$\theta$, where the physical process $\mathcal{P}(u;\theta)(\tmmathbf{x})=0$ represents a group of PDEs, or even constraints in other forms, e.g., explicit or implicit functions. Since inverse design is common in various complex scenarios, different problems can be formalized as either PDE-constrained optimization or, more generally, constrained optimization, depending on the form of $\mathcal{P}$. Note that, when we optimize $\theta$, the solution of PDEs $u
(\tmmathbf{x}; \theta)$ at given parameters $\theta$ is unknown and needs to be
solved using a traditional numerical solver or neural network surrogates. Here, the mathematical formulation of the inverse design is consistent with general inverse problems. If the design parameters $\theta$ are (part of ) the parameters of the physical systems, estimating optimal $\theta$ equals identifying system parameters. If $\theta$ denotes the control parameters or design parameters, then the formulation can be used for solving PDE Constrained Optimization (PDECO) problems, such as structural optimization or optimal control of PDEs.



Researchers used to adopt numerical methods to solve the inverse design formalized as an optimization problem, especially PDE-constrained optimization. Traditional methods can mainly be categorized as an  \textit{all-at-once} approach and a \textit{black-box} approach~\cite{herzog2010algorithms}. \textit{All-at-once} approaches, such as sequential quadratic programming (SQP)~\cite{boggs1995sequential}, optimize the state variables and parameters simultaneously, treating them independently, which only requires the PDE constraints to be satisfied at the end of optimization. However, when it comes to large-scale problems, \textit{all-at-once} methods become impractical. \textit{Black-box} methods, including first-order methods (e.g., gradient descent) and higher-order methods (e.g., Newton methods), use iterative schemes with repetitive evaluation of gradients $\frac{\partial\mathcal{J}}{\partial\theta}$. The adjoint method is most commonly employed to calculate the gradients. However, it requires costly solutions of the original PDEs and the adjoint PDEs with numerical solvers like FEM in every round of optimization. A rough estimate of the computational complexity of FEM can be $\mathcal{O}(dn^r)$, where $n$ is the dimension of state variables, $r$ is about $3$ for a simple solver, $d=1$ for a linear system and $d>1$ for a nonlinear system~\cite{xue2020amortized}. The high expense means that the optimization demands a large amount of computational resources with poor efficiency. Meanwhile, a large number of parameters to be optimized, which leads to higher degrees of freedom, could bring intractable complexity. Parameters in the form of continuous functions (e.g., the source function in a physical system) that are not finite-dimensional vectors also impose difficulties for the existing methods. For general constrained optimization problems, more challenges arise in some cases, including but not limited to non-differentiable physical process and lack of uniqueness~\cite{sun2021amortized}.

Based on the introduction above, problems of inverse design, i.e., identification or controlling physical systems, have been fundamental challenges because of their difficulty, low efficiency and multimodality. It is a promising direction that AI can help accelerate or improve existing methods of inverse design by introducing physics-informed machine learning paradigms.

\subsection{Neural Surrogate Models}
\label{sec_inverse_design_surrogate}

As mentioned above, the simulation of the physical system and the evaluation of the objective function often make use of traditional numerical methods like FEM, and the computational complexity and the demand for computing resources can be huge. As the scale and dimension of the system increase, the cost of numerical methods during optimization may become unacceptable. To accelerate the process, there has been interest in performing the optimization based on surrogate models instead of numerical solvers. Besides using machine learning algorithms such as random forests and Gaussian processes, more and more researchers are leveraging neural networks to model a physical system where an inverse design is conducted. The advantages of neural networks to approximate any measurable function, to handle high-dimensional and nonlinear problems, and to interpolate and extrapolate across the data contribute to its usage in the task of inverse design. Several typical paradigms of neural surrogate models are introduced below.

\textbf{With PINNs.} PINN~\cite{raissi2019physics} proposes to model the constraints of PDE system by minimizing physics-informed loss. One superiority of PINN is that it can successfully address inverse problems~\cite{chen2020physics,raissi2020hidden,tartakovsky2020physics,yazdani2020systems}, which are special cases of PDE-constrained optimization. Some research has also looked into solving inverse design or PDE-constrained optimization with PINNs describing a physics system.

Considering that PINN seamlessly introduces physical constraints to a neural network by incorporating the physics-informed loss, Mowlavi and Nabi~\cite{mowlavi2021optimal} extend the original PINN to problems like optimal control. With two neural networks representing the solution $u_w$ and the control parameters $\theta_v$, where $w\in W$ and $v\in V$, the inverse design can be solved with an augmented loss function as
\begin{equation}
\begin{aligned}
   \mathcal{L}=&\frac{\lambda_r}{N_r} \sum_{i = 1}^{N_r} \| \mathcal{F} (u_w ;
  \theta_v) (\tmmathbf{x}_i) \|^2 + \frac{\lambda_i}{N_i} \sum_{i = 1}^{N_i} \|
  \mathcal{I} (u_w ; \theta_v) (\tmmathbf{x}_i) \|^2 \\& + \frac{\lambda_b}{N_b}
  \sum_{i = 1}^{N_b} \| \mathcal{B} (u_w ; \theta_v) (\tmmathbf{x}_i) \|^2 +
  \lambda_{\mathcal{J}} \mathcal{J}(u_w(\tmmathbf{x}), \theta_v).
\end{aligned}
\end{equation}
Here, this method simply adds the objective function of the inverse design to the standard PINN loss terms with $\lambda_\cdot$ as scalar weights. To tune a series of hyperparameters, they propose a guideline for optimal control, which is categorized into validation (to ensure that the learned solution $u_{w^*}$ satisfies the PDE constraints) and evaluation (to accurately evaluate the performance of the optimized parameters $\theta_{v^*}$). This PINN-based method is compared with direct-adjoint-looping (DAL). Optimal control results in four physical systems, including Laplace, Burgers, Kuramoto-Sivashinsky, and Navier-Stokes equations, prove the capability of PINN in inverse design. Although this work mainly focuses on examining the feasibility of the original PINN in tasks of inverse design, the authors mention that the issue of balancing different objectives when training PINNs also exists for this problem.

To address the challenges due to the multiple loss terms, among which the PDE loss and the objective function are often not consistent, Lu \textit{et al.}~\cite{lu2021physics} propose PINN with hard constraints (hPINN) for inverse design, especially topology optimization. Unlike a soft-constraint method that directly minimizes the sum of PDE loss and the objective function, they take the equality and inequality constraints as hard constraints with the penalty method and the augmented Lagrangian method. The optimization objectives of these three methods are listed in order as
\begin{align}
    \mathcal{L} &= \mathcal{J}+\mu_{\mathcal{F}}\mathcal{L}_{\mathcal{F}}+\mu_{h}\mathcal{L}_{h},\\
    \mathcal{L}^k &= \mathcal{J}+\mu_{\mathcal{F}}^k\mathcal{L}_{\mathcal{F}}+\mu_{h}^k\mathbb{I}_{h>0}h^2,\\
    \mathcal{L}^k &= \mathcal{J}+\mu_{\mathcal{F}}^k\mathcal{L}_{\mathcal{F}}+\mu_{h}^k\mathbb{I}_{h>0\;or\;\lambda_h^k>0}h^2\nonumber\\&+\frac{1}{MN}\sum_{j=1}^M\sum_{i=1}^N\lambda_{i,j}^k\mathcal{F}_i(u;\theta)(\tmmathbf{x}_j)+\lambda_h^kh,
\end{align}
where $\mu_\cdot$ are coefficients, $\lambda_\cdot^k$ are Lagrangian multipliers, $h$ represents hard constraints, and $k$ denotes the $k$-th iteration, because the penalty method and the augmented Lagrangian method transform the constrained problem into a sequence of unconstrained problems.
Also, they present novel network architecture to strictly enforce the boundary conditions. The method is demonstrated with experiments on holography in optics and topology optimization in fluids.

Another study \cite{hao2022bi} proposes to adopt a bi-level optimization framework to handle the conflict PDE losses and objective loss. It adopts the following mathematical formulation:
\begin{eqnarray}
  \min_{\theta} &  & \mathcal{J} (w^{\ast}, \theta) \\
  s.t. &  & w^{\ast} = \tmop{\arg\min}_w \mathcal{L}_{PINN} (w, \theta) . \nonumber
\end{eqnarray}
It uses PINNs to solve PDE with current control variables in the inner loop and updates the control variables using hypergradients in the outer loop. The hypergradients are computed using Broyden's method based on implicit function differentiation \cite{liu2021investigating}. This method naturally links the traditional adjoint method for solving PDECO with prevailing methods based on PINNs, which has a large scope for exploration.

Other work makes use of PINNs to handle inverse design differently. ~\cite{barry2022physics} proposes Control PINN for optimal control. The network is designed to generate a triple of the system solution: the control parameters and the adjoint system state, together with spatial and temporal coordinates as input. The first order optimality condition of the Lagrangian is introduced to the standard PINN loss to perform the optimization under the constraints in a one-stage manner. ~\cite{antonelo2021physics} proposes Physics-Informed Neural Nets for Control (PINC), which modifies the network to output solutions based on initial states and control parameters, making it possible to make long-term predictions and suitable for control tasks. The model is combined with model-based predictive control (MPC) as a surrogate solver to perform control applications.

Physics-informed algorithms incorporate physical knowledge (PDEs) into the neural networks as soft constraints in loss terms. One strength is their flexibility and ease of implementation. This allows them to be extended to inverse design, by either optimizing the objective function along with PDE losses or using trained PINNs as surrogate solvers. However, the challenges concerning convergence when training PINNs can be pathological due to the imbalance among multiple loss terms, which may also exist for inverse design based on PINNs. Strategies like adaptive loss re-weighting~\cite{wang2021understanding,wang2022and} have the potential to improve the performance of optimization. Leveraging the advantages of traditional numerical methods for PINN methods may also ease the problems.

\textbf{With Neural Operators.} As introduced in Section~\ref{sec_neural_operator}, neural operators learn a mapping $G:\Theta\times\Omega\rightarrow\mathbb{R}^m$, which is a mapping from the input parameters/functions to the solution function under the physical constraints and can be queried for state variables at any arbitrary point in the spatio-temporal domain. This replaces the numerical PDE solvers or expensive physics simulators that are repeatedly called on during optimization.

Amortized Finite Element Analysis (AmorFEA) is developed by Xue \textit{et al.}~\cite{xue2020amortized}. It is inspired by the idea of amortized optimization in amortized variational inference. Its purpose is to predict PDE solutions with a neural network, based on the Galerkin minimization formulation of PDE. The neural network is trained to minimize the expected potential energy over the parameter space, from which the PDE can be derived. With the trained surrogate model, gradient-based optimization can be performed with only one forward and backward pass through the network to compute the gradients w.r.t. the control parameters. The authors conducted experiments on source control of a Poisson equation and inverse kinematics of a soft robot. The fact that not all PDE can be derived from potential energy limits the scope of applications in different physics systems.

Wang \textit{et al.}~\cite{wang2021fast} use physics-informed DeepONets (see Section~\ref{sec_operator_deeponet}) as a surrogate to achieve fast PDE-constrained optimization via gradient-based methods, even without paired training data. The self-supervised mechanism with physics-informed losses as soft constraints enables effective training with no need to call on numerical solvers. A two-stage framework is proposed to perform the optimization. It first trains the surrogate model for the given physical system and then minimizes the cost function w.r.t. the input functions, which can be described as
\begin{align}
    &w^\ast = \mathop{\arg\min}_{w\in W}\mathcal{L}_{PINO},\\
    &\min_{\theta} \mathcal{J}(G_{w^\ast}(\theta)(\tmmathbf{x}),\theta),
\end{align}
where $\mathcal{L}_{PINO}$ is defined in Equation~\ref{eq_pideeponet}. An additional neural network with learnable weight $\alpha$ is used to parameterize the input functions. Experimental results on optimal control of Poisson and heat equation and drag minimization of obstacles in a Stokes-flow demonstrate the capability of physics-informed DeepONets to perform PDE-constrained optimization. Compared with numerical solver and adjoint methods, the computational cost of their proposed method is much lower. However, the ability of physics-informed DeepONets to accurately solve complex PDE systems is the bottleneck in terms of extending this method to more complex scenarios.

A two-stage framework similar to that used by Wang \textit{et al.} is proposed by Hwang \textit{et al.}~\cite{hwang2022solving}. The first phase is to learn the solution operator under the PDE constraints with an autoencoder model. The decoder has two branches, one for solution $G_w(\theta)=(G^{sol}_w\circ G^{enc}_w)(\theta)$ and one for reconstruction $\tilde{\theta}_w(\theta)=(G^{rec}_w\circ G^{enc}_w)(\theta)$. The surrogate model can be trained in either a data-driven or data-free way. The second phase is to minimize the objective function based on the trained neural operator. To avoid the optimal parameters being outside the training domain, the reconstruction branch of the encoder is used to regularize the optimization. The whole framework can be described as
\begin{align}
    &\mathcal{L}_{sup} = \frac{1}{N}\sum_{k=1}^N L(G_w(\theta_k)(\tmmathbf{x}),u_k(\tmmathbf{x})),\\
    &\mathcal{L}_{res} = \frac{1}{N}\sum_{k=1}^N l_r^i+l_i^k+l_b^k,\\
    &\mathcal{L}^w_{rec} = \frac{1}{N}\sum_{k=1}^NL(\tilde{\theta}_w(\theta_k),\theta_k),\\
    & w^\ast = \mathop{\arg\min}_{w\in W} \mathcal{L}_{sup/res}+\lambda_1 \mathcal{L}^w_{rec},\\
    & \min_{\theta} \mathcal{J}(G_{w^\ast}(\theta)(\tmmathbf{x}),\theta)+\lambda_2 \mathcal{L}^{w^\ast}_{rec},
\end{align}
where $L$ is L2 relative error and $\lambda_{1/2}$ are hyperparameters. The authors also provide a theoretical analysis of the error estimates in the optimization phase in terms of the error in the training phase. They conducted experiments on source control of the Poisson equation, boundary control of the stationary Stokes equation, inverse design of a nonlinear wave equation, and force control of Burger's equation.

Sun \textit{et al.}~\cite{sun2021amortized} propose a two-stage neural network architecture that resembles an autoencoder to address the task of synthesis. This can be considered as a generalization of PDE-constrained optimization with state variables, referred to as realization in~\cite{sun2021amortized}, formalized as implicit functions $G_w(\theta)$ of input parameters $\theta$. The framework can be described as
\begin{align}
    &w^\ast = \mathop{\arg\min}_{w\in W}||G_w(\theta)(\tmmathbf{x})-\tilde{G}(\theta)(\tmmathbf{x})||,\\
    &\phi^\ast(\cdot) = \mathop{\arg\min}_{\phi(\cdot)}\mathbb{E}_{g\sim\mathcal{D}_{g}}\mathcal{J}_g(G_{w^\ast}(\phi(g))(\tmmathbf{x}),\phi(g)), 
\end{align}
where a neural network $\phi(\cdot)$ is trained to map the desired target of design $g$ to the design parameters $\theta=\phi(g)$ on the distribution $\mathcal{D}_g$ of $g$ with $\mathcal{J}_g$ as the corresponding objective function for $g$. Instead of the iterative scheme used by previous methods that directly optimize according to the given target via gradient descent, this method trains another network to predict the optimal design from the target. The performance is evaluated on the fiber extruder path planning of a 3D printer and the inverse kinematics of a soft robot. The method is proven to have performance close to AmorFEA, while the speed is faster by orders of magnitude. It also addresses the challenges of the non-differentiability of physical processes and the non-uniqueness of the design solution. The limitations of the approach are that the data-driven training demands a substantial number of samples and the objective needs to be quantifiable to be sent into the network.

Methods based on neural operators commonly address the task of inverse design with a two-stage framework. The operator is trained to solve the PDE system and then used as a differentiable surrogate for an approximate solution to perform optimization via gradient descent or neural prediction. The sequential process usually shows its superior efficiency when the objectives of design over the control parameters in the physical system are potentially changeable and diverse because the surrogate can be trained only once for different targets. However, the challenges remain in the training of neural operators for large-scale and complex physical systems in terms of data generation and training stability. Also, gradient-based optimization may have no effect on discrete and non-differentiable parameters.

\textbf{With Neural Simulators.} In comparison with neural operators, neural simulators refer to neural networks that learn a mapping $S:\Theta\rightarrow\mathbb{R}^n$. This mapping is from the input parameters/functions to the values of interest, which evaluate the specific performance or even directly evaluate the objective functions. The effect of these methods is similar to that of neural operators, i.e., to avoid expensive numerical simulations.

For PDE-constrained optimization, Lye \textit{et al.}~\cite{lye2021iterative} propose Iterative Surrogate Model Optimization (ISMO), an active learning algorithm that combines a quasi-Newtonian optimizer with a neural network as a surrogate. The basic idea is to adopt a neural network to model the map between the parameters and the observables (e.g., lift and drag coefficients for an air foil) with supervised learning, and to perform the standard optimization based on the approximate objective predicted by the trained model. To guarantee accurate approximation around the (local) minima of the cost function, ISMO augments the training set regularly with additional points around the estimated local minima, which are decided by the quasi-Newton method. 
Compared with surrogates using neural operators, ISMO only substitutes a numerical PDE solver with neural surrogate during the optimization, but still needs an expensive numerical solution when generating the training set, which dominates the cost of the algorithm.

Besides dynamic systems described by PDEs, data-driven surrogate models with neural simulators are more common for general physical systems where the physical processes are too complex to be embedded into the model or to be accurately formalized in mathematics, especially in structural optimization~\cite{hoyer2019neural,kim2020designing,bi2020scalable}, rotating machines~\cite{tucci2019deep,sasaki2019topology,doi2019multi}, etc. The architectures and approaches of neural simulators vary according to different scenarios and design objectives. In~\cite{kim2020designing}, a CNN and a fully-connected network are trained to respectively classify the type of detachment mechanism and predict interfacial stress distribution with the input of an adhesive pillar shape, and genetic optimization on the pillar shape is further performed. In~\cite{bi2020scalable}, a neural network is employed for a surrogate gradient given a specific design by learning the iterative history data. In addition, parallel computing is adopted in the learning process to enable online updating and speed up the topology optimization. For rotating machines, CNN is often trained based on cross-sectional images and evolutionary optimization, including a genetic algorithm (GA), is used for optimization. \cite{sasaki2019topology,doi2019multi} use GA to optimize the IPM motor based on the classification result of CNNs and FEM is used to evaluate the individual with a certain probability. \cite{doi2019multi} focuses on a multi-objective topology optimization concerning average torque and torque ripple. \cite{tucci2019deep} leverages an autoencoder to reduce dimension and maps the latent representation to the torque with a CNN. In this way, it eases the challenge that evolutionary optimization is not suited to deal with search in high-dimensional space. Neural networks as neural simulators to accelerate inverse design also have wide applications in different fields, such as aerodynamics~\cite{sun2019review}, photonics~\cite{pestourie2020active, peurifoy2018nanophotonic}, mechanics~\cite{messner2020convolutional}, etc.

Similar to neural operators, methods based on neural simulators provide a surrogate for system simulation and evaluation, especially for complex and non-differentiable systems, where the inverse design is often formalized as general constrained optimization. The surrogate models cannot be trained in a self-supervised manner like PINN or Physics-informed DeepONet, but must be trained with simulation data in most cases. This means that time-consuming data generation for network training is necessary as well. Also, taking specific values of interest as the output of neural surrogate makes it less flexible in terms of transferring across different optimization tasks. FEM is sometimes used to simulate or evaluate during optimization even with a neural surrogate~\cite{sasaki2019topology,doi2019multi}, which means that the efficiency issue of numerical simulation remains.

\subsection{Other Methods}
\label{sec_inverse_design_general}

Besides training neural networks as surrogate models to describe physical systems or embed physical knowledge, some other research leverages neural networks to play different roles in the process of inverse design. Several types of methods are listed below. 

\textbf{Neural Representation.} Parameterizing the parameters to be optimized with neural networks instead of in their original space may take advantage of the expression of neural networks in high dimensions, in order to achieve more highly-detailed and continuous representations. This is especially the case in structural topology optimization, where conventional methods are usually grid-based or mesh-based~\cite{hoyer2019neural,zehnder2021ntopo}. This approach can be combined with methods of neural surrogate models, as some research has already used neural networks to parameterize the control signals along with the surrogates~\cite{wang2021fast,mowlavi2021optimal}. The characteristics of the parameters may introduce inductive bias to further increase the ability of neural representation.

\textbf{Design Prediction.} Neural surrogate models substitute the usage of numerical methods for the simulation of physical systems and the evaluation of optimization goals as forward modeling, i.e., the solutions or outcomes to physical constraints such as PDEs. Reverse methods are also proposed to predict the optimal design from  optimization targets. In~\cite{sun2021amortized}, a network mapping from the desired target to the design parameters is trained along with a neural operator as surrogate models in a self-supervised way. Moreover, using neural networks to perform inverse design directly in a data-driven manner is also common in fields such as photonics~\cite{liu2018training}, airfoil optimization~\cite{sekar2019inverse} and materials~\cite{abueidda2020topology}.

\textbf{Data Generation.} ~\cite{yamasaki2021data} proposes data-driven topology design using deep generative models. Given initial material distributions, the trained VAE is used to generate novel samples, which are diverse but with inherited features of the training data, and may have superior performance compared to the original ones. By iterating the process, a series of satisfactory solutions can be obtained.

\subsection{Beyond Inverse Design}
\label{sec_inverse_problem}

In fact, inverse design is a realistic application of inverse problem, which also includes property identification, parameter inference and scientific discovery, which are critical problems in scientific researches. The formulation is consistent with Eq.~\eqref{eq_inverse_design_form}, while the objective function $\mathcal{J}$ usually describes the difference between the neural solutions and observations.

The original PINN~\cite{raissi2019physics} has demonstrated its potential in inverse problem of property identification and parameter inference under the unified framework with physics-informed losses. Then it's widely adopted to study inverse problems in various fields. In material science, physics-informed deep learning methods are used to discover new materials or identify properties of given materials from experimental/simulated data (inverse problems) \cite{lu2020extraction, haghighat2021physics}. In fluid dynamic, physics-informed algorithms are taken to learn physical quantities such as velocity and pressure from flow data (inverse problems) \cite{raissi2020hidden,jin2021nsfnets,mao2020physics,lou2021physics}. Specifically, \cite{raissi2020hidden} developed a physics-informed neural network framework to infer physical quantities of interest such as velocity and pressure from the spatial-temporal visualizations of fluids. They combined known physical priors (namely governing equations) with available data to draw inferences that may be useful for physical or biological problems where direct measurement is infeasible. 

Scientific discovery is another class of inverse problems, which intends to discover new physical laws or determine unknown parameters of the physical system from data. Physics-informed deep learning methods such as PINNs and neural operators are competitive choices to solve such problems \cite{bar2019unsupervised,takeishi2021physics}, especially for the cases of high dimension \cite{mackinlaymodel} or scarce and noisy data \cite{chen2021physics}. The authors of \cite{raissi2018deep} used two neural networks trained in a physics-informed manner with some noisy observations to discover the governing equation (which was originally partially known) as well as its solution. Besides, other methods are also proposed to address this task. For instance, PDE-Net~\cite{long2018pde} learns the specific form of PDE from data, while ~\cite{cranmer2020discovering} not only represented the unknown governing equation (physical laws) with a graph neural network in a supervised setting, but also extracted analytical expressions from the learned model via symbolic regression.

\subsection{Open Challenges and Future Work}
\label{sec_inverse_design_discussion}

Compared to the problems of neural solvers and neural operators, inverse design is a more complex and comprehensive task with multiple technical parts involving representation, simulation and optimization. Therefore, the methods, which adopt machine learning, especially deep learning to address inverse design, are more diverse. Besides methods discussed in previous sections, there are still other potential options to incorporate neural networks to handle inverse design, such as using reinforcement learning~\cite{jang2022generative}. 

The remaining challenges of inverse design are also various. We summarize some typical open problems as follows. 
\begin{itemize}
    \item \textbf{Neural Surrogate Modeling } Some of the open challenges are consistent with those of neural solvers and operators, including the balance of multiple loss terms and training convergence for physics-informed methods, the large amount of data demand for operator and simulator, etc. These were introduced in detail in the previous sections. Addressing these challenges may contribute to the development of machine learning algorithms for inverse design.
    \item \textbf{Large Scale Application } Inverse design are related to abundant practical application scenarios, where the physical systems have a large number of parameters to be optimized. This could lead to other challenges including but not limited to issues of curse of dimensions, extensions to large-scale scenarios and computational complexity of optimization.
    \item \textbf{Other Directions } Since solving the problems of inverse design involves multiple steps, the usage of neural network in different parts can be further exploited. Improving the efficiency of data utilization, incorporating the physical knowledge into the framework of inverse design and fusing with traditional numerical methods are potential directions for future researches.
\end{itemize}

\section{Computer Vision and Graphics}
\label{sec_computer_vision}
In this section, we describe how physical knowledge can be incorporated in computer vision and computer graphics.

\subsection{Problem Overview}

Computer vision involves diverse tasks including the prediction, generation, and analysis of digital images, videos, and even data like point clouds or sensor measurements, while computer graphics aims at rendering vivid scenes where physics law must be satisfied to achieve sense of reality. Both focus on the perception, generation and interaction with real-world environments, which are governed by plentiful physical rules. 

However, current data-driven deep learning methods are not sufficient. Deep neural networks trained from natural or synthesized datasets perform based on the underlying patterns of data, and could encounter challenges concerning stability, reliability and security, e.g., out-of-distribution problem (OOD)~\cite{shen2021towards,ye2022ood}, adversarial examples~\cite{dong2021adversarial,chen2021unrestricted}. Incorporating deep learning models with physical knowledge can further constrain the prediction to follow the laws and guarantee the preciseness. For instance, in motion tracking and pose analysis, physical constraints or simulations~\cite{yi2022physical,gartner2022trajectory} can boost the models’ physical plausibility and correctness to reduce unrealistic errors. 

Meanwhile, the interpretability of deep learning is still an open problem in machine learning~\cite{zhang2018visual}. The black-box nature of neural networks limits the trust in their grounded applications, especially in security-sensitive scenarios like auto-driving~\cite{zhang2022make,michaelis2019benchmarking} and medicine~\cite{finlayson2019adversarial}. Empowered by physical knowledge, the performance of models can be partially expressed and explained by formalized rules. In consecutive predictions with videos, recurrent predictors based on dynamical equations could restrict the future trajectories and improve the long-term performance~\cite{toth2019hamiltonian}.

\subsection{Physics-informed Computer Vision and Graphics}
Attempts to incorporate physical knowledge in deep learning based computer vision and computer graphics are still primary. We will conclude some methods in the following context, according to the parts of machine learning model to introduce physics.

\subsubsection{Data}

Latest works\cite{brown2020language,he2022masked} have demonstrated the capability of deep learning models to form learned priors and experiences from massive data. Directly introducing training data following the physical rules can help models learn the underlying physical structure and distribution, which is categorized as ``observational bias'' by~\cite{karniadakis2021physics}. 

There have been some datasets describing physical objects and scenes~\cite{gao2021objectfolder,gao2022objectfolder,wu2016physics}. In~\cite{mottaghi2016newtonian}, Visual Newtonian Dynamics (VIND) dataset is compiled, including more than 6000 videos aligned with Newtonian scenarios represented using game engines, and more than 4500 still images with their ground truth dynamics. This could help models understand the objects in a static scene in terms of the forces acting
upon it and its long term motion as response to those forces. Physics101 dataset~\cite{wu2016physics} contains thousands of video clips recording objects in simple physical scenarios like collision, to study the automatic learning of object-centric physical properties. Physical Concepts dataset~\cite{piloto2022intuitive} follows the violation-of-expectation (VoE) paradigm in developmental psychology and collects videos with VoE probes involving basic physical concepts like continuity, persistence, solidity. This corresponds to the learning of intuitive physics, also named naïve physics, that is semantically meaningful but hard to be directly incorporated in the learning procedure. There are also datasets~\cite{gao2021objectfolder,gao2022objectfolder} gathering objects with various physical properties.

\subsubsection{Architecture}

The architecture design of neural networks inspired or guided by physics can improve the efficacy and interpretability of models. In fact, the convolution operation in CNN~\cite{lecun2015deep} is based on the translation invariance and has achieved great success. 

There could be more space to for improvement when more complex physical knowledge comes in. For instance, Equivariant Object detection Network (EON)~\cite{bao2022equivariant} introduces object-level rotation equivariance to 3D object detection by designing a rotation equivariance suspension design to extract the equivariant seed features. In scenarios more specific to physical process, some works adopt neural networks as tool to express physical quantities and evolutions. In visual prediction for scenarios involving physical phenomena like collision, a typical way is to use an encoder to extract physical properties or states from images and perform future prediction with a dynamics predictor using different strategies, e.g. recurrent network~\cite{watters2017visual,piloto2022intuitive}, physical engines~\cite{wu2017learning}, optical flow~\cite{ye2018interpretable}, Hamiltonian Generative Network~\cite{toth2019hamiltonian}. The constraints on latent physical properties also vary. \cite{wu2017learning} proposed to explicitly estimate the physical object states and \cite{ye2018interpretable} preferred learning entangled representations while distinguishing whether the state is physical variable or intrinsic variable through dimension. Besides, Interaction networks~\cite{battaglia2016interaction} use a graphical representation of objects and relationships and model the effects of relations by a neural network. The current state of an object, aggregate effect of relations, and the external effects are recurrently fed into another neural network to predict next state. This corresponds to the evolution of Newtonian mechanics, where we aggregate the forces exerted by other objects or by an external force to determine acceleration, and update the state by the acceleration.

\subsubsection{Loss Function}

By embedding prior physical knowledge into the loss function of a machine learning model, it can guide the learning procedure and restrict the model to satisfy physics constraints. 

HNN~\cite{greydanus2019hamiltonian} uses Hamiltonian equations as the objective function to constrain the neural network to learn the quantity of Hamiltonian in the system, as introduced in Sec.\ref{sec_pinn_arch}. This is similar to PINN~\cite{raissi2019physics} and performs as soft constraint. Besides adopting formalized physical equations directly as objectives, another way is to reconstruct the observations following the physical process and take the reconstructed results to supervise the training. Neural radiance field (NeRF)~\cite{mildenhall2020nerf} models the volume density and a view-dependent color for any given position-view pair. With classical volume rendering techniques, we can generate differentiable images following physical law and train the network supervised by ground-truth inputs. This general framework is also utilized in various visual tasks, for either supervision or regularization. In deblurring, Reblur2Deblur \cite{chen2018reblur2deblur} reconstructs the blurry input by physics-based reblurring with an optical flow network and construct a self-supervised loss term as regularization. In computer graphics, PhyIR~\cite{li2022phyir} addresses the inverse rendering of complex material with a physics-based in-network rendering module to physically re-render realistic reflectance, which serves as a constraint by a re-render loss to realistic lighting effects. Physics simulator can also be used for supervision. For motion capture\cite{huang2022neural}, a physics simulator is taken as a supervisor to train a motion distribution prior for further motion sampling to capture physically plausible human motion from a monocular color video. Other forms of physics regularization are also possible, for example in depth estimation\cite{fei2019geo}, image dehazing\cite{morales2019feature}, 3D reconstruction\cite{chen2022aug}.

\subsubsection{Inference}

Another option is to impose physical constraints at inference stage, to adjust and correct the predictions to reach physical feasibility. 

In motion and pose analysis, physics knowledge of kinematics and dynamics, usually integrated into physics simulators, along with intuitive physical common sense are often embedded into the pipeline of inference. \cite{gartner2022trajectory} uses feature-complete physical simulation as a building block to incorporate more subtle physical effects such as sliding and rolling friction, or surfaces with varying degrees of softness. Physical Inertial Poser (PIP)~\cite{yi2022physical} achieves real-time motion tracking with satisfying accuracy using only six inertial sensors. The tracking is performed by combining a neural kinematics estimator and a physics-aware motion optimizer. The estimated kinematic parameters of motion status are further refined by the optimizer according to constraints of motion equation, friction cone, and the condition of no sliding.

\subsection{Open Challenges and Future Work}

There have been some primary explorations to combine computer vision with physical knowledge, while open challenges remain and demand future studies into them. Some of them are as follows.

\begin{itemize}
    \item \textbf{Enforcing Symmetry Constraints.} A lot of current papers try only to imitate the form of physical dynamics and ignore the underlying symmetry. For example, if we incorporate the translation invariance into the network design, the hypothesis space can be reduced, and we can train the neural networks more efficiently.
    \item \textbf{Learning Meaningful Representations.} When encoding images into a latent space or representing the effects from one object to another, the physical meanings of these representations are hardly examined. If we can map from image-based dynamics exactly to equation-based dynamics with clear physical explanations, we will be able to do simulation, dynamics discovery, and extraction of physical variables at the same time.
    \item \textbf{Formulated Representations of Intuitive Physics.} For vision tasks in daily scenarios, many rules of motion, collision, and interaction are described with intuitive physics, rather than rigorous physical equations. However, the unformulated representations of intuitive physics make it limited to utilize the knowledge in the learning framework, which is usually used in the form of constraints.
\end{itemize}

\section{Reinforcement Learning}
\label{sec_reinforcement_learning}
In this section, firstly we introduce the basic setting and the goal of reinforcement learning. Then, we discuss how physical knowledge can be incorporated into reinforcement learning.

\subsection{Problem Formulation}
Reinforcement learning is about an agent learning to act in an unknown environment to maximize its reward. One classic formulation of reinforcement learning is based on Markov Decision Processes (MDPs). An MDP defines the effects of all actions in all states of the environment, including the state transitions and rewards received by the agent.

Mathematically, an MDP can be defined as a tuple $(\mathcal{S}, \mathcal{A}, T, R, \gamma)$. Here, the state space $S$ is the set of all possible states in the environment, and the action space $A$ is the set of all possible actions that the agent can take. The transition function $T:\mathcal{S}\times\mathcal{A}\rightarrow\mathcal{S}$ defines the state transition from a given state when the agent performs a given action. The reward function $R:\mathcal{S}\times\mathcal{A}\rightarrow\mathbf{P}(\mathbb{R})$ defines the distribution of rewards when the agent performs a given action in a given state. $\gamma$ is a discount factor that balances short-term and long-term goals by discounting future rewards. The goal of the agent is to learn a policy $\pi: \mathcal{S}\rightarrow\mathbf{P}(\mathcal{A})$ to maximize the expected cumulative discounted reward $\sum_{t}\gamma^tr_t$, where for all $t$, $r_t\sim R(s_t, a_t), s_{t+1} = T(s_t, a_t), a_t\sim\pi(s_t)$.

Difficulties in reinforcement learning include sequential correlation between interactions and the environment, as well as learning by reward signals instead of learning by some "ground truth". As a result, a reinforcement learning algorithm usually learns by trial and error. In model-based reinforcement learning, we additionally learn a model of the environment to help the learning of the agent. In safe reinforcement learning, we want to avoid disastrous outcomes during training, so we may want to add some constraints to our policy.

Next, we discuss three aspects of incorporating physical knowledge: policy training, model training, and exploration guiding.

\subsection{Policy Training}
Many reinforcement learning tasks originate from real-world problems, and our understanding of these problems can help us design a friendly task to accelerate the training of the policy.

~\cite{yang2021reinforcement} focuses on reinforcement learning for mesh refinements when learning solutions of PDEs, which is inspired by adaptive mesh refinement (AMR). It uses the change in errors to guide the selection of refinements and achieve high error reductions.

~\cite{albarran2018measurement} considers a task where we want to adapt to an unknown reference quantum state by measurements. The method naturally provides many copies of the reference state, and it uses knowledge of quantum mechanics to design reward functions. In this way, it can achieve a high average fidelity with only hundreds of iterations.

~\cite{mirhoseini2020chip} introduces reinforcement learning to chip placement. It defines a reinforcement learning task to simulate the placement, including hard constraints on density a reward function dependent on the proxy wire length and congestion after the whole placement process. Experiments show that the algorithm can generate superhuman placements in under 6 hours.

\subsection{Model Training}
In model-based reinforcement learning, when the model is corresponding to parts of the physical world, our knowledge can help us to build a better model.

A recent work~\cite{lutter2021differentiable} incorporates the form of continuous dynamics into the learning of environment models and transfers the reinforcement learning agent in the environment model to the real environment. It demonstrates that physical models can work better in offline model-based reinforcement learning than black-box models, probably because of better extrapolation power. Similarly,~\cite{xie2016model} also estimates the dynamics based on a basic form of continuous dynamics via least squares. Besides, it plans over the estimated dynamics together with model-predictive control (MPC) to achieve a real-time policy with high performance.

In visual model-based reinforcement learning, the Object-Centric Perception, Prediction, and Planning (OP3) framework~\cite{veerapaneni2020entity} extracts representations of a variable number of objects and aggregates estimated effects from different sources to predict the next state. By planning over the model, their algorithm can outperform a state-of-the-art video prediction model at that time.

\subsection{Exploration Guiding}
In safe reinforcement learning, we need to add constraints to the policy as guidance in exploration. Usually, we need to use physical knowledge to find suitable constraints such that some safety requirements are satisfied.

~\cite{fulton2018safe} introduces differential dynamic logic informed by expert knowledge of the environment to create a sandbox for reinforcement learning. Specifically, it uses a logic formula to model safe transitions, and filter out actions that are not safe. If the formula is precise, the agent can get higher rewards without being unsafe during training.

The RL-CBF algorithm~\cite{cheng2019end} uses expert knowledge to construct a Gaussian Process (GP), which models the state-dependent uncertainty in an environment. The safe policy is the solution to a robust optimization problem over a confidence region of the Gaussian process assuming that the uncertainty is adversarial. In this way, it can approximately guarantee probabilistic safety only with graceful compromises.~\cite{fisac2018general} also uses a GP to describe uncertainty. They use the Hamilton–Jacobi–Isaacs (HJI) variational inequality instead to solve a robust policy and switch back to the robust policy when the agent is going to a dangerous region.

~\cite{ma2021model} views safe reinforcement learning as a constraint optimization problem, and introduces the generalized control barrier function (GCBF) together with the constraint policy gradient method to enhance the safety of the policy. In this way, the method can converge faster with fewer constraint violations.

\subsection{Open Challenges and Future Work}
Reinforcement learning has achieved impressive progress with physical knowledge, but there are still some open challenges to be solved in this field, which are shown below.
\begin{itemize}
    \item \textbf{Generic Modeling of Physical Tasks.} Currently, we mainly convert physical tasks into reinforcement learning tasks in a case-by-case way. If there is a general framework, we can easily deal with new physical tasks with proper expert knowledge. To do this, we may start with building a general framework within a subfield, and then try to expand it to a larger domain.
    \item \textbf{Solving High-Dimensional Problems.} A lot of physical problems are inherently high-dimensional, but reinforcement learning problems with high-dimensional state spaces are not sample-efficient, or even not tractable. To alleviate this, we usually resort to state space compressing such as doing reinforcement learning in a latent space constructed by a world model. However, learning a compact, meaningful latent space remains an open challenge. Besides, if the problem itself is already compact enough, we may only approximate the best policy by defining subproblems and combining their solutions.
    \item \textbf{Guaranteeing Safety in Complex, Uncertain Environments.} In safe reinforcement learning, there is still a trade-off between environment complexity and safety guarantee. In a complex, uncertain environment (e.g., there are transition noises estimated by a Gaussian Process over the state space), we can not get a probabilistic safety guarantee before training without strong assumptions. Even when we enable run-time checking, safety may still be compromised, or the checking itself does not scale to high-dimensional state spaces.
\end{itemize}

\section{Conclusion}
\label{sec:conclusion}

In this review, we have systematically investigated and summarized the field of physics-informed machine learning as seen through the eyes of machine learning researchers. First, we have identified and introduced the general concept of physics-informed machine learning. We suggest that there are several types of physical prior , i.e. PDEs/ODEs/SDEs, symmetry constraints and intuitive physics. They could be embedded into different parts of machine learning models, i.e. data, architecture, loss functions, optimization methods and inference algorithms. 
Then, we exhaustively presented existing methods, challenges, and future directions for these problems. Most of existing works focus on using neural networks for solving or identifying systems governed by PDEs/ODEs, i.e. neural simulation and inverse design. We have summarized the progress of these methods in detail.

From a methodological perspective, there are many open challenges for problems of physics-informed machine learning.
\begin{itemize}
    \item How to design a standardized dataset for problems of different physical priors is an open challenge. Datasets and benchmarks provide a fair environment to compare different algorithms and inspire researchers to discover new ones. In the field of physics-informed machine learning, such datasets are lacking due to the diversity of physical priors. A valuable, realistic dataset or benchmark containing either a single level or multiple levels of physical prior will play important role in boosting the machine learning community.
    
    \item Designing better optimization and inference algorithms incorporating or informed by physical priors is a valuable topic. While there are many works exploring model architectures inspired or to represent physical prior, optimization methods and inference algorithms receive less attention. Different from constraining the hypothesis space, novel optimization methods incorporating physical prior might be another choice to train physics-informed machine learning models. Moreover, when models are pre-trained, we might also need better inference algorithms to ensure the output satisfies physical laws.

    \item Scalable algorithms incorporating data with real-world physical prior or intuitive physics is a basic issue for building intelligent systems capable of real-world interaction. It might be a prevailing trend for developing machine learning models driven by both large data and physical prior with a broad range of applications in computer vision and robotics control. 
\end{itemize}

From the perspective of tasks of physics-informed machine learning, we also suggest several research directions and opportunities.
\begin{itemize}
    \item For neural simulation, existing optimization techniques and architectures are far from optimal. There is still a gap between PINNs and highly specialized traditional numerical methods like FEM/FVM/FDM/Spectral methods on both speed and accuracy. Promising future research directions include inventing novel and effective optimization targets, learning paradigms and neural architectures. 
    \item Inverse problems are fundamental problems in scientific discovery, computer vision as well as many other engineering domains. Many works show that methods based on PINNs and DeepONets achieve better results on inverse problems compared with traditional methods. However, the complexity and ill-posedness requires learning algorithms that better utilize data and physical knowledge. Moreover, inverse design, as a specific type of inverse problem has many  promising application scenarios in structural/topology optimization, optimal control and molecular/drug discovery. 
    \item For computer vision and reinforcement learning in real world, we need better algorithms for inference stage or training stages that incorporates physical prior. A possible direction is to model the real world environment with physical prior. The emergence of concepts like world models or NERF provides a possible method that we could learns the real world environment from data with the help of physical prior. Such models could then be used for training downstream models interacting with the world.
    
    \item  Theoretical analysis like convergence and generalization ability for physics-informed machine learning algorithms are still at a beginning. It is still a challenging task due to the difficulty of analyzing the training process of neural networks. We even don't know what is the theoretical benefits of introducing physical prior into machine learning.
\end{itemize}

We conclude that physics-informed machine learning will be a fundamental and essential topic of AI. There is still much potential for improving current methods of physics-informed machine learning.

\appendices




\ifCLASSOPTIONcaptionsoff
  \newpage
\fi



\bibliographystyle{IEEEtran}
\bibliography{ref.bib}
\end{document}